\PassOptionsToPackage{prologue,dvipsnames}{xcolor}
\documentclass{article}
\usepackage[preprint]{spconf}
\usepackage{spconf,amsmath,graphicx,booktabs,tikz, xspace}

\usetikzlibrary{calc, fit, backgrounds}

\usepackage{ifthen} 
\usepackage{hyperref}
\usepackage{cleveref}

\renewcommand{\paragraph}[1]{\noindent\textbf{#1}}
\newcommand{\argmin}{\operatornamewithlimits{argmin}}
\def\x{{\mathbf x}}
\def\u{{\mathbf u}}
\def\v{{\mathbf v}}
\def\f{{\mathbf f}}
\def\d{{\mathbf d}}
\def\T{{\mathbf T}}
\def\L{{\cal L}}

\newcommand{\textfrac}[2]{$\frac{#1}{#2}$}
\copyrightnotice{}
               \toappear{\parbox{\textwidth}{ \footnotesize\linespread{1}\selectfont
               \copyright\ IEEE 2025. Personal use of this material is permitted. Permission from IEEE must be obtained for all other uses, in any current or future media, including reprinting/republishing this material for advertising or promotional purposes, creating new collective works, for resale or redistribution to servers or lists, or reuse of any copyrighted component of this work in other works.}}
\title{MS-RAFT-3D: A Multi-scale Architecture for\\ 
Recurrent Image-based 
Scene Flow 
}
\newcommand{\anonymous}[1]{#1} 

\name{\anonymous{Jakob Schmid \qquad Azin Jahedi \qquad Noah Berenguel Senn \qquad Andrés Bruhn}\thanks{\anonymous{The authors thank the Deutsche Forschungsgemeinschaft (DFG, German} \anonymous{Research Foundation) -- Project-ID 251654672 -- TRR 161 (B04) for fund-} \anonymous{ing. Furthermore, Azin Jahedi thanks the International Max Planck Research} \anonymous{School for Intelligent Systems (IMPRS-IS) for supporting.}}}
\address{\anonymous{Computer Vision Group, Institute for Visualization and Interactive Systems,
University of Stuttgart}}
\begin{document}
%
\maketitle
\begin{abstract}
Although multi-scale concepts have recently proven useful for recurrent network architectures in the field of optical flow and stereo, they have not been considered for image-based scene flow so far. 
Hence, based on a single-scale recurrent scene flow backbone, we develop a multi-scale approach 
that generalizes successful hierarchical ideas from optical flow to image-based scene flow. 
By considering suitable concepts for the feature and the context encoder, the overall coarse-to-fine framework 
and the training loss, we succeed to design a scene flow approach that outperforms the current state of the art on 
KITTI and Spring by 
8.7\%
(3.89 vs.\ 4.26)
and 
65.8\% 
(9.13 vs.\ 26.71), respectively. 
 Our code is available at 
 {\url{https://github.com/cv-stuttgart/MS-RAFT-3D}}.
\end{abstract}
\section{Introduction}
\label{sec:intro}

With applications such as autonomous driving \cite{Menze2015_KITTI}, scene understanding \cite{Mustafa2020_SceneFlowDynScences} and robot-assisted surgery \cite{Chen2024_SceleFlowSurgicalRobot} the estimation of 3D motion from RGB-D or stereo image sequences plays an important role in computer vision. The corresponding motion field, which is also referred to as scene flow \cite{Vedula1999_SceneFlow}, provides valuable information on the interaction of objects in 3D space that cannot be inferred by computing 
its projection
on the 2D image plane, the so-called optical flow.

While classical variational methods \cite{Li2008_VariationalSceneFlowMS,Basha2010_3DSceneFlow} and early learning based approaches \cite{PWOC-3D} for computing the scene flow were traditionally based on coarse-to-fine approaches -- both to handle large displacements and to make the estimation more robust w.r.t to the initialization -- the success of recurrent neural networks in optical flow estimation triggered by the single-level RAFT architecture \cite{RAFT} lead to the development of similar single-scale approaches also in the field of scene flow \cite{RAFT-3D,CamLiFlow,CamLiRAFT}.
In this context, RAFT-3D 
\cite{RAFT-3D} was proposed that combines RAFT's recurrent updates based on a hierarchical all pairs cost volume with a pointwise SE(3)-based rigid body parametrization
that allows to group regions of similar motion. 
At the same time, it does not exploit higher-level knowledge such as object classes or pixel-wise semantic information and hence does not require additional supervision rendering
it a flexible but yet highly accurate backbone for many follow-up works, including 
multi-modality \cite{Zhou2023_RAFT3DMultiModal} and cross-scale cost volume
extensions \cite{ScaleRAFT}.

Recent developments in optical flow estimation \cite{MS-RAFT+,CCMR,Xu2023_Unifying} and stereo matching \cite{Lipson2021_RAFTStereo,Li2022_CREStereo,Xu2023_Unifying}, however, show that even for recurrent approaches 
the use of multi-scale concepts can still be beneficial. This includes, amongst others, the design of more sophisticated multi-scale feature and context encoders, the sharing of architectural components over scales, the initialization of finer levels using results from coarse scales, as well as the application of multi-scale losses during training. Even in the context of scene flow estimation from point clouds, which constitutes a somewhat orthogonal research direction to image-based scene flow\footnote{Recently, there have been first approaches to combine image- and point-cloud-based scene flow methods; see \cite{CamLiRAFT,Zhou2023_RAFT3DMultiModal}.
However, those methods also do not combine multi-scale concepts with recurrent-based estimation.}
, the combination of multi-scale ideas and recurrent concepts turned out to be valuable \cite{Cheng2023_MSPointSceneFlow}. Hence, it is surprising that such multi-scale strategies have not been considered in the field of recurrent image-based scene flow estimation so far. Consequently, the goal of this work is to investigate the generalization of these concepts to recurrent image-based scene flow.

\smallskip

\noindent \textbf{Contributions} We propose MS-RAFT-3D -- a hierarchical approach for image-based scene-flow estimation that combines the recurrent single-scale architecture of RAFT-3D with (i) a coarse-to-fine framework that exploits the underlying 
parametrization by initializing finer scales
by upsampling the
SE(3) field and the corresponding motion embeddings,
(ii) a U-Net-based feature encoder with multi-scale feature aggregation
, (iii) a top-down context encoder that outperforms the pre-trained ResNet50 encoder originally used by RAFT-3D while reducing the number of total parameters by more than 17M, (iv) training with a generalized multi-scale multi-iteration loss that guides the estimation of all relevant scene flow components, and (v) the consideration of a recent stereo method that provides highly accurate initial disparities.
While 
ablations show the benefits of the considered architectural concepts, a comparison to the literature demonstrates that our approach sets a new 
state of the art 
(SOTA) in image-based scene flow estimation on the KITTI and Spring benchmark, where it outperforms the previously leading approaches by 8.7\% and 65.8\%, respectively.

\smallskip

\noindent \textbf{Related Work} For the related (sub-)tasks of stereo and optical flow estimation, 
Lipson {\em et al.}~\cite{Lipson2021_RAFTStereo}, Li {\em et al.}~\cite{Li2022_CREStereo} and Jahedi {\em et al.}~\cite{MS-RAFT+,CCMR} 
combine recurrent and multi-scale ideas within a single estimation framework. 
Moreover, 
Xu {\em et al.}~\cite{Xu2023_Unifying} and Pan {\em et al.}~\cite{Pan2024_JointRAFTStereoRAFT} 
seek to improve results for stereo and optical flow by jointly training shared recurrent multi-scale architectures for both tasks (in \cite{Xu2023_Unifying} also with depth). A generalization of these ideas to scene flow, however, has not been investigated in the aforementioned works.

Regarding image-based 
scene flow approaches, on the other hand, the underlying models {\em either} employ a multi-scale strategy \cite{PWOC-3D,CamLiFlow} {\em or} they rely on recurrent updates
\cite{RAFT-3D,CamLiRAFT}.
The only exception constitutes the method of Ling {\em et al.}~\cite{ScaleRAFT} that focuses, however, on {\em cross-scale matching} with a corresponding correlation volume and scale look-up operator instead of considering a hierarchical estimation. Moreover, 
Cheng {\em et al.}~\cite{Cheng2023_MSPointSceneFlow} 
propose to combine multi-scale and recurrent concepts, however, in the context of estimating scene flow from {\em point clouds}.
Finally, it is important to note that all previously mentioned recurrent image-based scene flow approaches \cite{RAFT-3D, CamLiRAFT, ScaleRAFT} work on {\em \textfrac{1}{8} of the original resolution}. Hence, in contrast to our method that operates on up to \textfrac{1}{2} of the resolution, they require much larger upsampling factors which, in turn, impacts the detailedness of their results.

\section{Method}
\label{sec:method}



Given a pair of RGB-D images 
our approach predicts a dense field of SE(3) transformations.
In case of stereo images, \ie when the depth is not available, it is
estimated with a 
stereo method as in 
\cite{RAFT-3D,CamLiFlow,CamLiRAFT,SF2SE3}.
In our approach we consider CroCo-Stereo \cite{CroCo} -- a recent 
highly accurate stereo method.


Conceptually, our approach is based on RAFT-3D \cite{RAFT-3D}, a single-scale approach that groups rigid moving pixels and iteratively refines a scene flow prediction at \textfrac{1}{8} of the original image resolution. Its recurrent network structure is similar to 
RAFT \cite{RAFT} 
for optical flow. First, features are computed for both input images. For all pairs of features from the first and second image, a similarity score is computed and stored in a 4D cost volume. Finally, a recurrent network block then iteratively produces rigid motion embeddings to group rigid moving objects and a revision map for the scene flow.

\begin{figure}
    \centering
    \resizebox{\columnwidth}{!}{\input{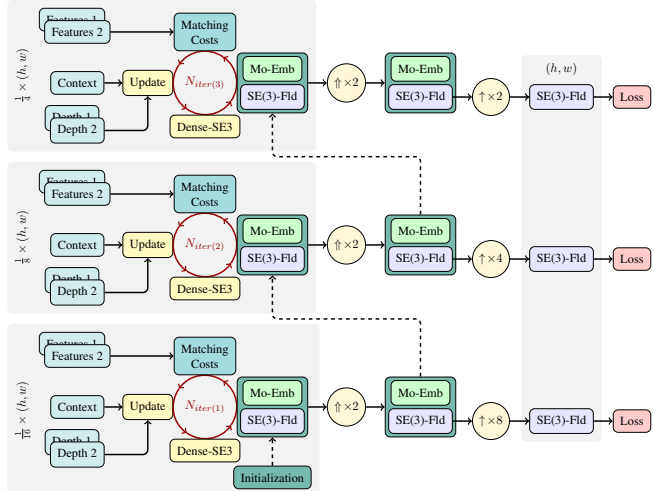}}
    \vspace{-6mm}
    \caption{Architecture of MS-RAFT-3D.
    }
    \label{fig:architecture}
    \vspace{-2mm}
\end{figure}

In contrast,
we propose a multi-scale estimation scheme to obtain scene flow. As shown in \Cref{fig:architecture}, we perform coarse-to-fine initializations for the rigid motion embeddings and the SE(3) motion field.
Starting at the coarsest scale with a resolution of \textfrac{1}{16}, we initialize the SE(3) motion field with the identity transformation and the rigid motion embeddings with 0.
After multiple refinement iterations on the coarsest scale, the SE(3) motion field and the rigid motion embeddings are upsampled with a learned $\times 2$ convex upsampling mask \cite{RAFT} and used to initialize the next finer scale until the matching is performed at the final scale of \textfrac{1}{4}. 
Because the edges of the motion field and the rigid motion embeddings are expected to coincide, the upsampling mask is shared between both.
As in \cite{RAFT-3D}, upsampling of the SE(3) field is done in the Lie algebra.
Note that while we emphasize here on the 3-scale model, we also present a 4-scale variant (finest scale at $\frac{1}{2}$) in \Cref{sec:results}. 


\noindent\textbf{Encoders} 
The matching process starts by extracting image features from both frames and context features from the first frame.
For computing the image features we use a Siamese U-Net-based encoder as in \cite{CCMR}.
While the feature consolidation in this type of encoder is beneficial for 
image features, based on our ablations, it is not helpful for context features.
Hence, as context encoder, we use a simpler top-down extractor
resulting 
in an improved performance and, at the same time, 
a much leaner design compared to 
RAFT-3D's ResNet50 \cite{He2016_ResNet} encoder.
Using these two encoders, image and context features are output at \textfrac{1}{16}, \textfrac{1}{8}, and \textfrac{1}{4} 
resolutions.
Since context features are 
inputs of the shared (across scales) recurrent unit (in the update block in \Cref{fig:architecture}), a $1\times1$ \textit{conv} ensures that the number of context channels coincides for all scales.

\noindent\textbf{Cost Volume} 
Using the image features, we construct a 4D cost volume containing matching costs for all pairs of features from the first and the second image on all flow prediction resolutions. The cost volume in each scale is pooled twice over the last two dimensions, where the cost entries are queried based on the current estimate.
Note that the number of levels in the correlation pyramid is reduced to 2 due to the lower resolution on the coarsest scale compared to RAFT-3D; {\em cf.} 
\cite{MS-RAFT+}.
We also employ the on-demand cost-sampling \cite{MS-RAFT+} to enable training and inference on the high-resolution images. In this case, the full 4D cost volume is not pre-computed, but the matching scores are computed in each refinement step.
\noindent\textbf{Flow Refinement} 
At each scale of our hierarchical approach, the flow refinement follows the same procedure as RAFT-3D \cite{RAFT-3D}.
In this context, the refinement is partially performed via the update module (see \Cref{fig:architecture}), where a GRU unit is used to iteratively refine the SE(3) field as well as the rigid motion embeddings. Note that each scale's context features are split over the channels into two parts: one of them initializes the hidden-state of the current scale (Con-1), while the other is used in the flow refinement of that scale (Con-2).\linebreak
During each flow refinement iteration, the matching costs for a local neighborhood around the current flow are considered. 
Costs, context features (Con-2), 
current 2D flow, inverse depth residual (\ie difference of projected inverse depth maps), the current SE(3) field along with the current rigid motion embeddings are passed through the recurrent unit and a refinement step is performed. The hidden-state of the GRU cell is then updated which is used to generate flow revisions (\ie corrections of the flow),
confidence weights, rigid motion embeddings and inverse edge weights.
As in \cite{RAFT-3D}, we use the bi-Laplacian smoothing
to smooth the rigid motion embeddings $\v$ only at
positions where the inverse edge weights $\mathbf w_x,\mathbf w_y$ are large (\ie within objects) by solving
\[
    \u^\ast = \argmin_{\u} \left\{ 
    \| \u - \v \|^2 + 
    \| D_x \u \|_{\mathbf w_x}^2 + \| D_y \u \|_{\mathbf w_y}^2 
    \right\}
\]
where $D_x$ and $D_y$ are first-order finite difference operators.
Afterwards, the Dense-SE3 layer (see \Cref{fig:architecture}) updates the SE(3) transformation for each pixel by computing that transformation which fits the motion of all pixels in the neighborhood that have a similar rigid motion embedding and a high confidence weight. 
This is achieved 
by solving
\[
    \T^\ast \!\!\!=\! \argmin_{\T}\!\! \sum_{i \in \Omega}\! \sum_{j \in \mathcal N_i}\!\! \sigma(-\|\u_i^\ast - \u_j^\ast\|^2) \cdot \| \pi(\T_i \cdot \pi^{\!-1}\!(\x_j)) - \mathbf t_j \|_{\mathbf c_j}^2
\]
with the SE(3) transformation field $\T$, the image domain $\Omega$, the neighborhood $\mathcal N_i$ around pixel $i$, the pixel location $\x_j$, the sigmoid function $\sigma$, the camera projection $\pi$ that contains the inverse depth, the revised target position $\mathbf t$ and the confidence weights $\mathbf c$.
This equation is solved by a damped Gauss-Newton solver that only takes one step per flow refinement iteration.
In our multi-scale approach, we use a fixed neighborhood radius of 32 in the Dense-SE3 layer regardless of the scale, \ie, the 
considered area gets smaller on finer scales.



\noindent\textbf{Supervision} 
We generalize the idea of the multi-scale multi-iteration loss from 
\cite{MS-RAFT+} to the scene flow loss of RAFT-3D \cite{RAFT-3D}.
More precisely, after each flow refinement iteration, we upsample the flow revision predicted by the update network and the SE(3) motion field first with the convex upsampling mask to twice the resolution and then with bilinear upsampling to the full input resolution.
For the upsampled SE(3) field $\T$, the induced optical flow $\f_{\rm est}$ and the induced inverse depth change $\d_{\rm est}$ are calculated as
\[
    (\f_{\rm est}, \d_{\rm est})^\top = \pi(\T \cdot \pi^{\!-1}(\x)) - \x \; .
\]

The loss for 
iteration $i$ at scale $s$
is then given by
\[
    \L_{s,i} = \| \f_{\rm est}^{s,i} - \f_{\rm gt} \|_1 + w_d \cdot \| \d_{\rm est}^{s,i} - \d_{\rm gt} \|_1 + w_{\rm rev} \cdot \| \f_{\rm rev}^{s,i} - \f_{\rm gt} \|_1
\]
with the optical flow portion $\f_{\rm rev}$ of the revision predicted by the GRU, the optical flow ground truth $\f_{\rm gt}$, the inverse depth change ground truth $\d_{\rm gt}$ and the weights $w_{d}=250$, $w_{\rm rev}=0.2$.
Finally, the overall multi-scale multi-iteration loss reads
\[
    \L = \sum_{s=1}^{N_{\rm scales}\vphantom{(}} \sum_{i=1}^{N_{{\rm iter}(s)}} \gamma^{R_{s,i}} \L_{s,i}
\]
with 
decay factor
$\gamma=0.8$ and 
remaining iterations
$R_{s,i} = N_{{\rm iter}(s)}- i + \sum_{k=s+1}^{N_{\rm scales}} N_{{\rm iter}(k)}$
after iteration $i$ at scale $s$.

%
%

\section{Results and Experiments}
\label{sec:results}
\nocite{Mayer2016_SceneFlow,Spring}

\begin{table}[b!]
    \centering
    \scalebox{0.85}{
    \begin{tabular}{lccccccccc}
        \toprule

        & \multicolumn{2}{c}{Optical Flow} & \multicolumn{3}{c}{Scene Flow}\\
        \cmidrule(lr){2-3} \cmidrule(lr){4-6}
        Model                     &                1px &                EPE &                0.05 &               0.1 &                EPE\\
        \midrule
        RAFT-3D \cite{RAFT-3D}                  &             13.6x &              2.46 &             12.2x &               8.5x &             0.062x\\
        \midrule
        RAFT-3D-CroCo                  &             12.83 &              2.08 &             12.93 &               9.15 &             0.0546\\        
        MS-RAFT-3D                &  \phantom{0}\underline{8.47} &  \underline{1.74} & \underline{10.01} &  \underline{7.17} & \underline{0.0442}\\
        MS-RAFT-3D+               &     \phantom{0}\textbf{7.52} &     \textbf{1.65} &     \phantom{0}\textbf{7.72} &     \textbf{5.59} &    \textbf{0.0365}\\
        \bottomrule
    \end{tabular}
    }
    \caption{Comparison between MS-RAFT-3D(+) and RAFT-3D on FT for the evaluation
    split from \cite{RAFT-3D} 
    using ground truth disparities. 
    Inlier metrics reported in \cite{RAFT-3D} have been converted.
    {\bf Bold} = best result, \underline{underline} = second best result.    
    }
    \label{tab:results_fth}
\end{table}

\begin{table*}
    \centering
    \scalebox{0.85}{
    \begin{tabular}{l ccc ccc ccc ccc}
        \toprule
        & \multicolumn{3}{c}{Disparity 1} & \multicolumn{3}{c}{Disparity 2} & \multicolumn{3}{c}{Optical Flow} & \multicolumn{3}{c}{Scene Flow} \\
        \cmidrule(lr){2-4} \cmidrule(lr){5-7} \cmidrule(lr){8-10} \cmidrule(lr){11-13}
        Method                             &   bg &   fg &  all &   bg &   fg &  all &   bg &   fg &  all &   bg &    fg &  all\\
        \midrule
        RAFT-3D       \cite{RAFT-3D}       & 1.48 & 3.46 & 1.81 & 2.51 & 9.46 & 3.67 & 3.39 & 8.79 & 4.29 & 4.27 & 13.27 & 5.77\\
        SF2SE3        \cite{SF2SE3}        & \underline{1.40} & \underline{2.91} & \underline{1.65} & 2.20 & 7.66 & 3.11 & 3.17 & 8.79 & 4.11 & 3.75 & 13.15 & 5.32\\
        RigidMask+ISF \cite{RigidMask_ISF} & 1.53 & 3.65 & 1.89 & 2.09 & 8.92 & 3.23 & 2.63 & 7.85 & 3.50 & 3.25 & 13.08 & 4.89\\
        CamLiFlow     \cite{CamLiFlow}     & 1.48 & 3.46 & 1.81 & 1.92 & 8.14 & 2.95 & 2.31 & 7.04 & 3.10 & 2.87 & 12.23 & 4.43\\
        CamLiRAFT     \cite{CamLiRAFT}     & 1.48 & 3.46 & 1.81 & \underline{1.91} & 8.11 & 2.94 & \textbf{2.08} & 7.37 & \underline{2.96} & \textbf{2.68} & 12.16 & 4.26\\
        \midrule
        RAFT-3D-CroCo                    & \textbf{1.38} & \textbf{2.65} & \textbf{1.59} & 2.15 & 7.20 & 2.99 & 2.91 & 7.41 & 3.66 & 3.43 & 10.94 & 4.68\\
        MS-RAFT-3D                         & \textbf{1.38} & \textbf{2.65} & \textbf{1.59} & 1.99 & \underline{6.17} & \underline{2.68} & 2.43 & \textbf{5.75} & 2.98 & 2.95 &  \phantom{0}\textbf{9.47} & \underline{4.04}\\
        MS-RAFT-3D+                        & \textbf{1.38} & \textbf{2.65} & \textbf{1.59} & \textbf{1.81} & \textbf{5.50} & \textbf{2.42} & \underline{2.22} & \underline{5.99} & \textbf{2.85} & \underline{2.75} & \phantom{0}\underline{9.59} & \textbf{3.89}\\
        \bottomrule
    \end{tabular}
    }
    \caption{Best ranking published methods on the KITTI benchmark. 
    }
    \label{tab:results_kitti}
\end{table*}

\begin{table*}
   \centering
   \scalebox{0.85}{
    \begin{tabular}{l ccc ccc ccc cc}
        \toprule
       & \multicolumn{3}{c}{Disparity 1} & \multicolumn{3}{c}{Disparity 2} & \multicolumn{3}{c}{Optical Flow} & \multicolumn{2}{c}{Scene Flow}\\
        \cmidrule(lr){2-4} \cmidrule(lr){5-7} \cmidrule(lr){8-10} \cmidrule(lr){11-12}
        Model                     &                1px &                 D1 &                EPE &                1px &                 D2 &                EPE &                1px &                 Fl &                EPE &                1px &                 SF\\
        \midrule

        PWOC-3D \cite{PWOC-3D}            & \underline{18.23} &  \underline{5.89} &  \underline{1.340} &             18.98 &              6.19 &              1.393 &             14.46 &              5.41 &              2.813 &             26.71 &              9.99\\
        \midrule
        RAFT-3D-CroCo             &     \phantom{0}\textbf{7.14} &     \textbf{2.71} &     \textbf{0.471} &     \phantom{0}{7.36} &     {2.75} &     {0.481} &     \phantom{0}{4.12} &     {1.56} &     {0.454} &    \phantom{0}{9.70} &     {3.59}\\
       MS-RAFT-3D             &     \phantom{0}\textbf{7.14} &     \textbf{2.71} &     \textbf{0.471} &     \phantom{0}\underline{7.29} &     \underline{2.74} &     \underline{0.480} &     \phantom{0}\underline{3.70} &     \textbf{1.41} &     \underline{0.377} &    \phantom{0}\underline{9.28} &     \textbf{3.43}\\
        MS-RAFT-3D+            &     \phantom{0}\textbf{7.14} &     \textbf{2.71} &     \textbf{0.471} &     \phantom{0}\textbf{7.24} &     \textbf{2.73} &     \textbf{0.477} &            \phantom{0}\textbf{3.52} &            \underline{1.43} &            \textbf{0.359} &           \phantom{0}\textbf{9.13} &            \underline{3.47}\\
        \bottomrule
    \end{tabular}
    }
    
    \caption{
    Best ranking published methods on the Spring benchmark.
    }
 \label{tab:results_spring}
\end{table*}

\noindent\textbf{Models}
We investigate a 3-scale model 
(MS-RAFT-3D), operating on \textfrac{1}{16}, \textfrac{1}{8} and \textfrac{1}{4} of the original image resolution and a 4-scale variant
(MS-RAFT-3D\textbf{+}) that adds a scale with resolution \textfrac{1}{2} (see architecture in \textit{supplementary material}: 
\Cref{sec:ms_raft_3d+_architecture}). 
For the latter, we increase the Dense-SE3 neighborhood radius to 64 due to the higher resolution.
To ensure numerical stability, in the presence of rounding errors, 
we increase the damping factor of the Gauss-Newton solver.
Other architectural choices are based on the ablations of the 3-scale variant.
 Similar to \cite{MS-RAFT+}, we use [4, 6, 8] recurrent flow update iterations for MS-RAFT-3D and [4, 5, 5, 6] iterations for MS-RAFT-3D+ during training and inference.
Our models have 27.5M and 27.9M parameters in total, of which 20.7M and 20.9M are the context encoder.
Compared to RAFT-3D \cite{RAFT-3D} which has 44.8M parameters,
we need around 17M parameters less.

\noindent\textbf{Training} 
Our network is pre-trained on FlyingThings (FT) \cite{Mayer2016_SceneFlow} for 200k iterations with a batch size of 3 using $368 \times 768$ patches and a maximum learning rate of $2 \cdot 10^{-4}$.
We then proceed with fine-tuning on KITTI \cite{Menze2015_KITTI} or Spring \cite{Spring}.
During fine-tuning, we train for 30k iterations with a batch size of 4 and a maximum learning rate of $6.25 \cdot 10^{-6}$.
Patch sizes for KITTI and Spring are $256 \times 480$ and $640 \times 1248$, respectively.
Importantly, following our baseline RAFT-3D \cite{RAFT-3D}, we use the \textit{ground truth disparity} for FT and \textit{estimated disparities} on KITTI and Spring.

\subsection{Results} 
\noindent \textbf{FlyingThings} First, using ground truth disparities, we compare our two multi-scale models on the FT split from \cite{RAFT-3D} to the original single-scale RAFT-3D approach as well as to a model we trained ourselves with the improved code provided by the RAFT-3D authors\footnote{Available at \texttt{https://github.com/princeton-vl/RAFT-3D}}.
This model, which we call RAFT-3D-CroCo, is fine-tuned 
and evaluated \textit{in later experiments with disparities from CroCo-Stereo} (as our multi-scale methods) and hence serves as second baseline in addition to RAFT-3D. \Cref{tab:results_fth} reports the corresponding endpoint errors (EPE) as well as the outlier 
metrics 1px, 0.05 and 0.1, \ie the percentage of pixels that are not within 1 pixel of the optical flow ground truth or within 0.05 or 0.1 units of the scene flow ground truth.
Our models outperform RAFT-3D and RAFT-3D-CroCo (the latter with better results due to the improved code) across all metrics with improvements in the range from 15.6\% to 41.4\%.
Moreover, the 4-scale version has large gains over the 3-scale version in the scene flow metrics.
Note that these improvements are \textit{solely} due to the multi-scale architecture and \textit{not} due to the improved stereo method, as \textit{ground truth} disparity is used in this experiment.

\begin{figure*}
\centering{
\setlength\tabcolsep{1pt}
\newcommand*\imgtrimtop{3cm}
\tikzset{labelstyle/.style={anchor=north west, fill=black, inner sep=1, text opacity=1, fill opacity=0.7, scale=0.5, xshift=3, yshift=-3, text=white}}
\tikzset{imgstyle/.style={inner sep=0,anchor=south west}}
\tikzset{rectstyle/.style={draw=black,densely dotted}}
\tikzset{rectstyle2/.style={draw=white, densely dotted}}
\newcommand{\drawlabel}[1]{%
    \draw (img.north west) node[labelstyle] {#1};
}
\newcommand{\drawrect}{
    \coordinate (A1) at (1.4,0.4);
    \coordinate (A2) at (1.75,0.65);

    \coordinate (B1) at (0.1,0.1);
    \coordinate (B2) at (0.8,0.6);
    \draw[rectstyle2] (B1) rectangle (B2); 
}

\newcommand{\croppedincludegraphics}[2][]{%
    \includegraphics[trim=0 9.0cm 0 3.8cm, clip, #1]{#2}%
}

\begin{tabular}{ccccc}
 target disparity & \textit{D2} error & optical flow & \textit{Fl} error& \textit{SF} error\\
\begin{tikzpicture}
\draw (0, 0) node[imgstyle] (img) {
\includegraphics[width=0.195\textwidth]{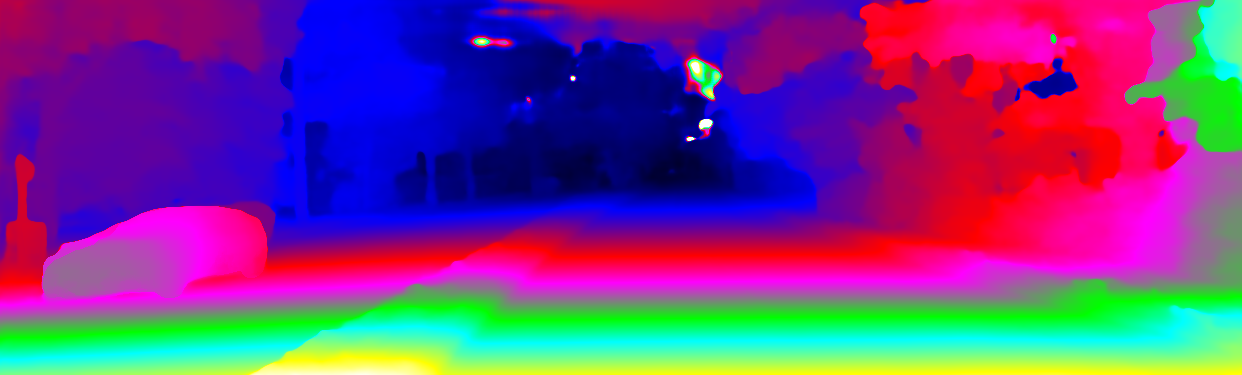}};
\drawrect
\drawlabel{CamLiRAFT [11]}
\end{tikzpicture} &
\begin{tikzpicture}
\draw (0, 0) node[imgstyle] (img) {
\includegraphics[width=0.195\textwidth]{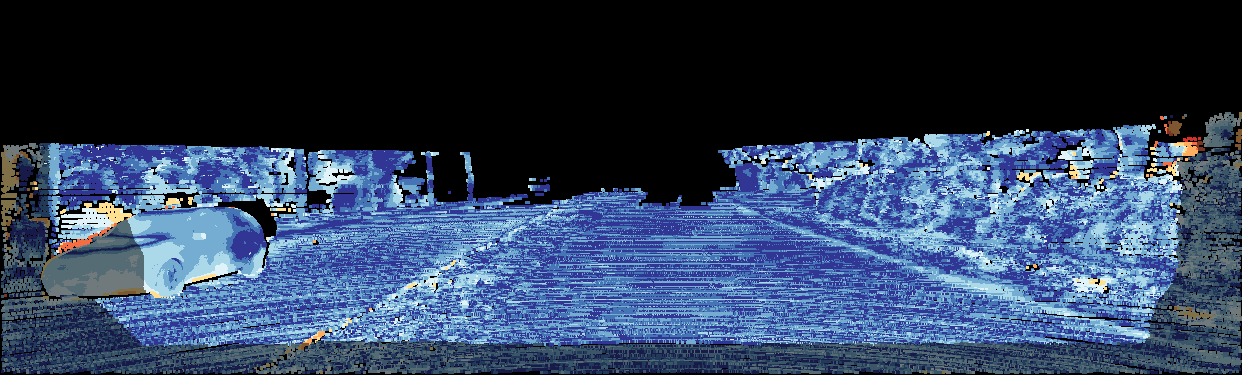}};
\drawrect
\drawlabel{D2-fg: 1.90, D2-bg: 0.85, D2-all: 0.98}
\end{tikzpicture} &
\begin{tikzpicture}
\draw (0, 0) node[imgstyle] (img) {
\includegraphics[width=0.195\textwidth]{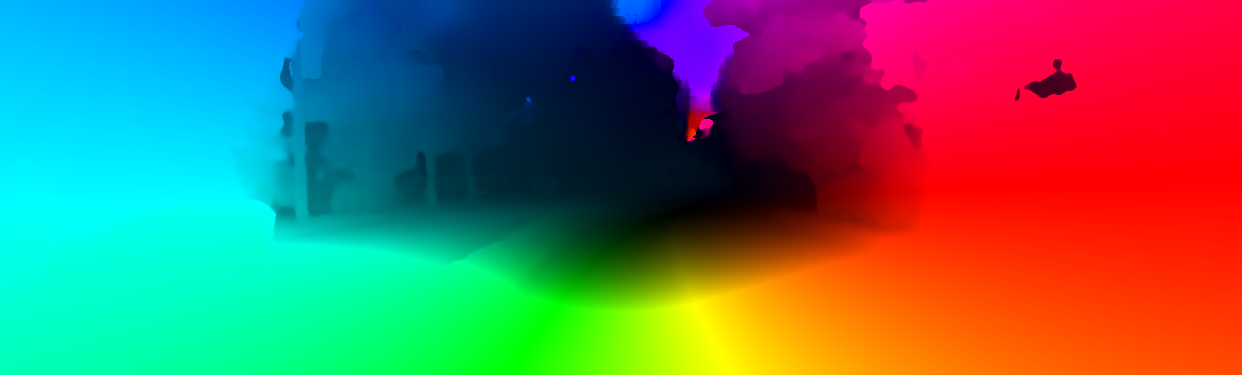}};
\drawrect
\end{tikzpicture} &
\begin{tikzpicture}
\draw (0, 0) node[imgstyle] (img) {
\includegraphics[width=0.195\textwidth]{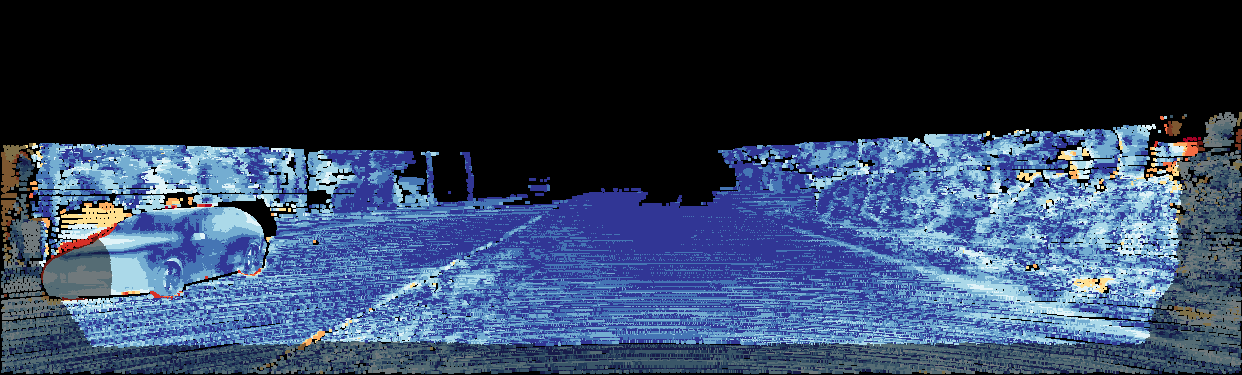}};
\drawrect
\drawlabel{Fl-fg: 2.36, Fl-bg: 1.43, Fl-all: 1.54}
\end{tikzpicture} &
\begin{tikzpicture}
\draw (0, 0) node[imgstyle] (img) {
\includegraphics[width=0.195\textwidth]{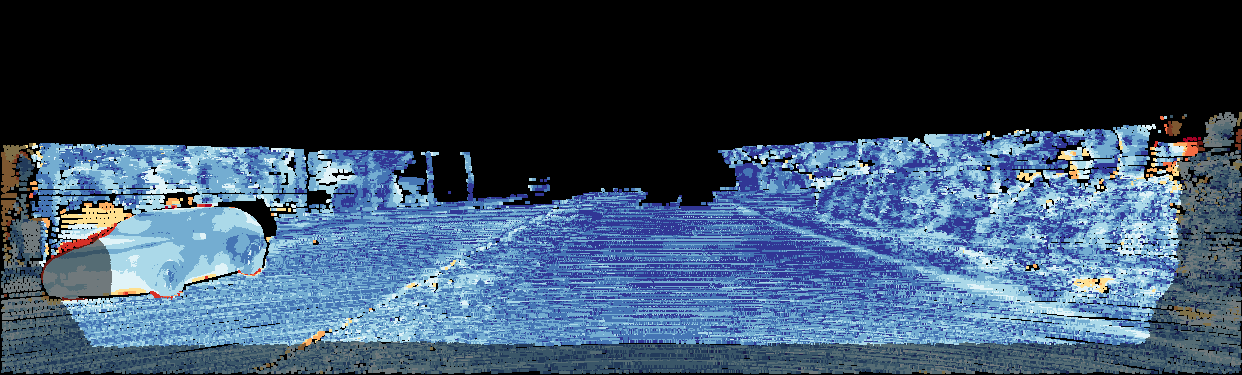}};
\drawrect
\drawlabel{SF-fg: 3.22, SF-bg: 1.45, SF-all: 1.67}
\end{tikzpicture}
\\[-0.6mm]
\begin{tikzpicture}
\draw (0, 0) node[imgstyle] (img) {
\includegraphics[width=0.195\textwidth]{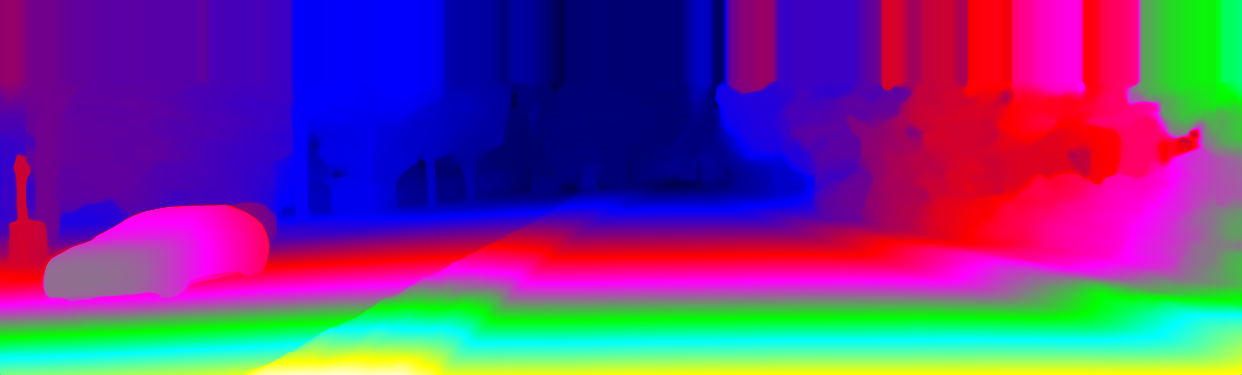}};
\drawrect
\drawlabel{MS-RAFT-3D+}
\end{tikzpicture} &
\begin{tikzpicture}
\draw (0, 0) node[imgstyle] (img) {
\includegraphics[width=0.195\textwidth]{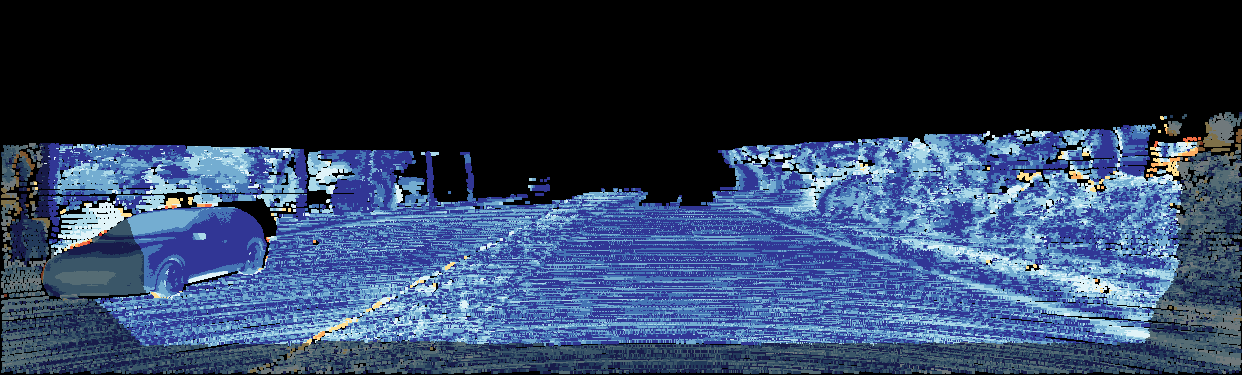}};
\drawrect
\drawlabel{D2-fg: 0.47, D2-bg: 0.64, D2-all: 0.62}
\end{tikzpicture} &
\begin{tikzpicture}
\draw (0, 0) node[imgstyle] (img) {
\includegraphics[width=0.195\textwidth]{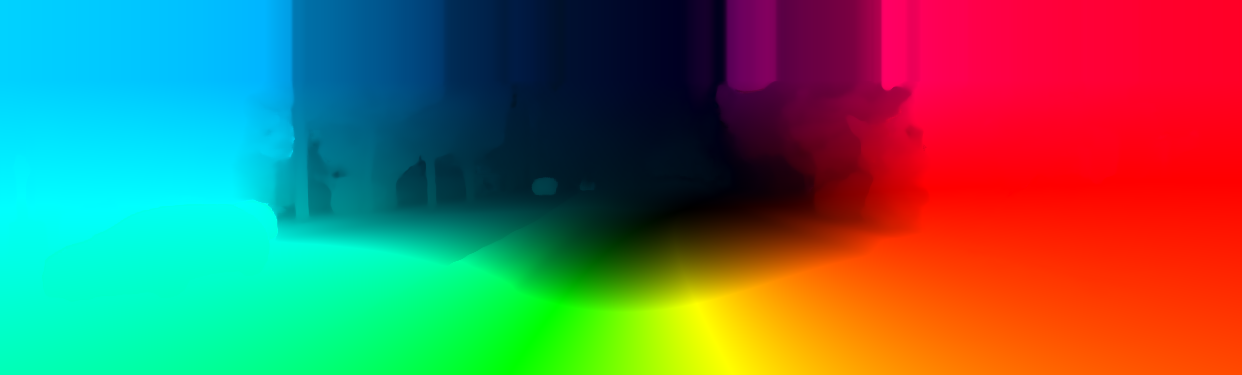}};
\drawrect
\end{tikzpicture} &
\begin{tikzpicture}
\draw (0, 0) node[imgstyle] (img) {
\includegraphics[width=0.195\textwidth]{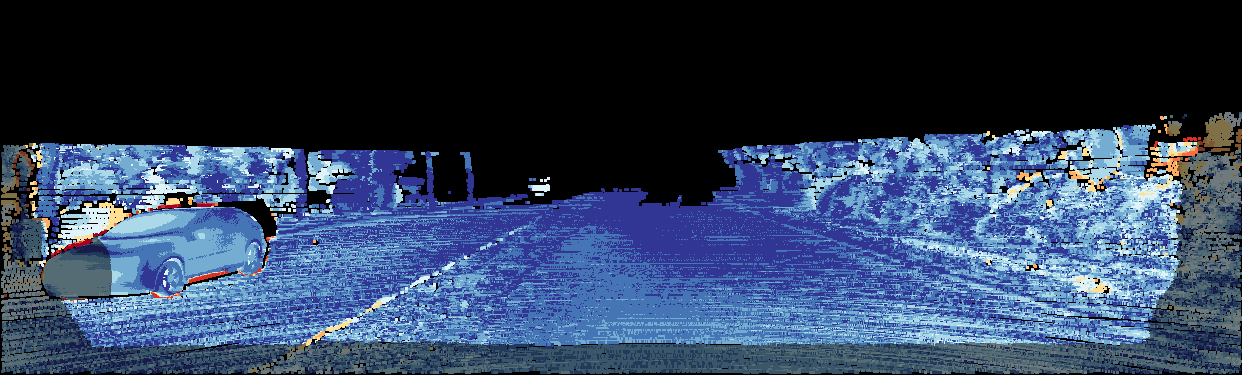}};
\drawrect
\drawlabel{Fl-fg: 1.27, Fl-bg: 1.34, Fl-all: 1.34}
\end{tikzpicture} &
\begin{tikzpicture}
\draw (0, 0) node[imgstyle] (img) {
\includegraphics[width=0.195\textwidth]{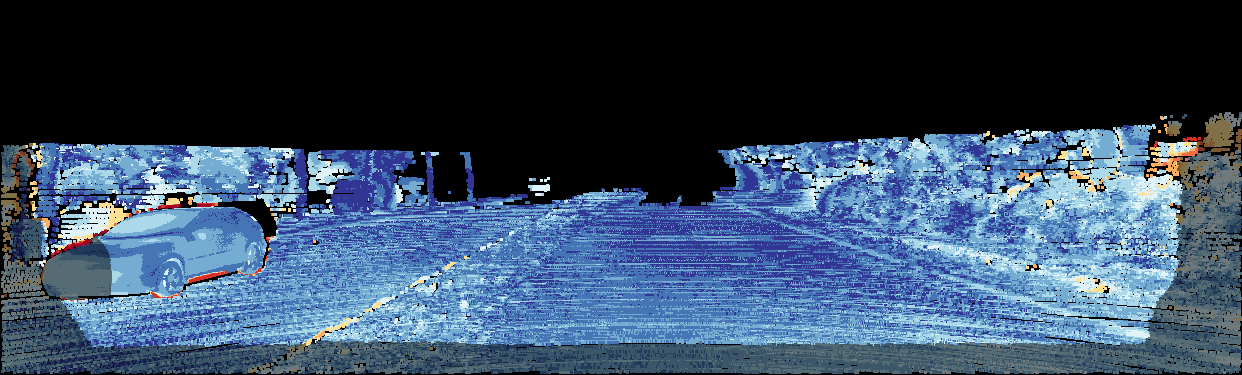}};
\drawrect
\drawlabel{SF-fg: 1.27, SF-bg: 1.38, SF-all: 1.37}
\end{tikzpicture}
\\[-0.6mm]
\begin{tikzpicture}
\draw (0, 0) node[imgstyle] (img) {
\includegraphics[width=0.195\textwidth]{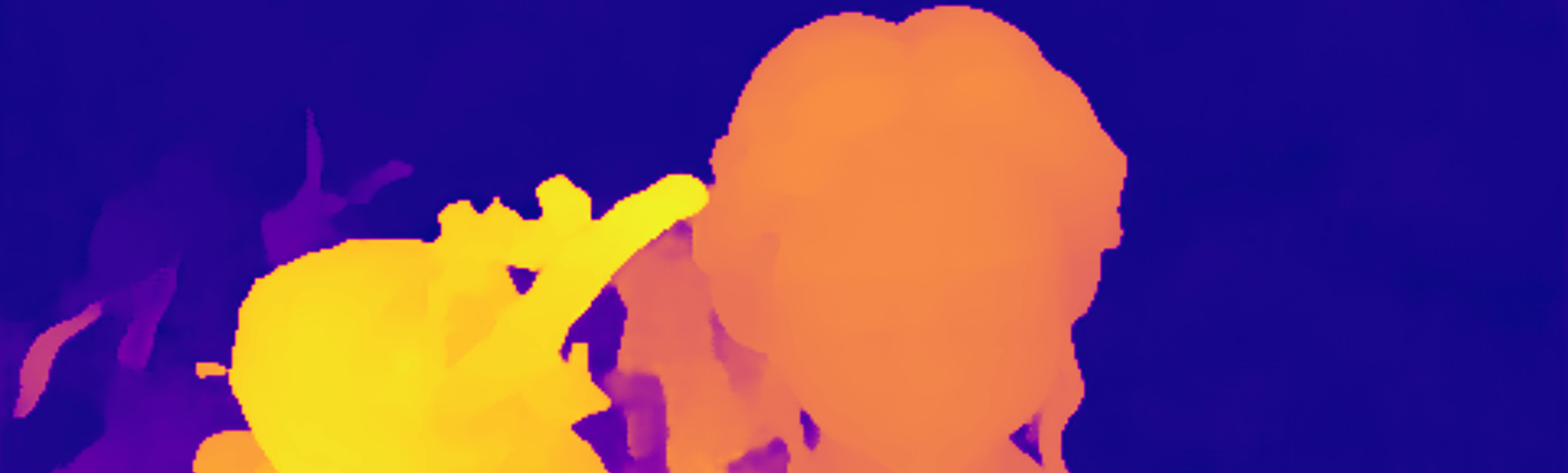}};
\drawlabel{PWOC-3D [7]}
\end{tikzpicture} &
\begin{tikzpicture}
\draw (0, 0) node[imgstyle] (img) {
\includegraphics[width=0.195\textwidth]{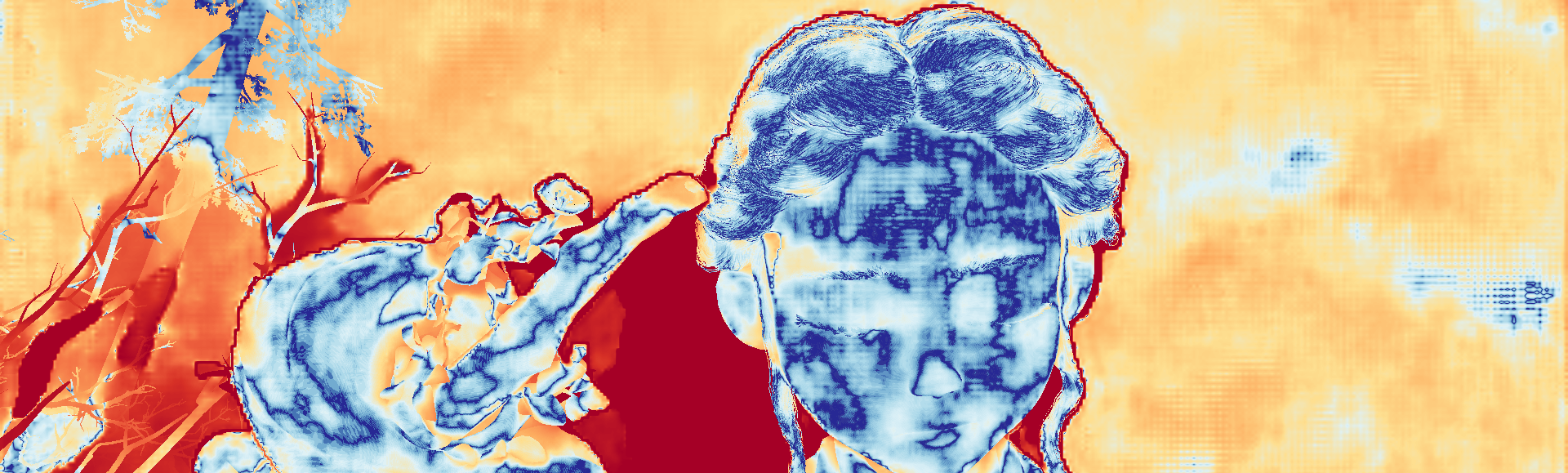}};
\end{tikzpicture} &
\begin{tikzpicture}
\draw (0, 0) node[imgstyle] (img) {
\includegraphics[width=0.195\textwidth]{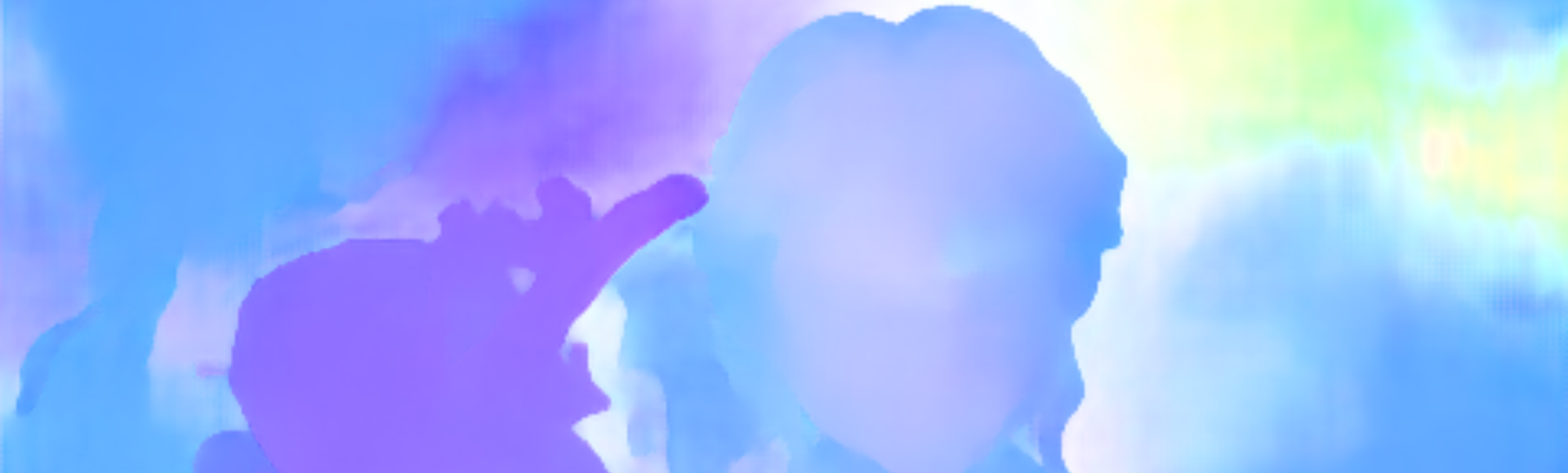}};
\end{tikzpicture} &
\begin{tikzpicture}
\draw (0, 0) node[imgstyle] (img) {
\includegraphics[width=0.195\textwidth]{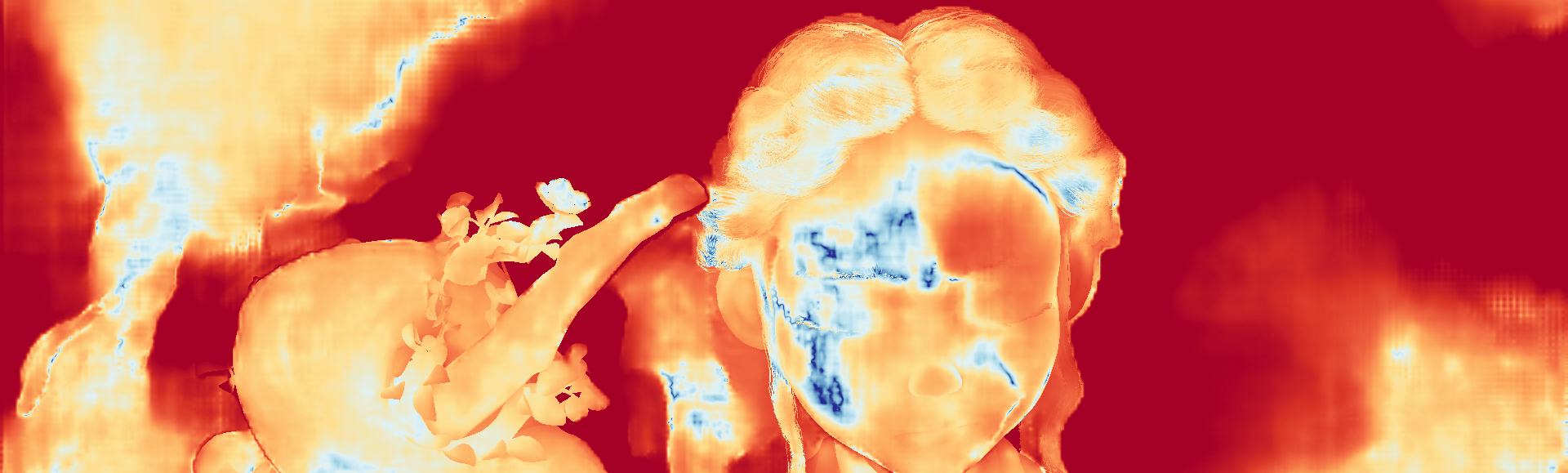}};
\end{tikzpicture} &
\begin{tikzpicture}
\draw (0, 0) node[imgstyle] (img) {
\includegraphics[width=0.195\textwidth]{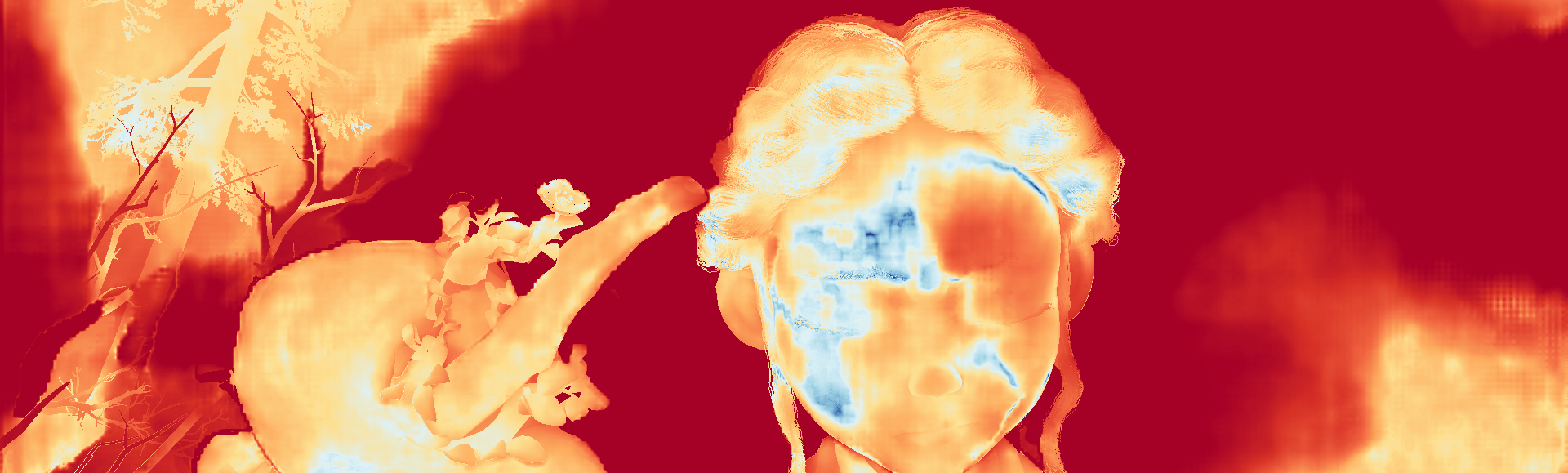}};
\end{tikzpicture}

\\[-0.6mm]
\begin{tikzpicture}
\draw (0, 0) node[imgstyle] (img) {
\includegraphics[width=0.195\textwidth]{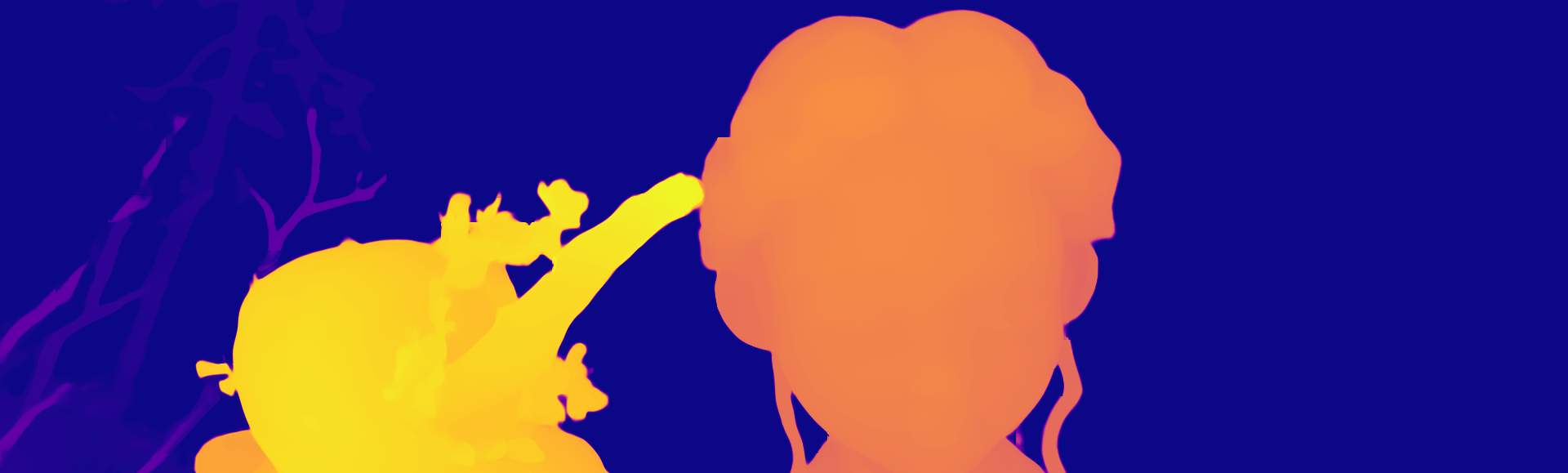}};
\drawlabel{MS-RAFT-3D}
\end{tikzpicture} &
\begin{tikzpicture}
\draw (0, 0) node[imgstyle] (img) {
\includegraphics[width=0.195\textwidth]{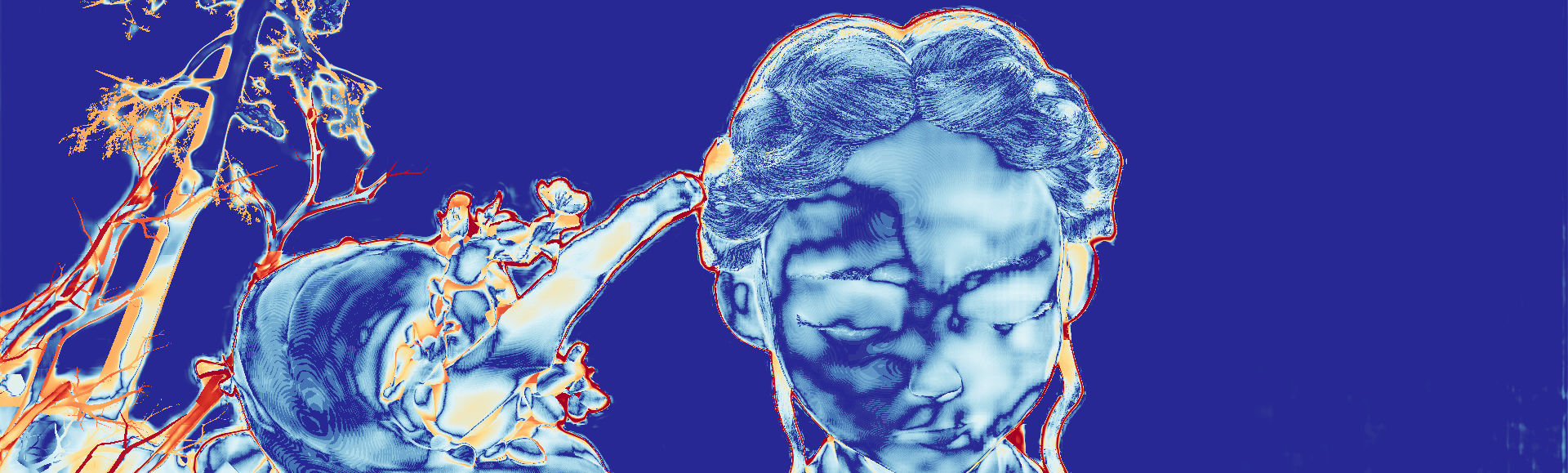}};
\end{tikzpicture} &
\begin{tikzpicture}
\draw (0, 0) node[imgstyle] (img) {
\includegraphics[width=0.195\textwidth]{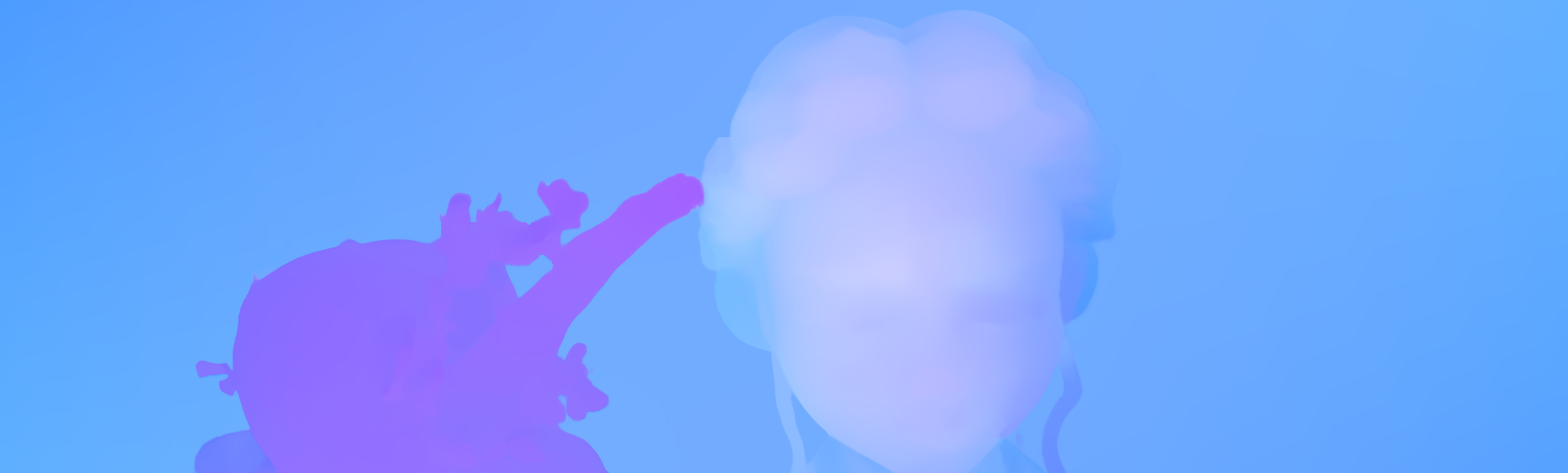}};
\end{tikzpicture} &
\begin{tikzpicture}
\draw (0, 0) node[imgstyle] (img) {
\includegraphics[width=0.195\textwidth]{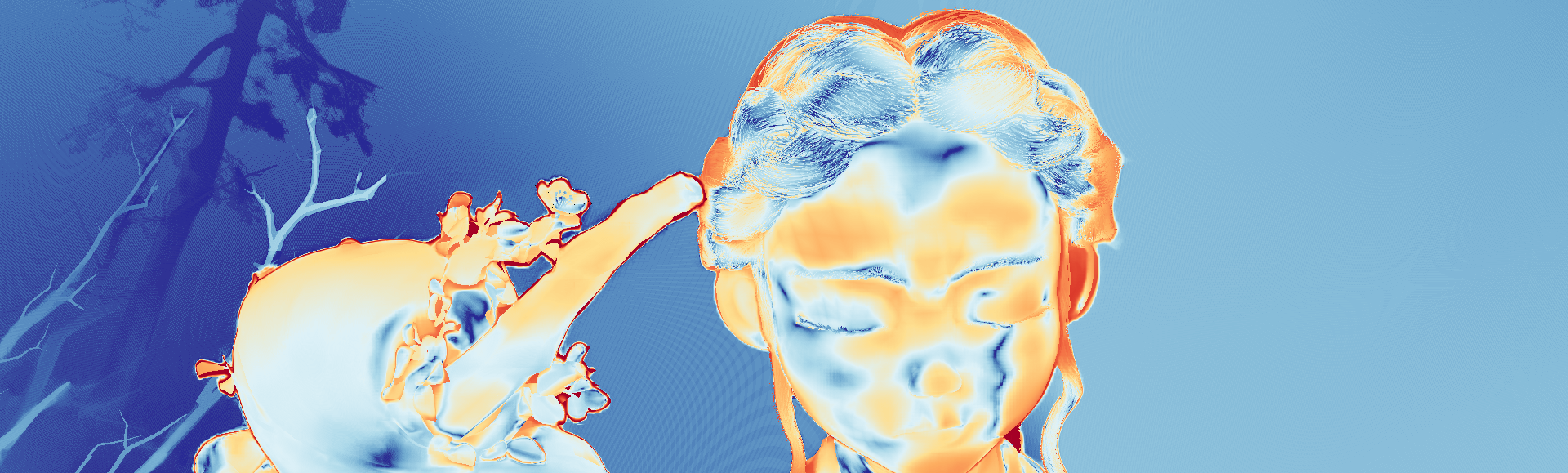}};
\end{tikzpicture} &
\begin{tikzpicture}
\draw (0, 0) node[imgstyle] (img) {
\includegraphics[width=0.195\textwidth]{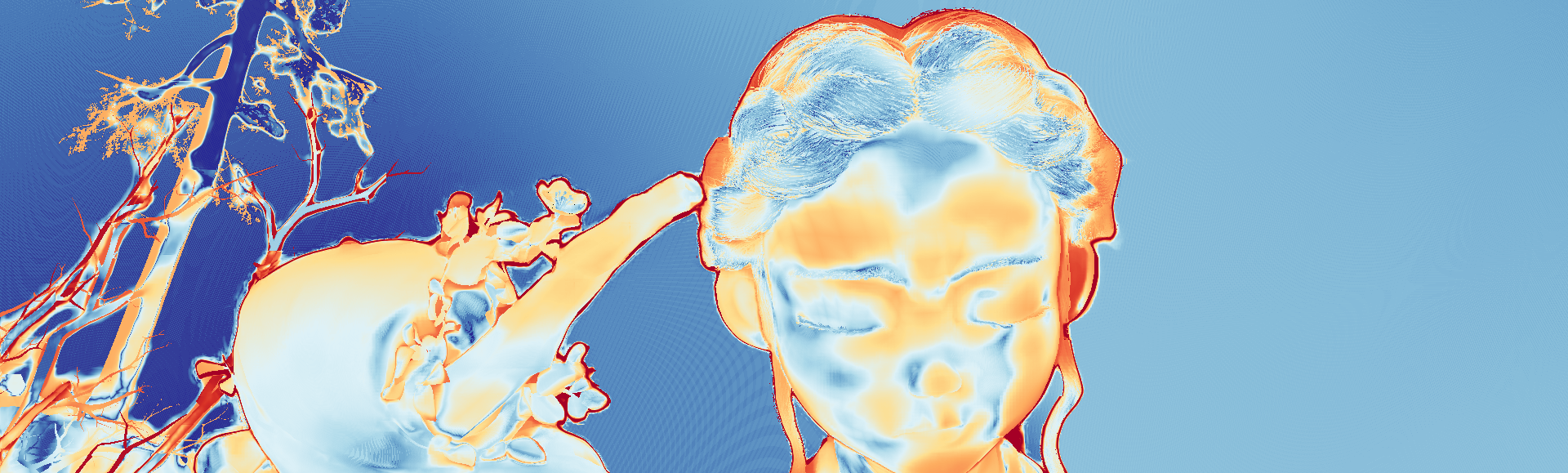}};
\end{tikzpicture}
\end{tabular}
}
\vspace{-2mm}
\caption{Visual comparison of our results to the current SOTA on KITTI (\textit{top}) and Spring (\textit{bottom}). 
}
\label{fig:visual_comparison}
\end{figure*}

\noindent \textbf{KITTI and Spring}
Furthermore, we submitted our results to the KITTI and Spring benchmark to compare it to the current SOTA; see \Cref{tab:results_kitti,tab:results_spring}. 
On KITTI, our multi-scale methods consistently outperform 
both baselines reducing the scene flow outliers up to 13.7\% for 
MS-RAFT-3D
and up to 16.9\% for 
MS-RAFT-3D+.
They also perform favorably against other published methods with improvements of 8.7\% compared to the current SOTA (3.89 vs.\ 4.26).
Our approaches are particularly accurate in the dynamic foreground (fg) regions, which contain independently moving objects.
This is confirmed by the visualizations in \Cref{fig:visual_comparison}.

On Spring, our 
multi-scale models perform once again better than the RAFT-3D-CroCo baseline in all metrics.
Note that, since other methods on the benchmark including the original RAFT-3D method, are not fine-tuned on the dataset, we refrain from showing them and only compare to the best fine-tuned approach: PWOC-3D. 
Also in this case our multi-scale models set a new SOTA with 65.8\% improvements (9.13 vs.\ 26.71). This also becomes explicit from \Cref{fig:visual_comparison}.

\begin{table}[t!]
    \centering
    \scalebox{0.85}{
    \begin{tabular}{lccc}
        \toprule
         {\em Ablations 1/2}               &               D2 &               Fl & SF\\
        \midrule
        \multicolumn{3}{l}{\textbf{Feature Encoder}} \\
        \midrule
        Top-down RAFT-like        & \underline{4.18} & \phantom{0}\underline{8.70} & \phantom{0}\underline{9.05} \\
        {\bf CCMR}             &    \textbf{2.87} &    \phantom{0}\textbf{6.23} & \phantom{0}\textbf{6.66}    \\
        \midrule
        \multicolumn{3}{l}{\textbf{Context Encoder}} \\
        \midrule
        {\bf Top-down RAFT-like}         &    \textbf{2.87} &    \phantom{0}\textbf{6.23} & \phantom{0}\textbf{6.66}    \\
        CCMR             & \underline{2.98} & \phantom{0}\underline{6.59} & \phantom{0}\underline{7.03} \\
        RAFT-FPN         &           {3.24} &           \phantom{0}{7.32} & \phantom{0}7.74             \\
        ResNet without Enhancement          &           {3.47} &           \phantom{0}{8.21} & \phantom{0}8.66             \\
        ResNet with Enhancement       &           {3.62} &           \phantom{0}{8.07} & \phantom{0}8.45             \\
        \midrule
        \multicolumn{3}{l}{\textbf{Scale Initialization}} \\
        \midrule
        Re-Init HS, Re-Init Mo-Emb       & \underline{3.13} &           \phantom{0}{7.32} & \phantom{0}7.74             \\
        {\bf Re-Init HS, Upsample Mo-Emb}      &    \textbf{2.87} &    \phantom{0}\textbf{6.23} & \phantom{0}\textbf{6.66}    \\
        Upsample HS, Re-Init Mo-Emb      &           {3.69} &           \phantom{0}{9.33} & \phantom{0}9.78             \\
        Upsample HS, Upsample Mo-Emb     &           {3.39} & \phantom{0}\underline{6.75} & \phantom{0}\underline{7.23} \\
        \midrule
        \multicolumn{3}{l}{\textbf{bi-Laplacian Embedding Smoothing}} \\
        \midrule
        {\bf Smoothing}     &    \textbf{2.87} &    \phantom{0}\textbf{6.23} & \phantom{0}\textbf{6.66}    \\
        No Smoothing  & \underline{6.05} & \underline{11.34} & \underline{11.83}    \\
        \bottomrule
    \end{tabular}
    }
    \caption{Results on KITTI (train) after pre-training on FT.}
    
    \label{tab:ablations_initial}
\end{table}

\begin{table}[t!]
    \centering
    \scalebox{0.85}{
    \begin{tabular}{lccc}
        \toprule
        {\em Ablations 2/2} \phantom{xxxxxxxxxxxxxxxxi}                &               D2 &                Fl & SF \\
        \midrule
        \multicolumn{3}{l}{\textbf{Context Encoder Channels} finest to coarsest scale} \\
        \midrule
        {}[128, 192, 256]   &    \textbf{2.87} & \phantom{1}\underline{6.23} & \phantom{1}\underline{6.66} \\
        {}[256, 320, 384]             &           {3.59} &           \phantom{1}6.31 & \phantom{1}6.74             \\
        {\bf {}[320, 384, 512]}             & \underline{3.13} &    \phantom{1}\textbf{5.83} & \phantom{1}\textbf{6.21}    \\
        \midrule
        \multicolumn{3}{l}{\textbf{Dense-SE3 Neigborhood Size}} \\
        \midrule
        {\bf 32 Neighbors}  &    \textbf{3.13} &    \phantom{1}\textbf{5.83} & \phantom{1}\textbf{6.21}    \\
        256 Pixels    & \underline{3.36} & \phantom{1}\underline{6.35} & \phantom{1}\underline{6.78} \\
        \bottomrule
    \end{tabular}
    }
    \caption{Results on KITTI (train) after pre-training on FT.}
    \label{tab:ablations_rest}
    \vspace{-1mm}
\end{table}

\subsection{Ablations}
We ablate our architecture with 3-scale models pre-trained on FT and evaluated on the KITTI (train) dataset.
Initially, our experiments are performed using a small context encoder with
[128, 192, 256] channels (prior to applying the $1\times1$ \textit{conv}) on finest to coarsest layers, respectively.
The results are reported in \Cref{tab:ablations_initial}.
In later ablations, we increase the number of channels in the context encoder to 
[320, 384, 512]; see 
results in \Cref{tab:ablations_rest}.
Please note that, in the following, RAFT-like encoders refer to extensions of top-down RAFT \cite{RAFT} encoders to 
output features at the corresponding scales. 

\paragraph{Feature Encoder}
We compare the U-Net-based feature encoder employed by CCMR \cite{CCMR} to a simpler RAFT-like version (without multi-scale feature consolidation).
 The CCMR encoder achieves better performance in this case.

\paragraph{Context Encoder}
In addition to the employed top-down RAFT-like context encoder, we also investigate the U-Net-based encoder from CCMR and another variant using a Feature Pyramid Network (FPN) \cite{Lin2017_FPN}.
Since RAFT-3D uses ResNet50 \cite{He2016_ResNet} as context encoder,
we also show results of a ResNet50 encoder without and with a lightweight feature enhancement unit.
We find that, in this case, neither ResNet-based encoders nor feature consolidation nor U-Net-based encoders (CCMR) bring improvements over the top-down RAFT-like encoders.

\paragraph{Scale Initialization}
We investigate if retaining the information from the previous scale by upsampling the hidden state (HS) and the rigid motion embeddings (Mo-Emb) or discarding the information by re-initalizing is more useful.
Re-initializing the hidden state and upsampling the rigid motion embeddings gives the best results.

\paragraph{bi-Laplacian Embedding Smoothing}
This step
requires solving a large linear system, which takes up a large portion of the runtime and scales badly with increasing resolution.
However, without smoothing the performance 
clearly suffers.

\paragraph{Context Encoder Channels}
RAFT-3D employs large output channels in its context encoder, 
thus, we also investigate increasing the number of channels throughout the
context encoder, before applying the $1\times1$ \textit{conv} to unify the number of channels across scales.
We study versions with 256, 384 and 512 channels on the coarsest scale.
The number of channels on the finer scales is adapted accordingly (see \textit{supp.\ mat.}).
Overall, the largest encoder gives the best results.
Surprisingly, the smallest encoder still leads in the D2 metric.

\paragraph{Dense-SE3 Neighborhood Size}
In the Dense-SE3 layer all pixels within a neighborhood contribute to the optimization objective.
By decreasing the radius on coarser scales we can keep the area of the image that contributes the same.
However, this increases the impact of pixels close to the center.
Using a neighborhood with radius 32 across all scales performs better than decreasing the radius such that the corresponding area on the
original resolution is always 256 pixels.

\section{Conclusion}
\label{sec:conclusion}

We presented a novel recurrent multi-scale image-based scene flow approach that leverages successful concepts from optical flow to the field of scene flow estimation. 
Using a U-Net-based feature encoder, a lean 
context encoder,
a coarse-to-fine estimation scheme based on a joint upsampling of the $SE(3)$ field and the motion embeddings, a multi-scale  loss over all iterations, 
along with the consideration of high-quality disparities from a recent stereo method
, our approach achieves SOTA results on both KITTI and Spring benchmarks.

\vfill
\pagebreak
\label{sec:refs}
\bibliographystyle{IEEEbib-abbrev}
\bibliography{refs}
\cleardoublepage

\section*{Supplementary Material}


\def\x{{\mathbf x}}
\def\u{{\mathbf u}}
\def\v{{\mathbf v}}
\def\f{{\mathbf f}}
\def\d{{\mathbf d}}
\def\T{{\mathbf T}}
\def\L{{\cal L}}

In the following, we first show the architecture of our 4-scale model. Then we elaborate on our employed context encoder and finally, we demonstrate more visual results on the KITTI \cite{KITTI} and the Spring \cite{Spring} benchmark.
\section{Architecture of MS-RAFT-3D+}
\label{sec:ms_raft_3d+_architecture}
\begin{figure}[h!]
    \vspace{1mm}
    \centering
    \resizebox{0.90\columnwidth}{!}{\input{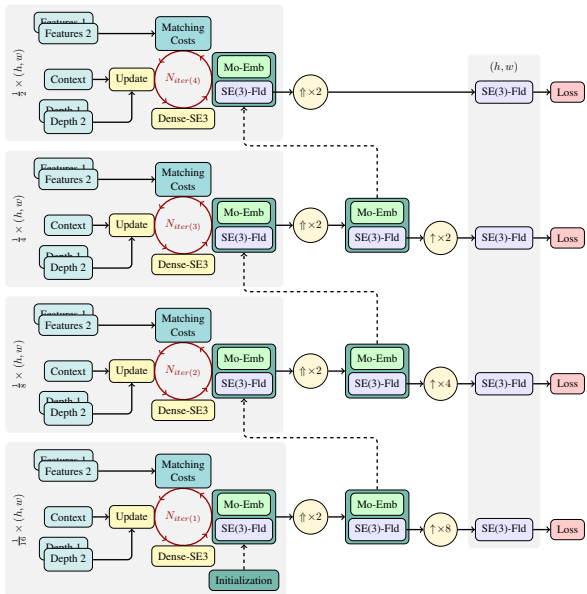}}
    \vspace{0mm}
    \caption{Architecture of MS-RAFT-3D+.}
    \label{fig:architecture_4scale}
    \vspace{-3mm}
\end{figure}

\Cref{fig:architecture_4scale} shows the architecture of our 4-scale MS-RAFT-3D+ model.
It can be seen that in addition to the three scales at [$\frac{1}{16}$, $\frac{1}{8}$, $\frac{1}{4}$], the SE(3) field is also refined at $\frac{1}{2}$ resolution. This allows to capture more details from images. Besides, no bilinear upsampling is needed to upsample the SE(3) field to full resolution, as the results after convex upsampling are already at full resolution. Note that for computing the matching costs, we used the on-demand cost computation from \cite{MS-RAFT+}.

\section{Context Encoder}
\label{sec:context_encoder}
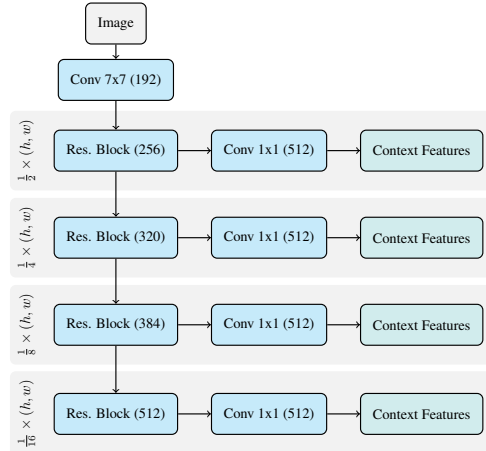
\begin{figure}[h!]
    \vspace{-2mm}
    \centering
    \resizebox{0.75\columnwidth}{!}{

\begin{tikzpicture}[]

\tikzset{every matrix/.style={ampersand replacement=\&,column sep=0.8cm,row sep=1.1cm}};
\tikzset{layer/.style={draw,thick,rounded corners,inner sep=.3cm,minimum height=1cm}};
\tikzset{connection/.style={->, thick, rounded corners}}
\tikzset{tensor/.style={fill=white!80!BlueGreen}};
\tikzset{encoder/.style={fill=white!80!cyan}};

\matrix{
    \node[layer,fill=gray!10!white] (img) {Image};
    \\[-0.75cm]
    \node[layer,encoder] (layer0) {Conv 7x7 (192)};
    \\[-0.4cm]
    \node[layer,encoder] (layer1) {Res. Block (256)}; \&
    \node[layer,encoder] (out_conv1) {Conv 1x1 (512)}; \& \node[layer,tensor] (out1) {Context Features}; 
    \\
    \node[layer,encoder] (layer2) {Res. Block (320)};  \&
    \node[layer,encoder] (out_conv2) {Conv 1x1 (512)}; \& \node[layer,tensor] (out2) {Context Features}; 
    \\
    \node[layer,encoder] (layer3) {Res. Block (384)}; \& 
    \node[layer,encoder] (out_conv3) {Conv 1x1 (512)}; \& \node[layer,tensor] (out3) {Context Features}; 
    \\[0.05cm]
    \node[layer,encoder] (layer4) {Res. Block (512)}; \&
    \node[layer,encoder] (out_conv4) {Conv 1x1 (512)}; \& \node[layer,tensor] (out4) {Context Features}; 
    \\
};

\draw[connection] (img) -- (layer0);

\foreach \layer in {1,2,3,4} {
    \pgfmathtruncatemacro{\prevlayer}{\layer - 1}
    \pgfmathtruncatemacro{\resolution}{pow(2,\layer)}

    \draw[connection] (layer\prevlayer) -- (layer\layer);
    \draw[connection] (layer\layer) -- (out_conv\layer);
    \draw[connection] (out_conv\layer) -- (out\layer);

    \node[
        fit=(layer\layer)(out_conv\layer)(out\layer),
        label={[name=label\layer,anchor=south,rotate=90,yshift=0.2cm]left:$\frac{1}{\resolution} \times (h,w)$}
    ] (scale\layer) {};
    \begin{scope}[on background layer]
        \node[
            thick,rounded corners,fill=gray!10!white,inner sep=0.075cm,
            fit=(scale\layer)(label\layer)
        ] (scale\layer) {};
    \end{scope}
}

\end{tikzpicture}}
    \vspace{0mm}
    \caption{Structure of the context encoder for four scales.}
    \label{fig:architecture_context_encoder_3scale}
    \vspace{-4mm}
\end{figure}

We use a simple top-down feature extractor to compute context features. The architecture is shown in \Cref{fig:architecture_context_encoder_3scale}. The numbers in brackets show the number of channels that is output by each module. 
Note that the number of context encoder channels in the ablations of the main paper correspond to the residual blocks, before applying the $1\times1$ \textit{conv}. Essentially, the update unit (which is responsible for computing the residual flow) is shared among scales. This means, inputs of that module at each scale must have the same number of channels. We realize this by employing $1\times1$ \textit{conv}s, which are activation-free (see \cite{CCMR}).
Please note that \Cref{fig:architecture_context_encoder_3scale} shows the context encoder for the 4-scale model.
In the case of our 3-scale model, the output of the first residual block at $\frac{1}{2}$ resolution is not passed through a $1\times1$ \textit{conv} and is not output by the encoder.

\section{Qualitative Results}
\label{sec:samples}
\begin{figure*}
\centering{
\setlength\tabcolsep{1pt}
\newcommand*\imgtrimtop{3cm}
\tikzset{labelstyle/.style={anchor=north west, fill=black, inner sep=1, text opacity=1, fill opacity=0.7, scale=0.5, xshift=3, yshift=-3, text=white}}
\tikzset{imgstyle/.style={inner sep=0,anchor=south west}}
\tikzset{rectstyle/.style={draw=black,densely dotted}}
\tikzset{rectstyle2/.style={draw=white, densely dotted}}
\newcommand{\drawlabel}[1]{%
    \draw (img.north west) node[labelstyle] {#1};
}
\newcommand{\drawrect}{
    \coordinate (A1) at (1.4,0.4);
    \coordinate (A2) at (1.75,0.65);

    \coordinate (B1) at (0.1,0.1);
    \coordinate (B2) at (0.8,0.6);
    \draw[rectstyle2] (B1) rectangle (B2); 
}

\newcommand{\croppedincludegraphics}[2][]{%
    \includegraphics[trim=0 9.0cm 0 3.8cm, clip, #1]{#2}%
}

\begin{tabular}{ccccc}
 target disparity & \textit{D2} error & optical flow & \textit{Fl} error& \textit{SF} error\\
\begin{tikzpicture}
\draw (0, 0) node[imgstyle] (img) {
\includegraphics[width=0.195\textwidth]{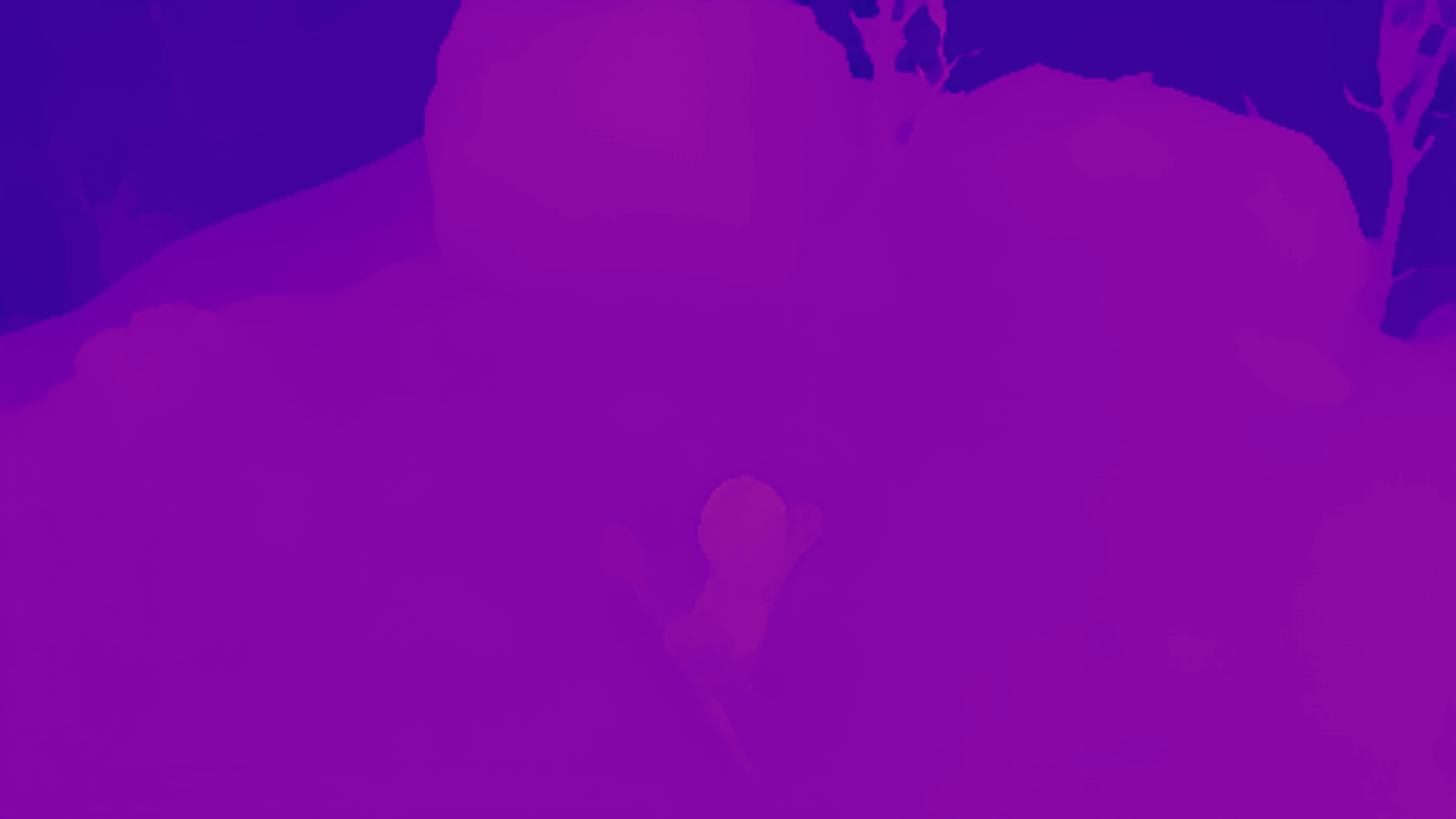}};
\drawlabel{PWOC-3D \cite{PWOC-3D}}
\end{tikzpicture} &
\begin{tikzpicture}
\draw (0, 0) node[imgstyle] (img) {
\includegraphics[width=0.195\textwidth]{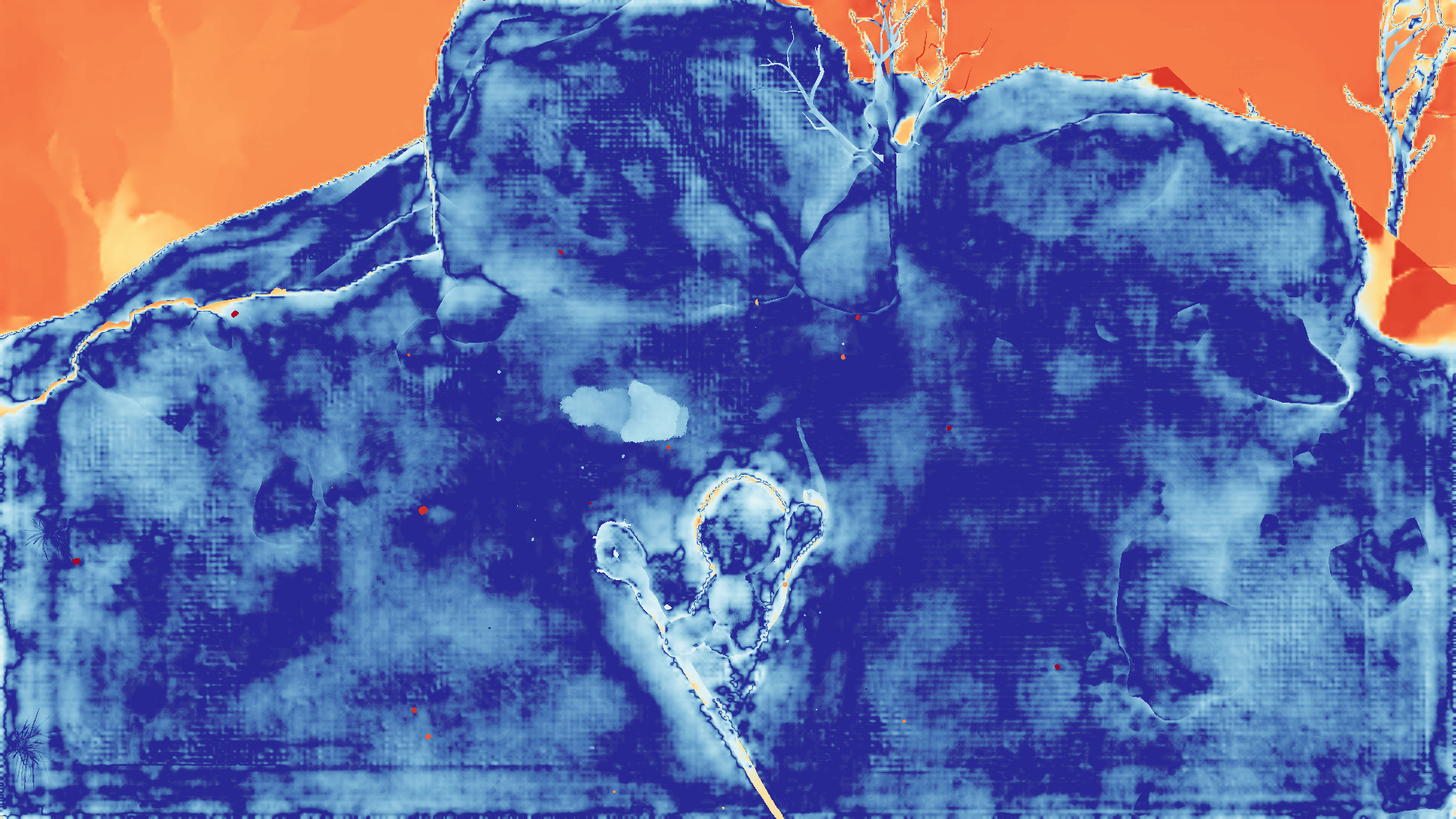}};
\end{tikzpicture} &
\begin{tikzpicture}
\draw (0, 0) node[imgstyle] (img) {
\includegraphics[width=0.195\textwidth]{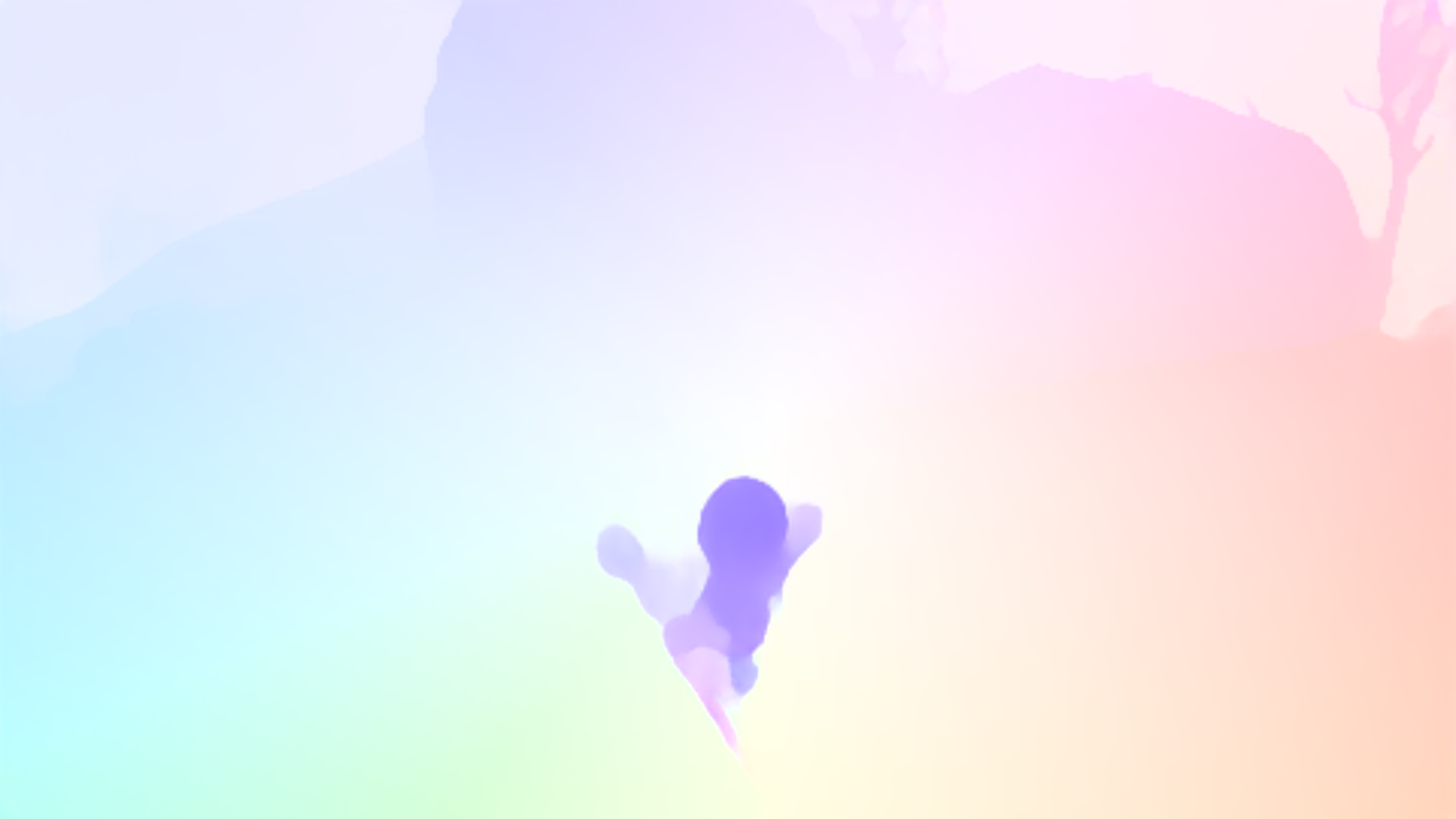}};
\end{tikzpicture} &
\begin{tikzpicture}
\draw (0, 0) node[imgstyle] (img) {
\includegraphics[width=0.195\textwidth]{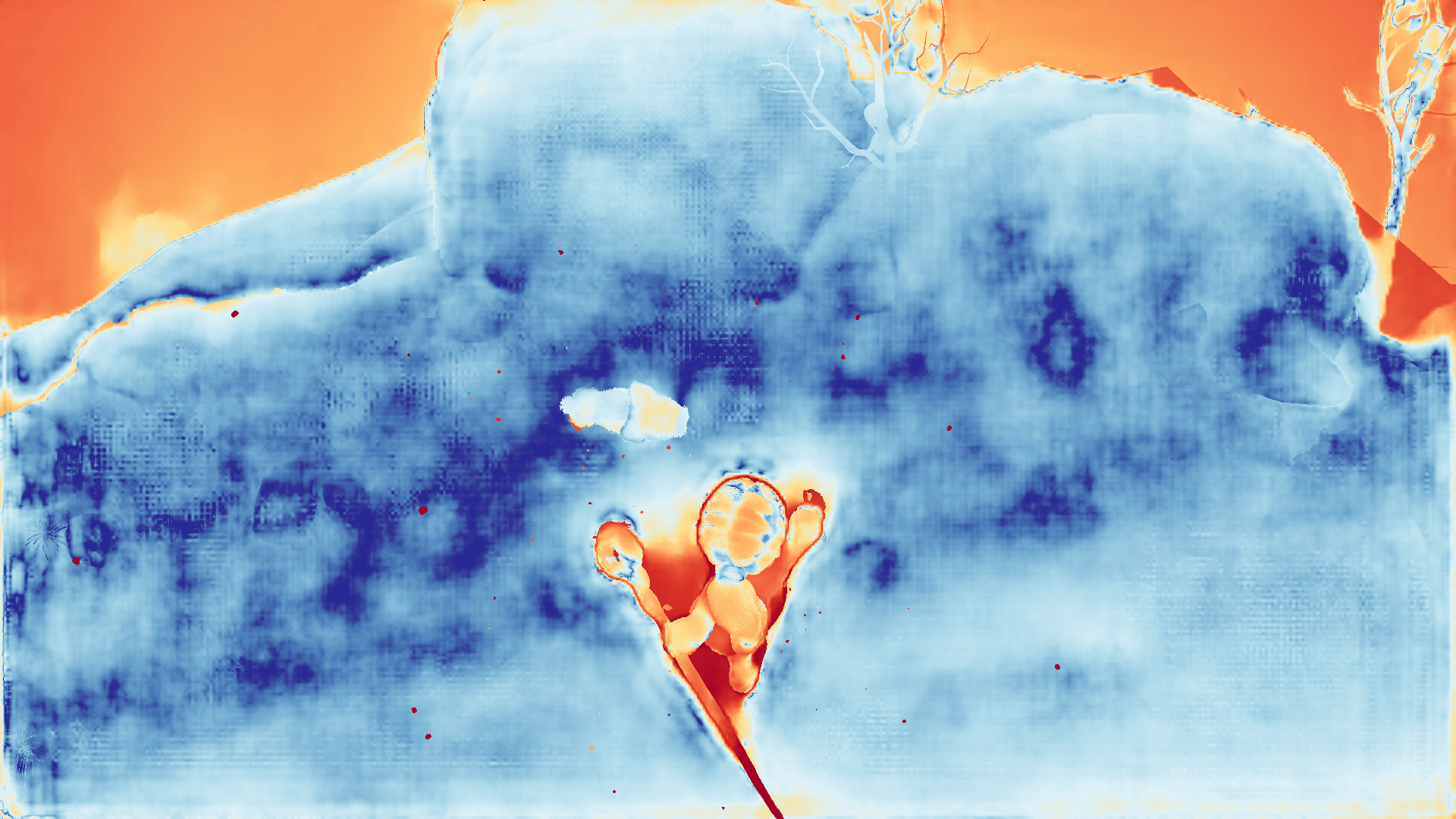}};
\end{tikzpicture} &
\begin{tikzpicture}
\draw (0, 0) node[imgstyle] (img) {
\includegraphics[width=0.195\textwidth]{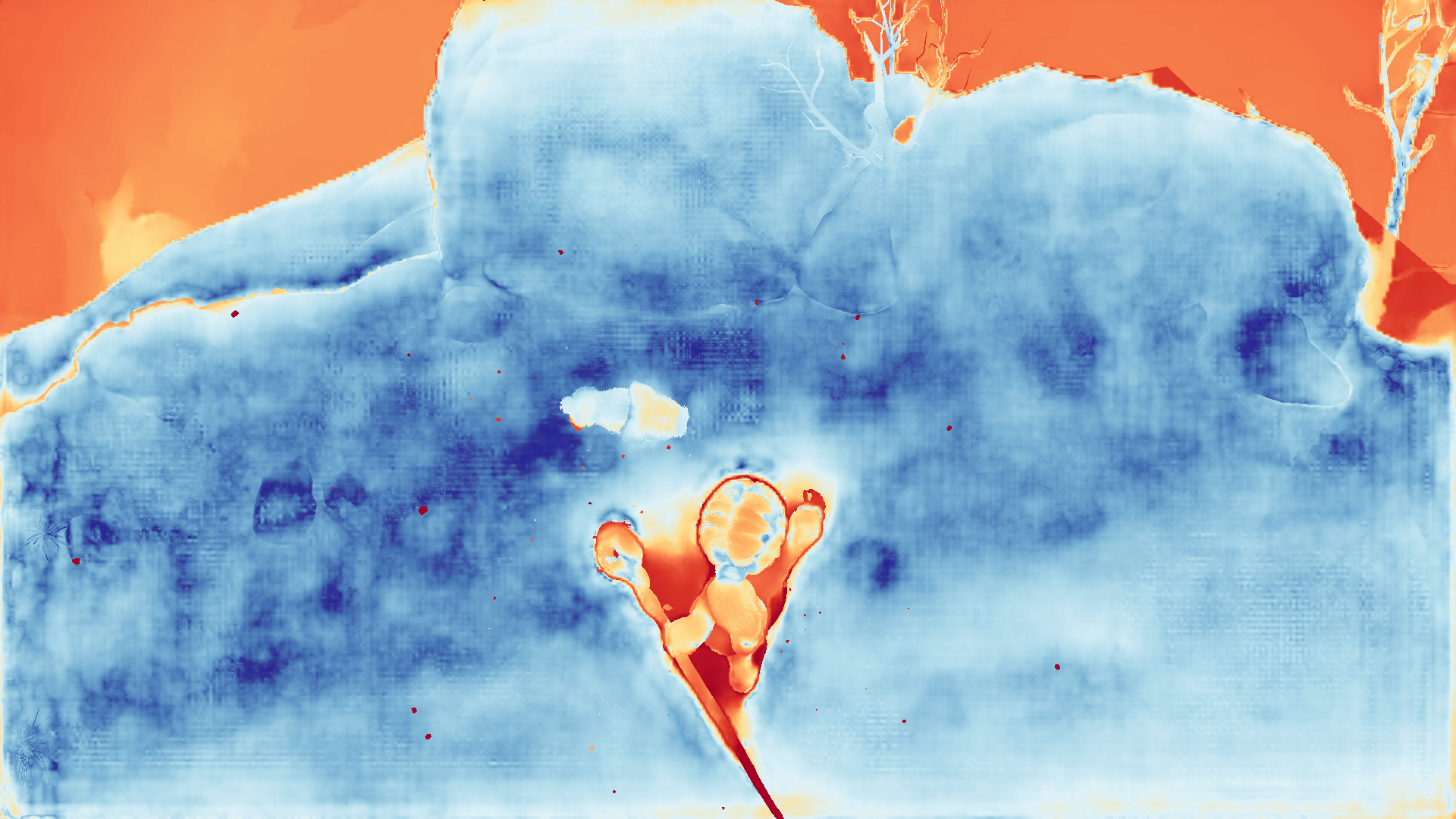}};
\end{tikzpicture}

\\[-0.6mm]
\begin{tikzpicture}
\draw (0, 0) node[imgstyle] (img) {
\includegraphics[width=0.195\textwidth]{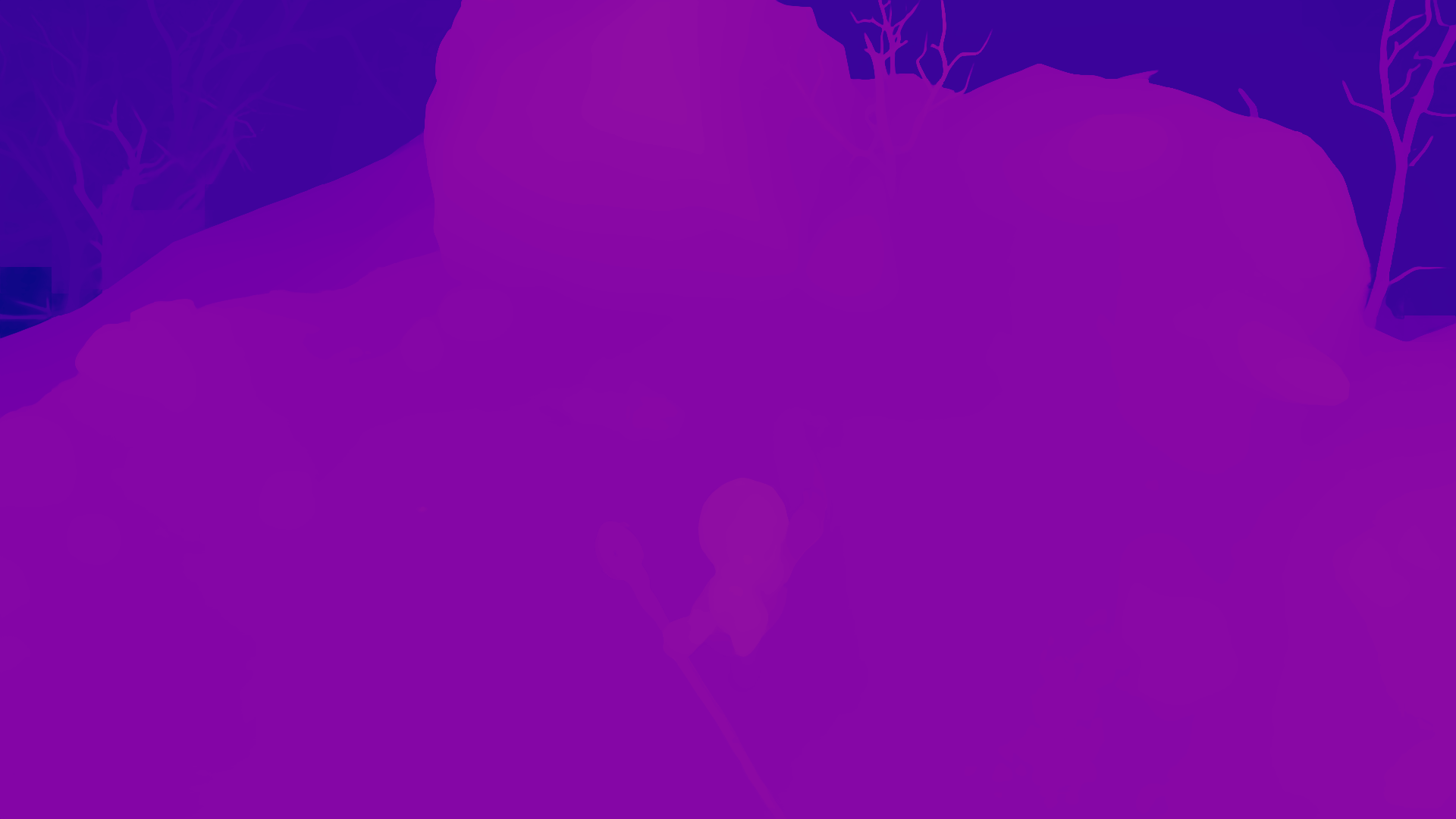}};
\drawlabel{MS-RAFT-3D}
\end{tikzpicture} &
\begin{tikzpicture}
\draw (0, 0) node[imgstyle] (img) {
\includegraphics[width=0.195\textwidth]{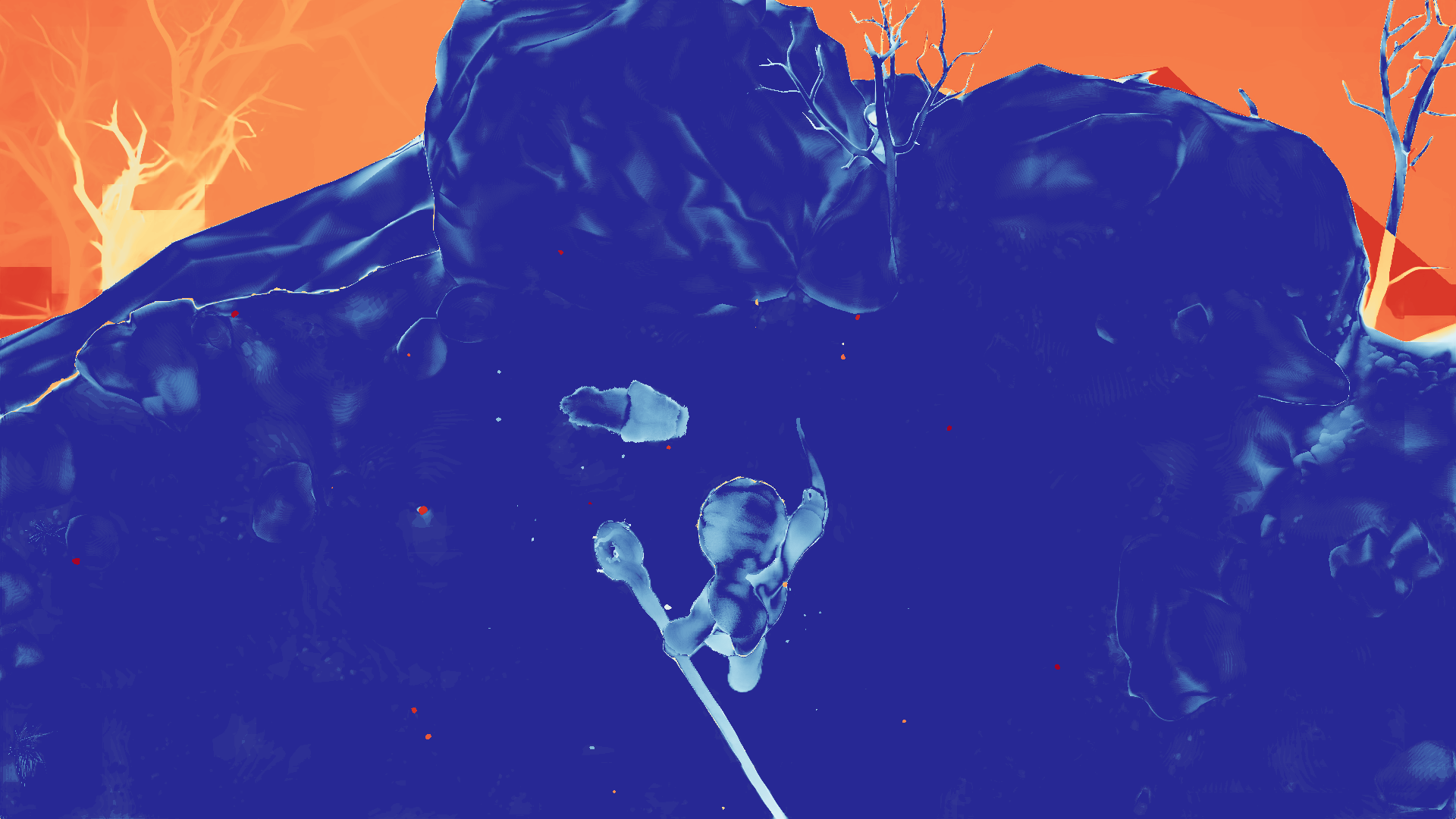}};
\end{tikzpicture} &
\begin{tikzpicture}
\draw (0, 0) node[imgstyle] (img) {
\includegraphics[width=0.195\textwidth]{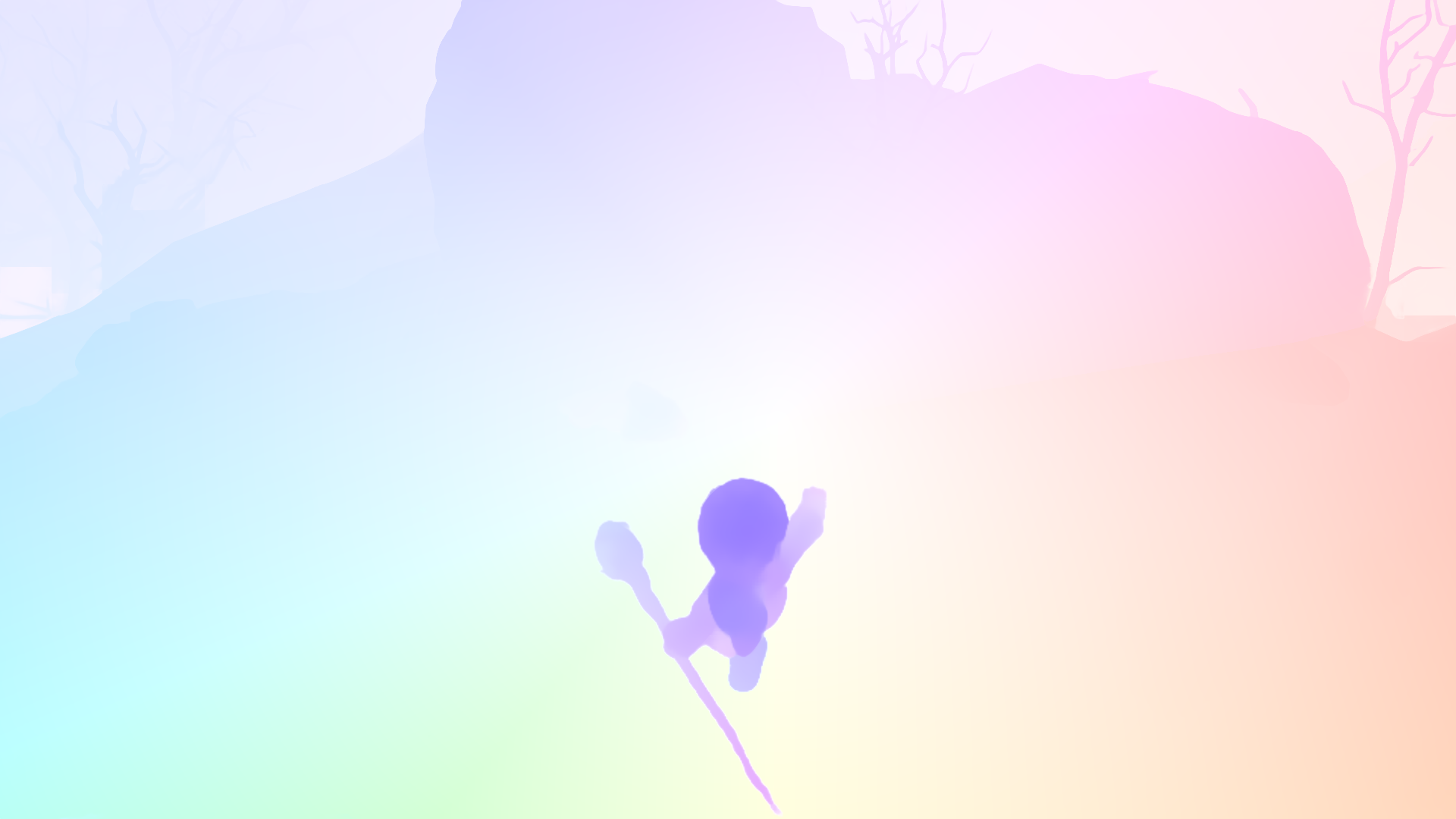}};
\end{tikzpicture} &
\begin{tikzpicture}
\draw (0, 0) node[imgstyle] (img) {
\includegraphics[width=0.195\textwidth]{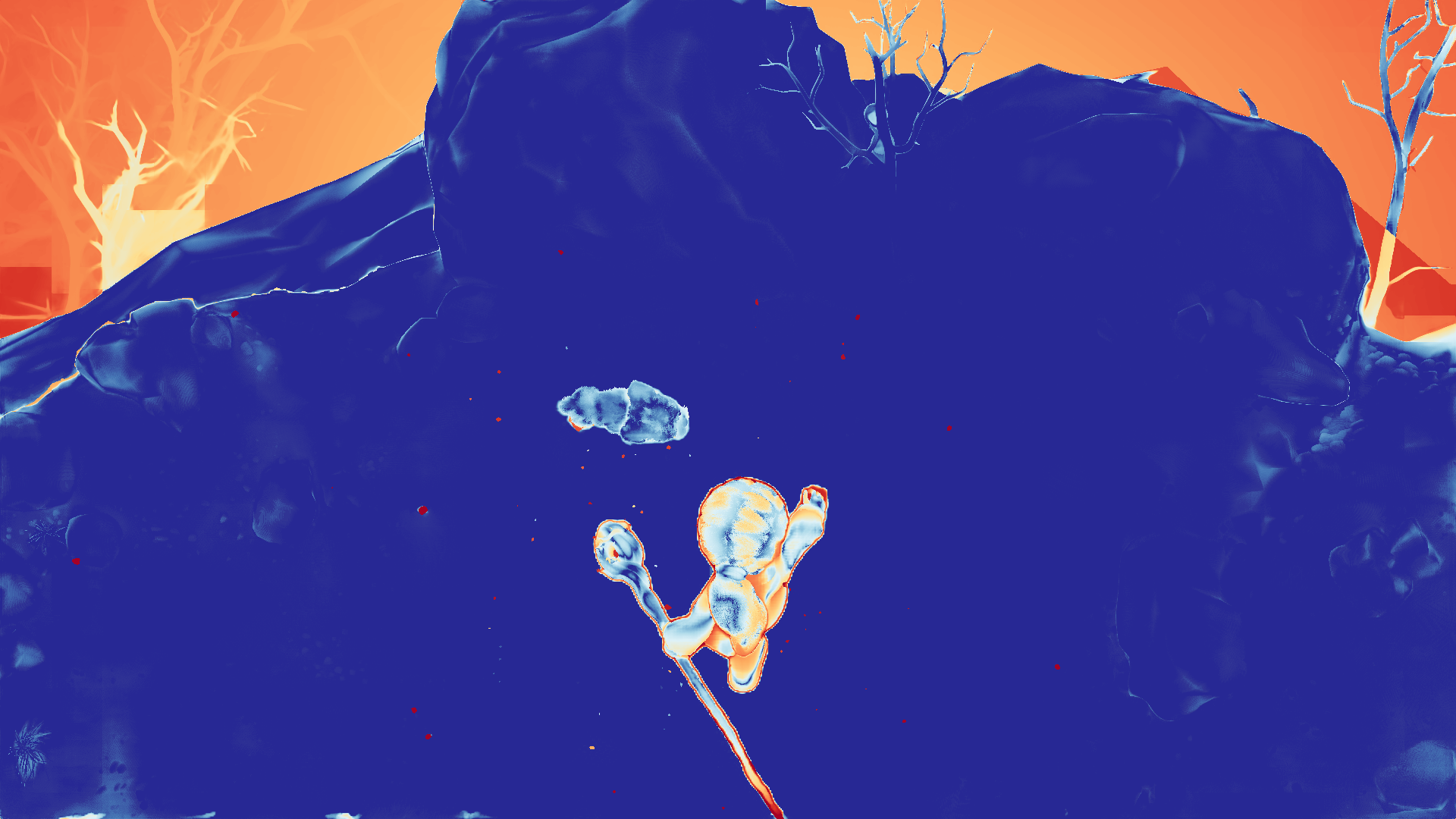}};
\end{tikzpicture} &
\begin{tikzpicture}
\draw (0, 0) node[imgstyle] (img) {
\includegraphics[width=0.195\textwidth]{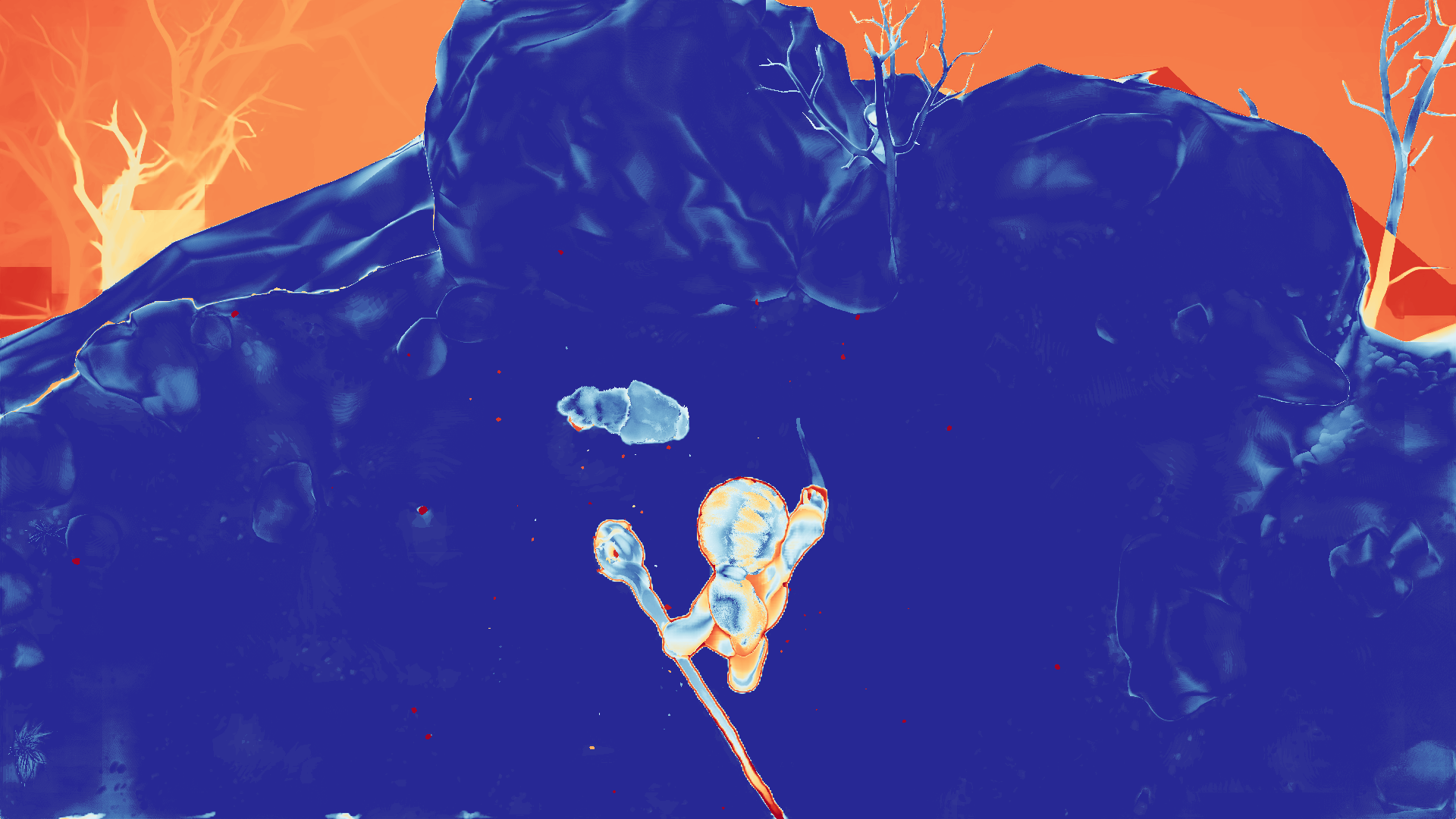}};
\end{tikzpicture}
\\[-0.6mm]

\\[-0.6mm]
\begin{tikzpicture}
\draw (0, 0) node[imgstyle] (img) {
\includegraphics[width=0.195\textwidth]{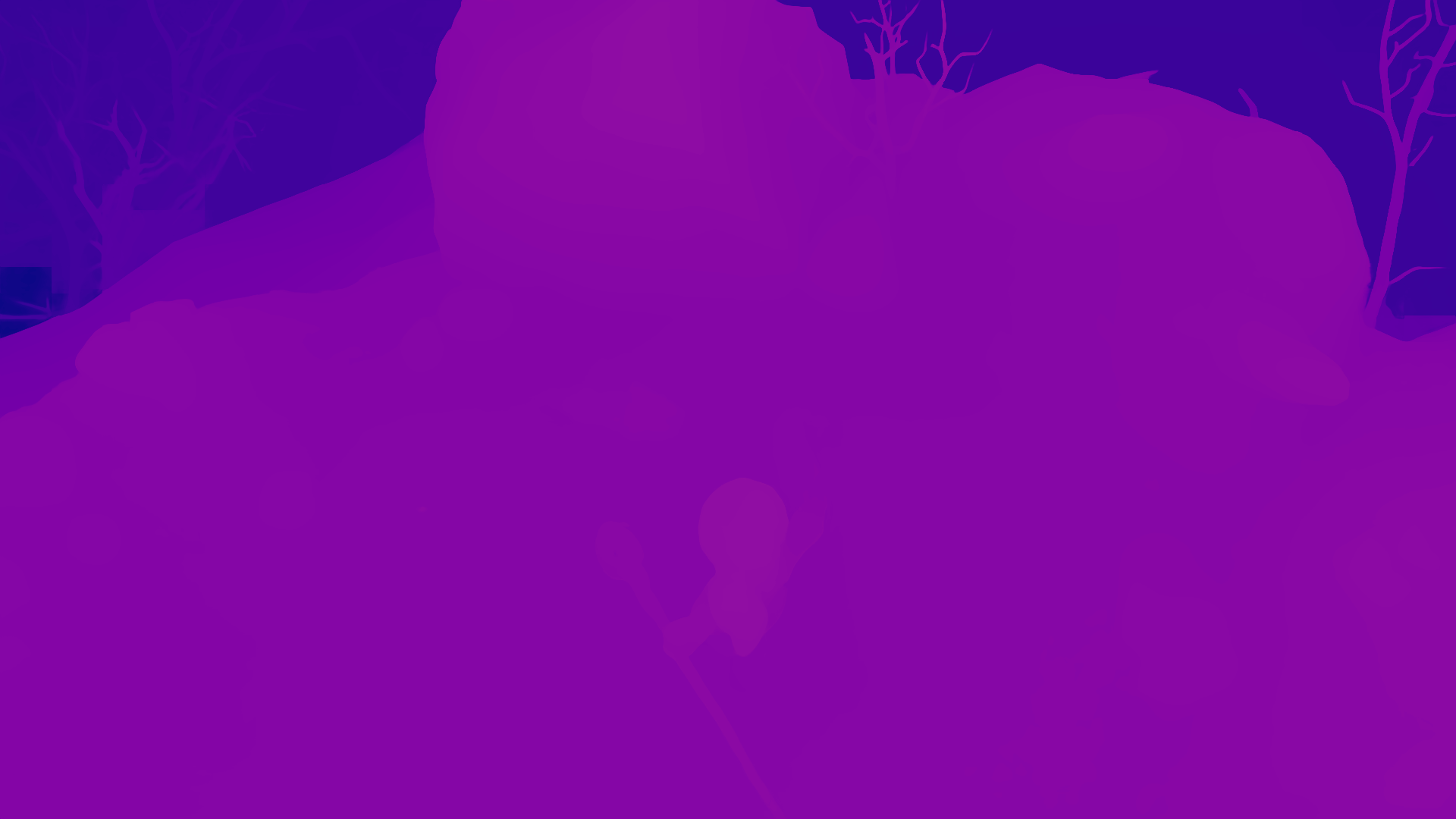}};
\drawlabel{MS-RAFT-3D+}
\end{tikzpicture} &
\begin{tikzpicture}
\draw (0, 0) node[imgstyle] (img) {
\includegraphics[width=0.195\textwidth]{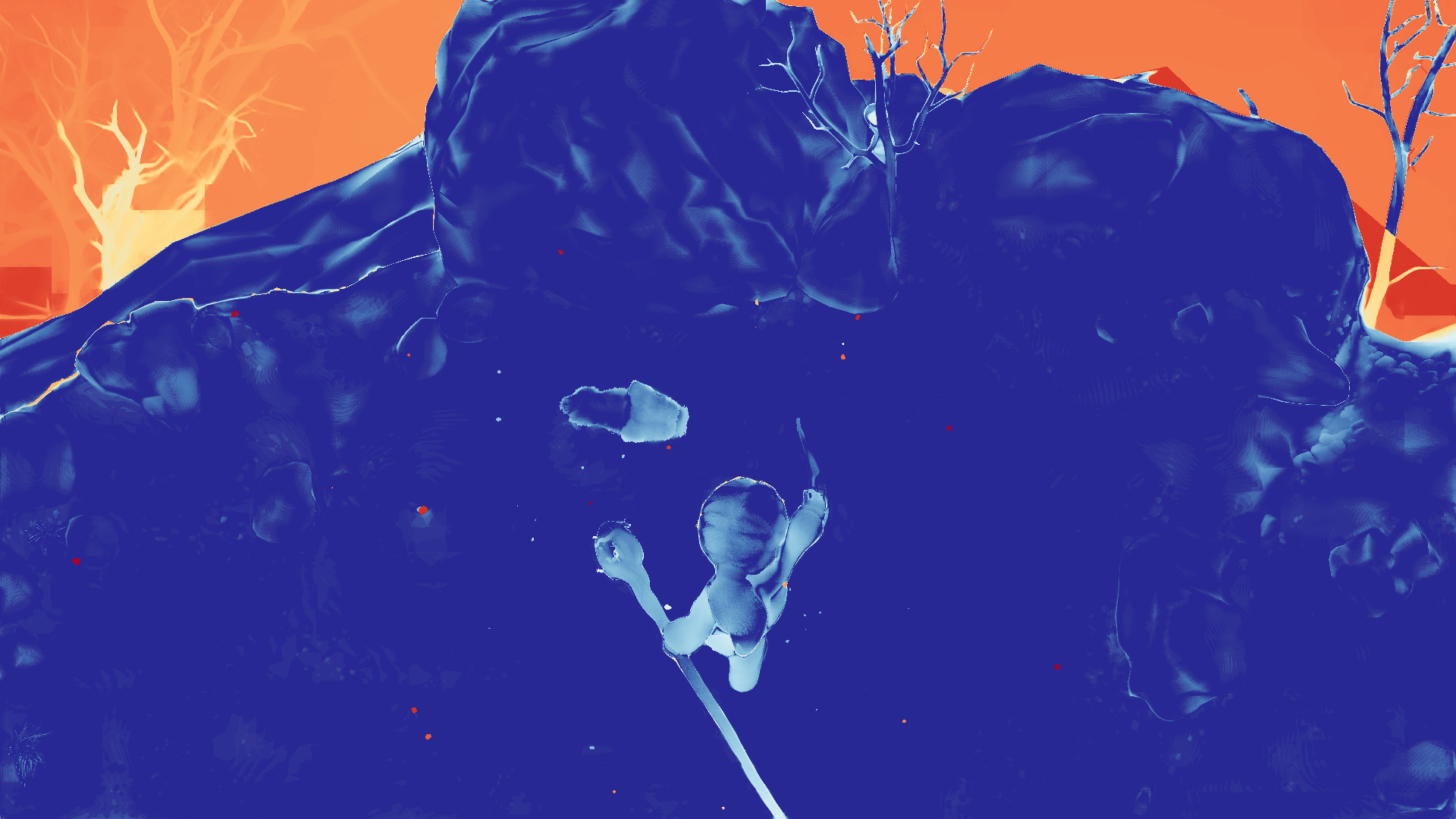}};
\end{tikzpicture} &
\begin{tikzpicture}
\draw (0, 0) node[imgstyle] (img) {
\includegraphics[width=0.195\textwidth]{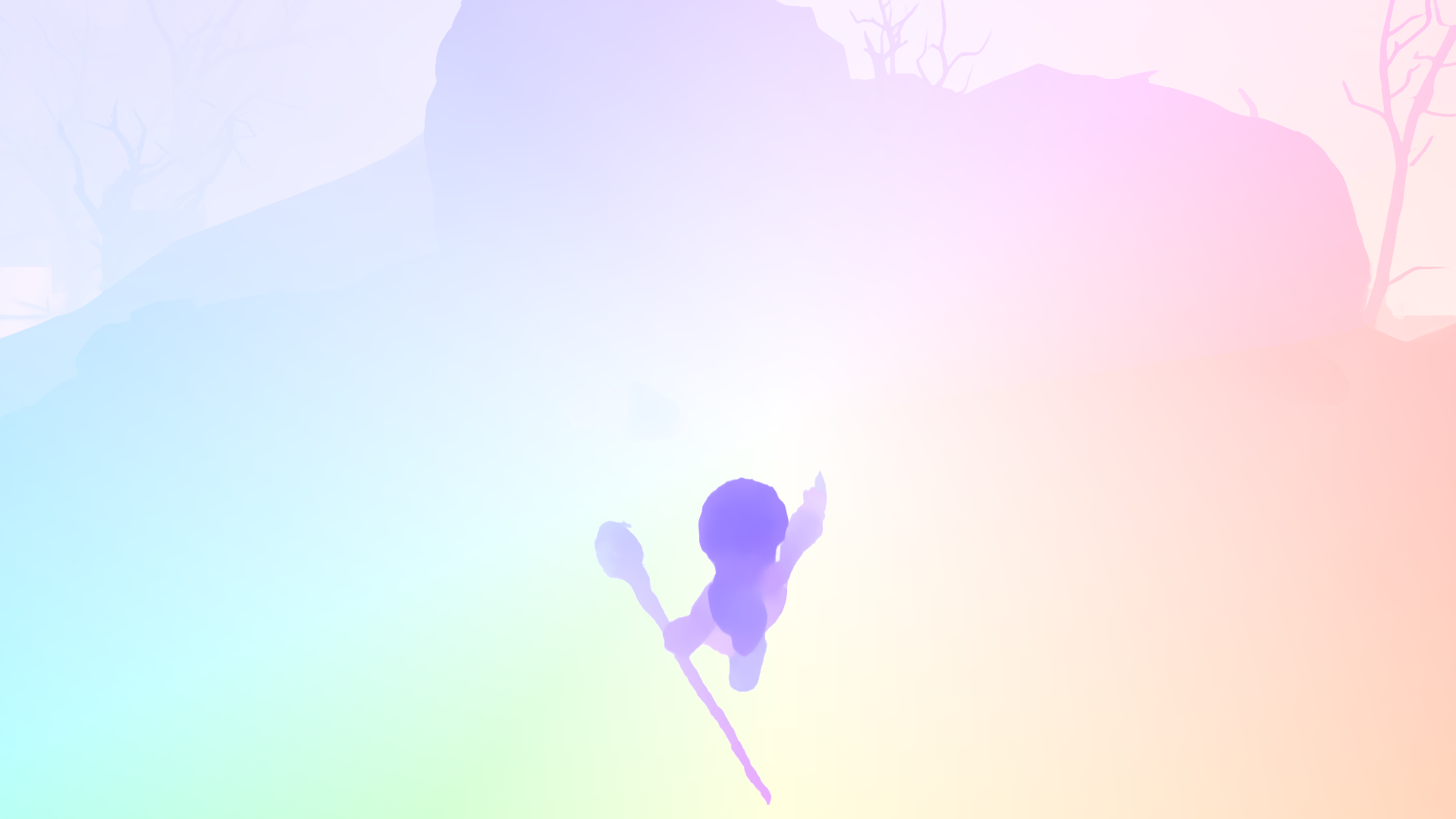}};
\end{tikzpicture} &
\begin{tikzpicture}
\draw (0, 0) node[imgstyle] (img) {
\includegraphics[width=0.195\textwidth]{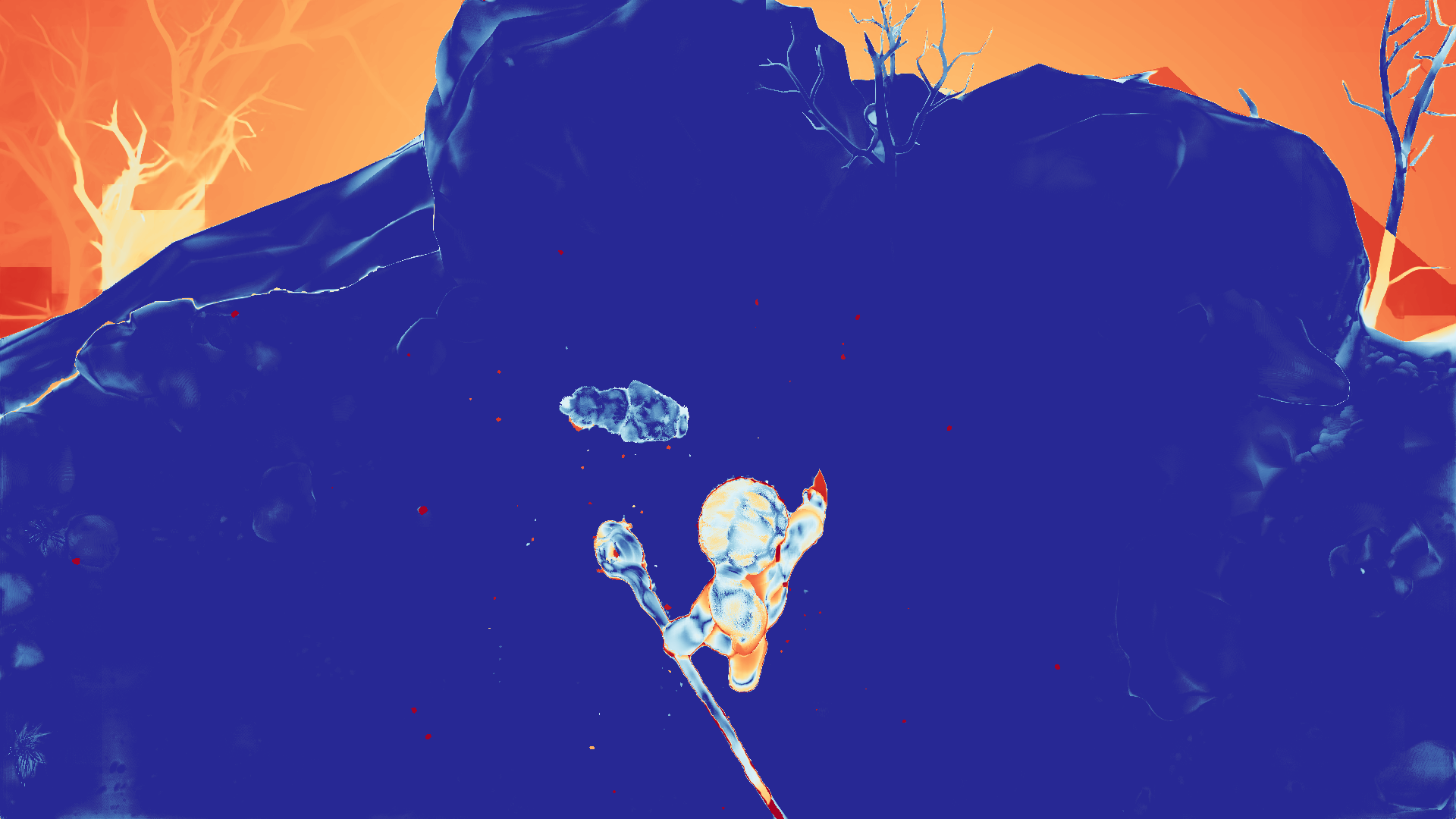}};
\end{tikzpicture} &
\begin{tikzpicture}
\draw (0, 0) node[imgstyle] (img) {
\includegraphics[width=0.195\textwidth]{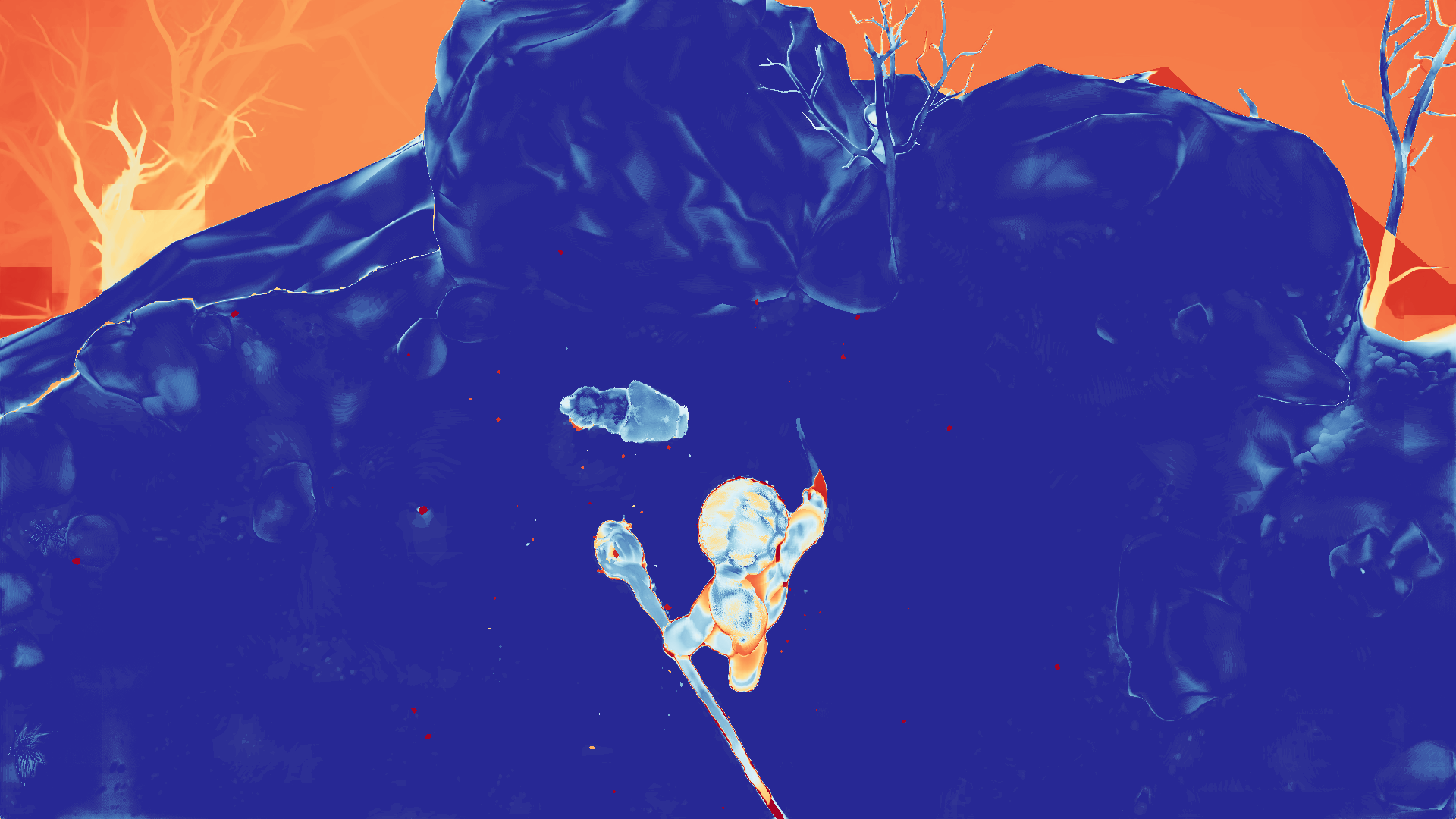}};
\end{tikzpicture}
\\[-0.6mm]

\begin{tikzpicture}
\draw (0, 0) node[imgstyle] (img) {
\includegraphics[width=0.195\textwidth]{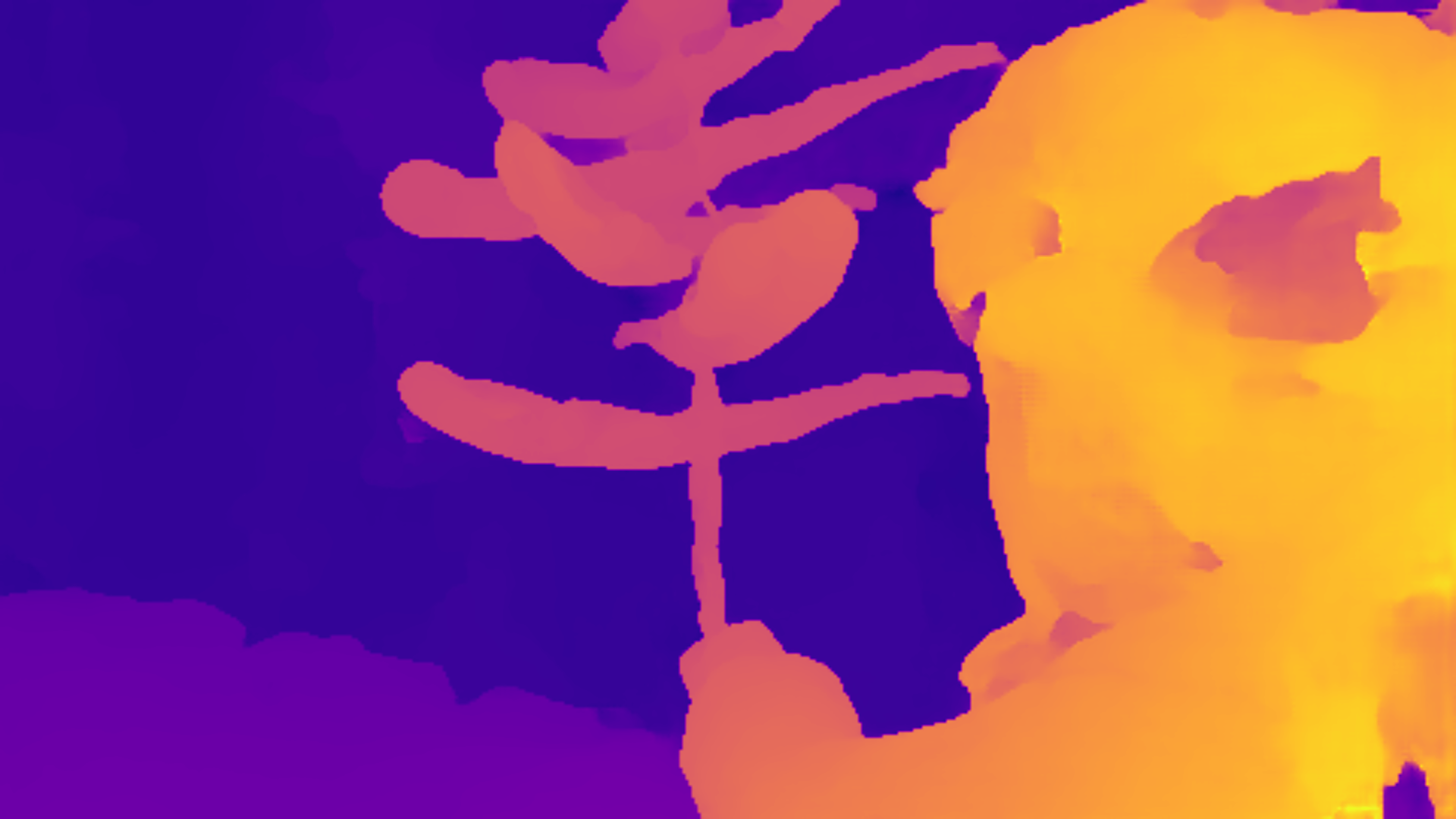}};
\drawlabel{PWOC-3D \cite{PWOC-3D}}
\end{tikzpicture} &
\begin{tikzpicture}
\draw (0, 0) node[imgstyle] (img) {
\includegraphics[width=0.195\textwidth]{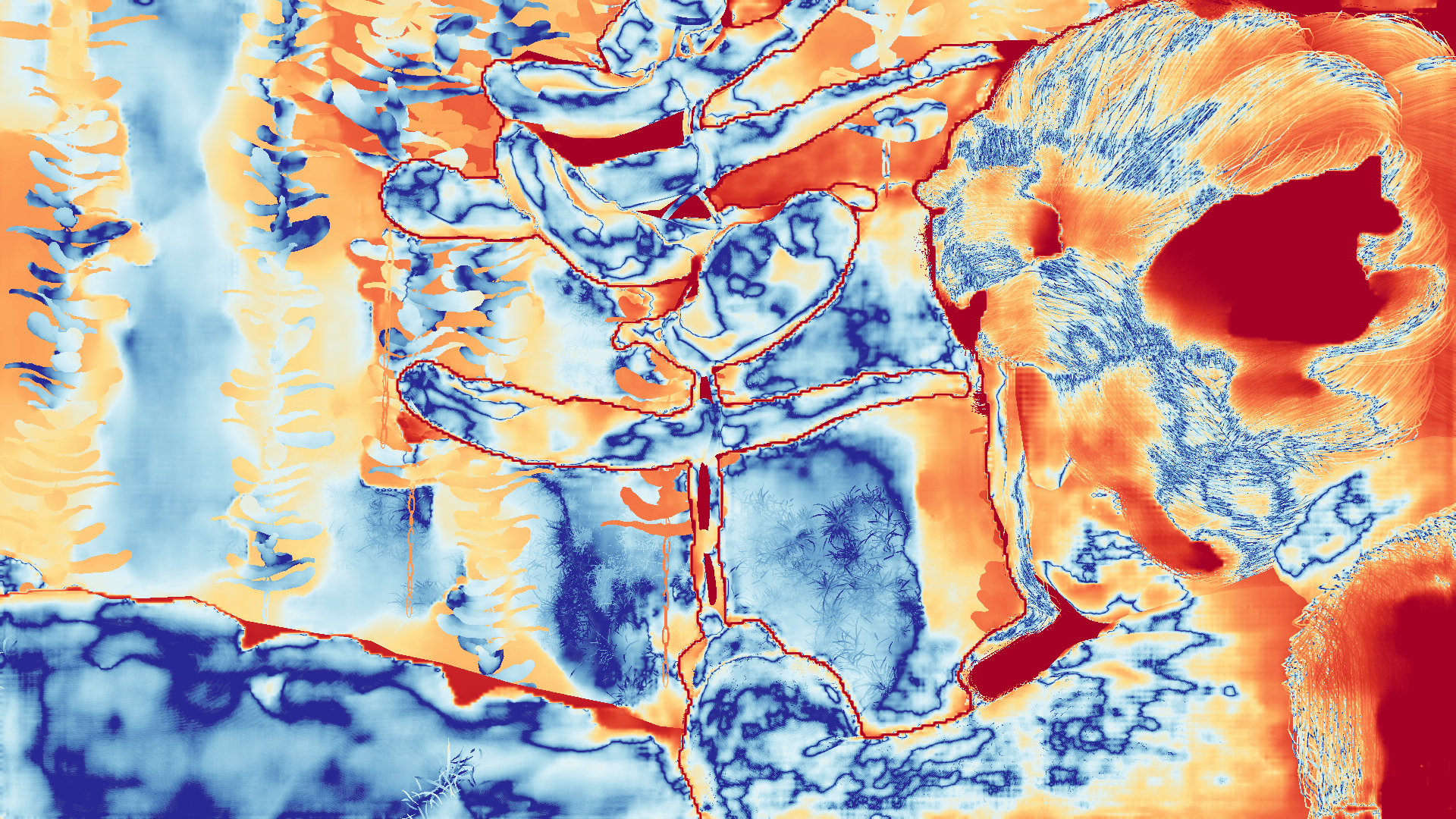}};
\end{tikzpicture} &
\begin{tikzpicture}
\draw (0, 0) node[imgstyle] (img) {
\includegraphics[width=0.195\textwidth]{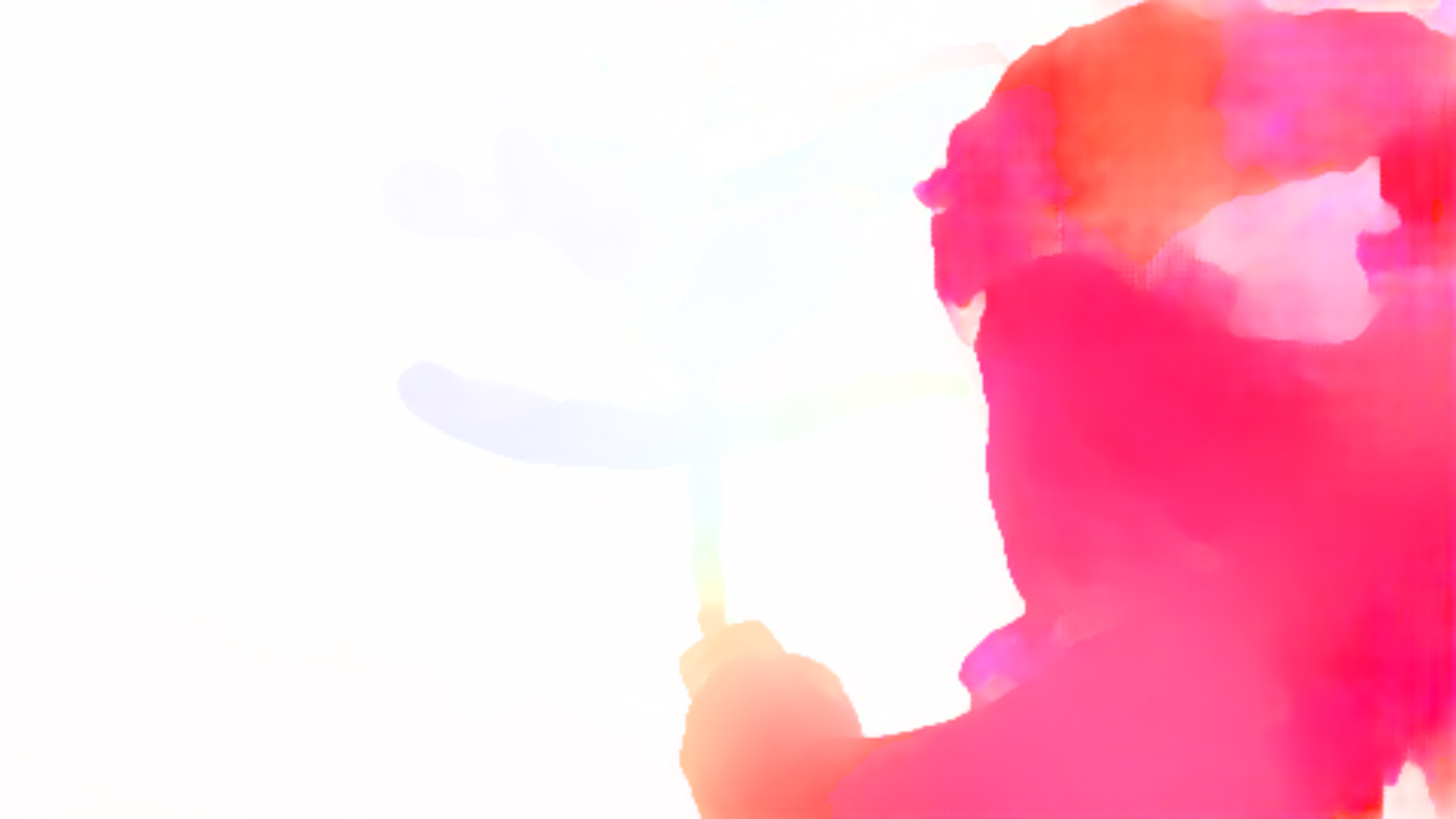}};
\end{tikzpicture} &
\begin{tikzpicture}
\draw (0, 0) node[imgstyle] (img) {
\includegraphics[width=0.195\textwidth]{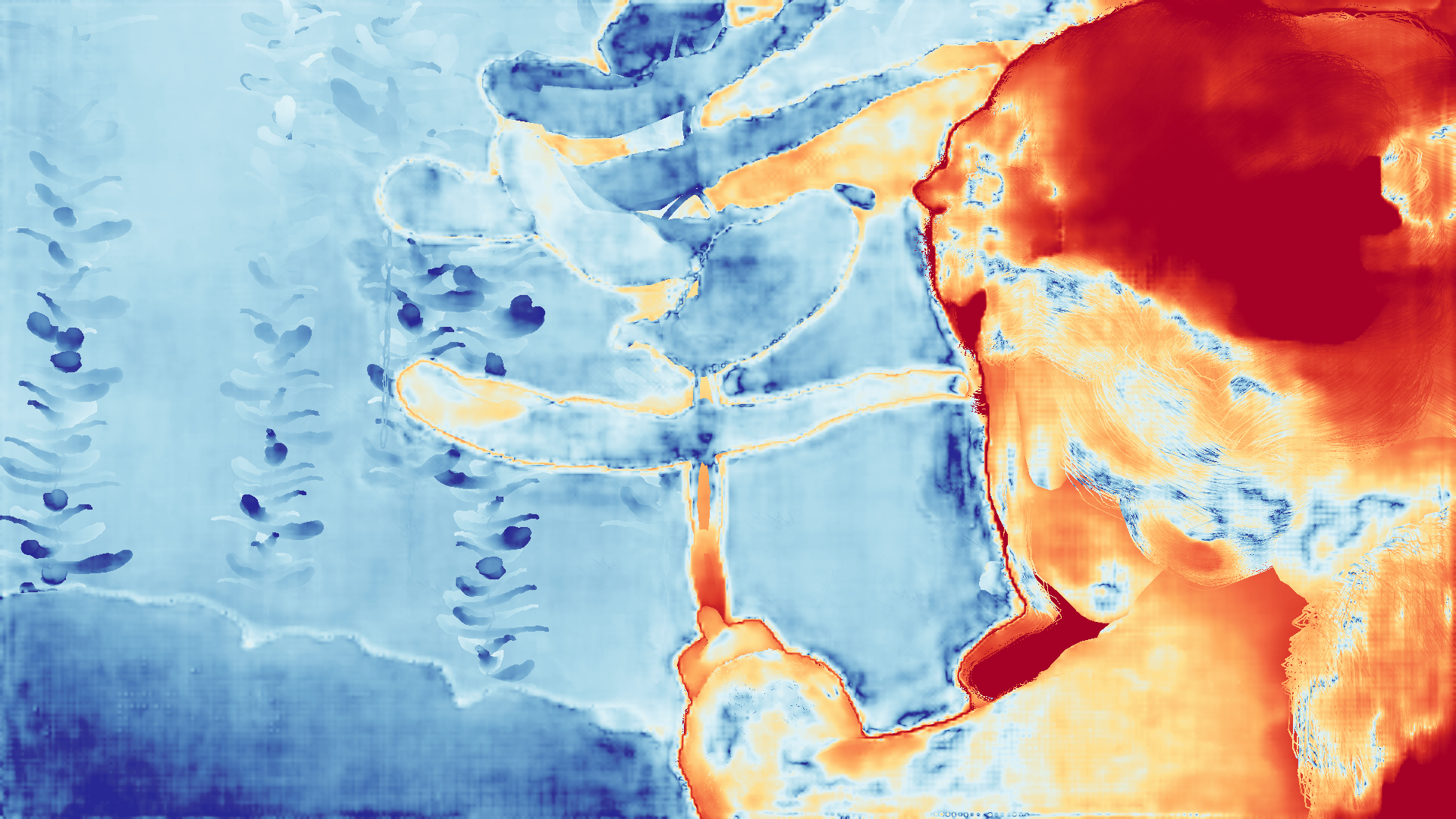}};
\end{tikzpicture} &
\begin{tikzpicture}
\draw (0, 0) node[imgstyle] (img) {
\includegraphics[width=0.195\textwidth]{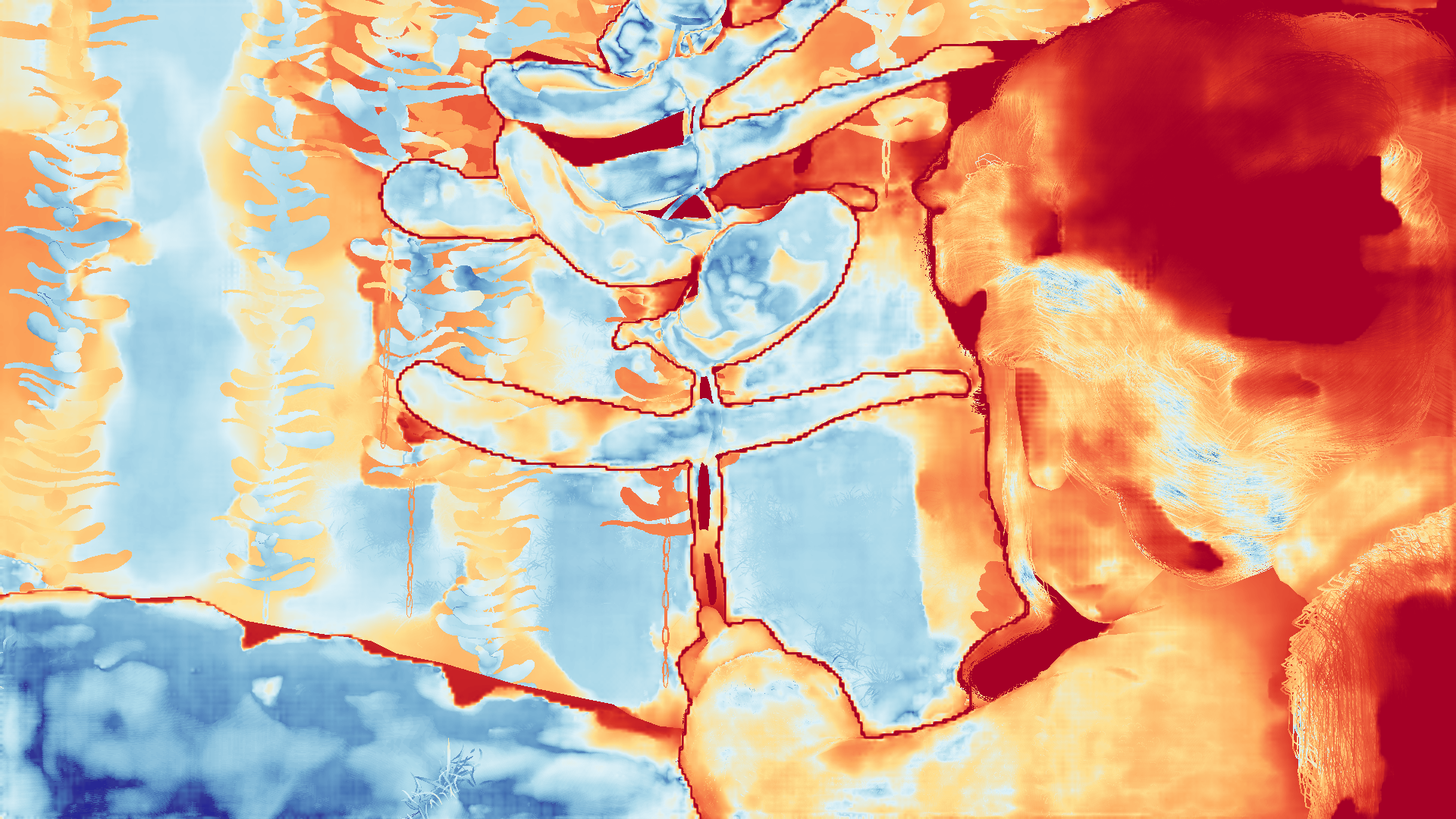}};
\end{tikzpicture}

\\[-0.6mm]
\begin{tikzpicture}
\draw (0, 0) node[imgstyle] (img) {
\includegraphics[width=0.195\textwidth]{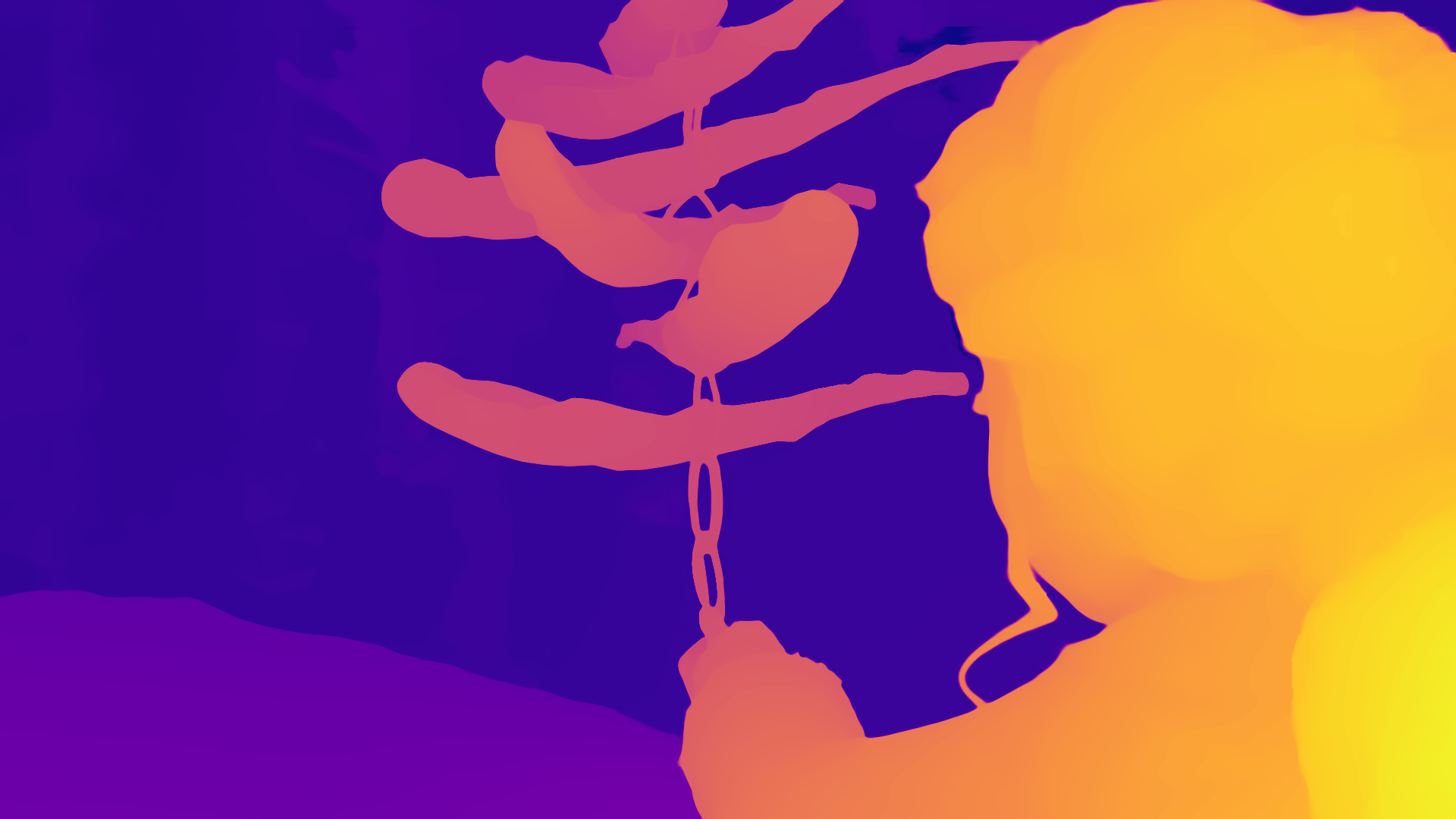}};
\drawlabel{MS-RAFT-3D}
\end{tikzpicture} &
\begin{tikzpicture}
\draw (0, 0) node[imgstyle] (img) {
\includegraphics[width=0.195\textwidth]{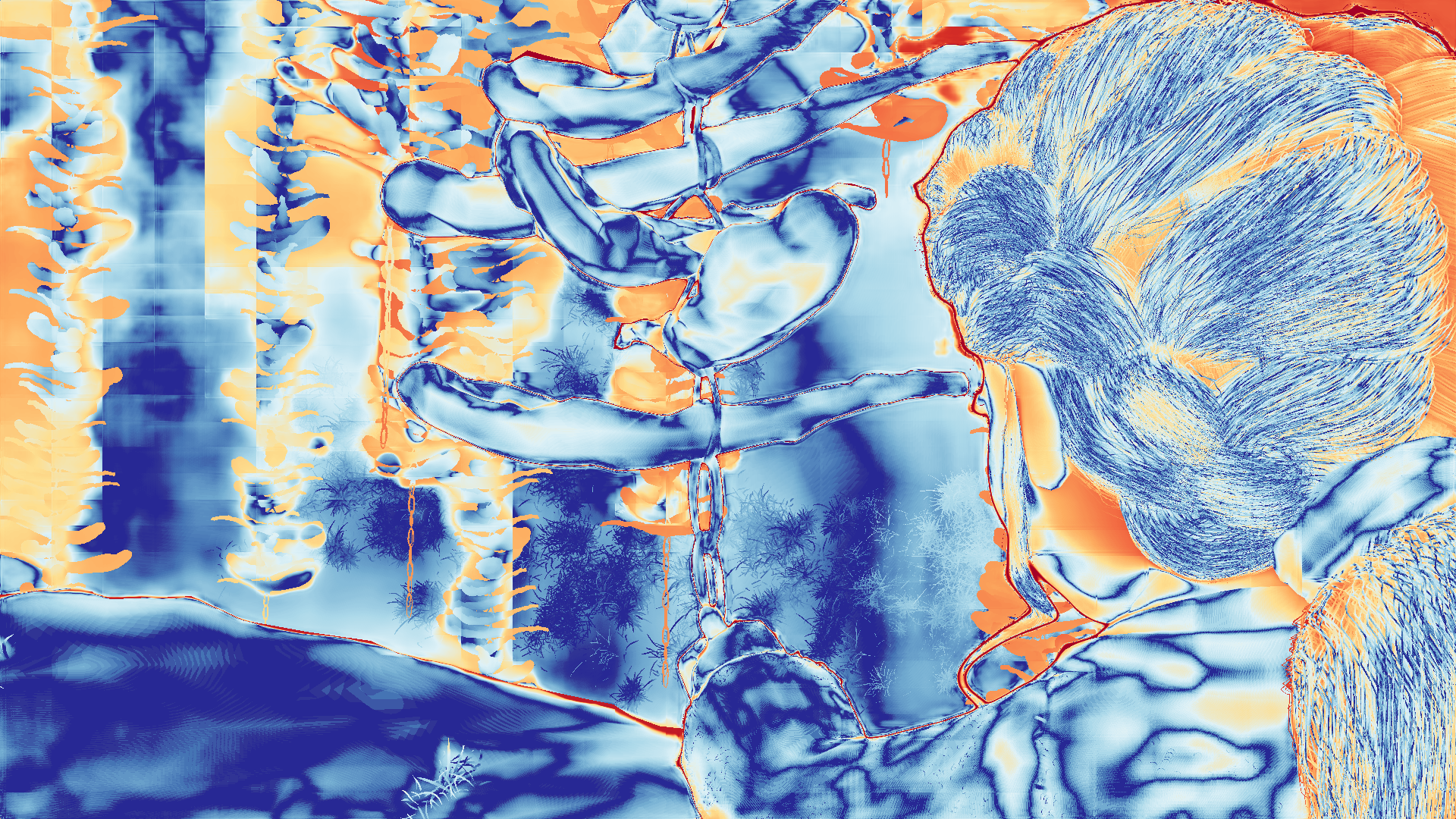}};
\end{tikzpicture} &
\begin{tikzpicture}
\draw (0, 0) node[imgstyle] (img) {
\includegraphics[width=0.195\textwidth]{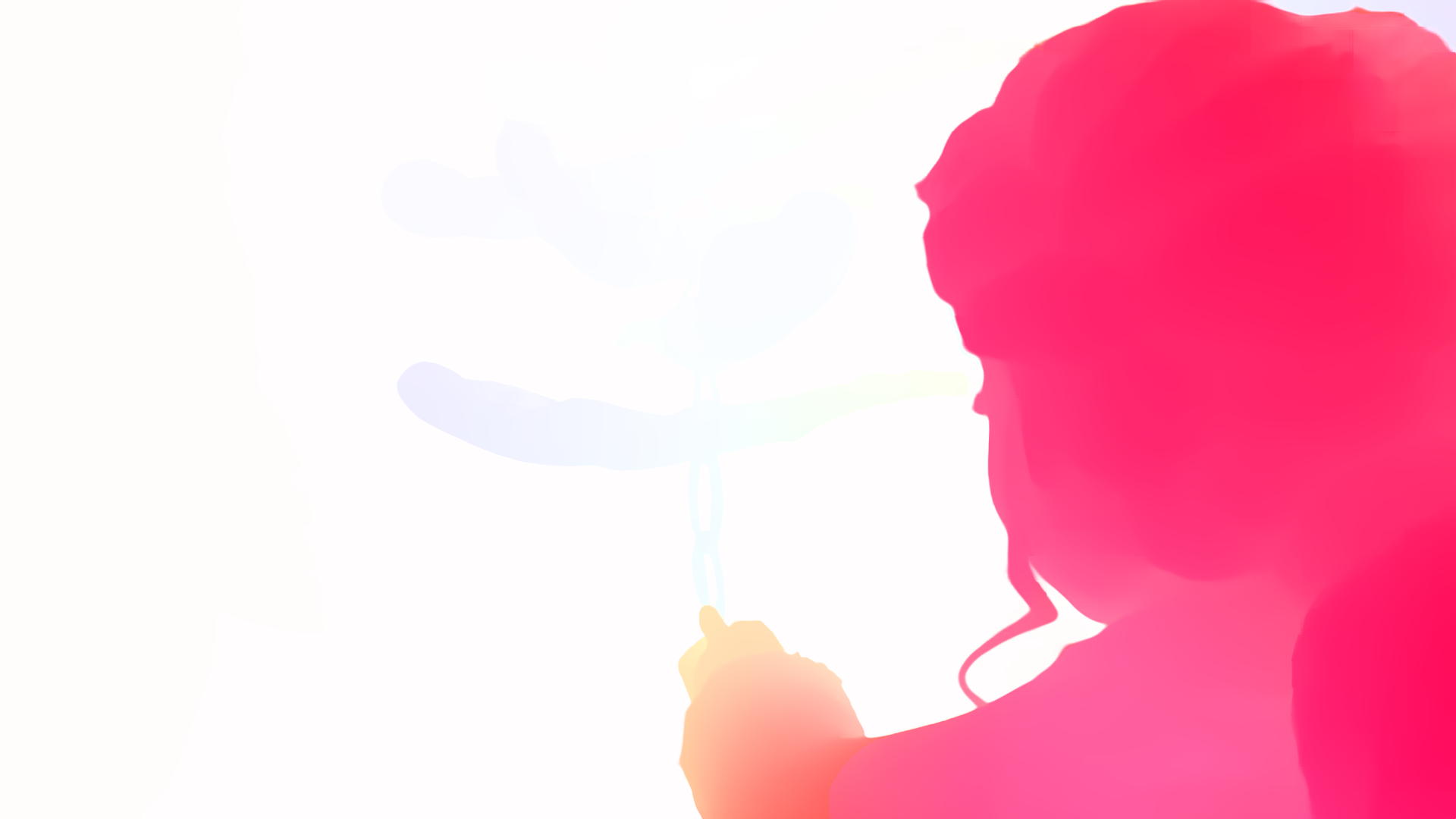}};
\end{tikzpicture} &
\begin{tikzpicture}
\draw (0, 0) node[imgstyle] (img) {
\includegraphics[width=0.195\textwidth]{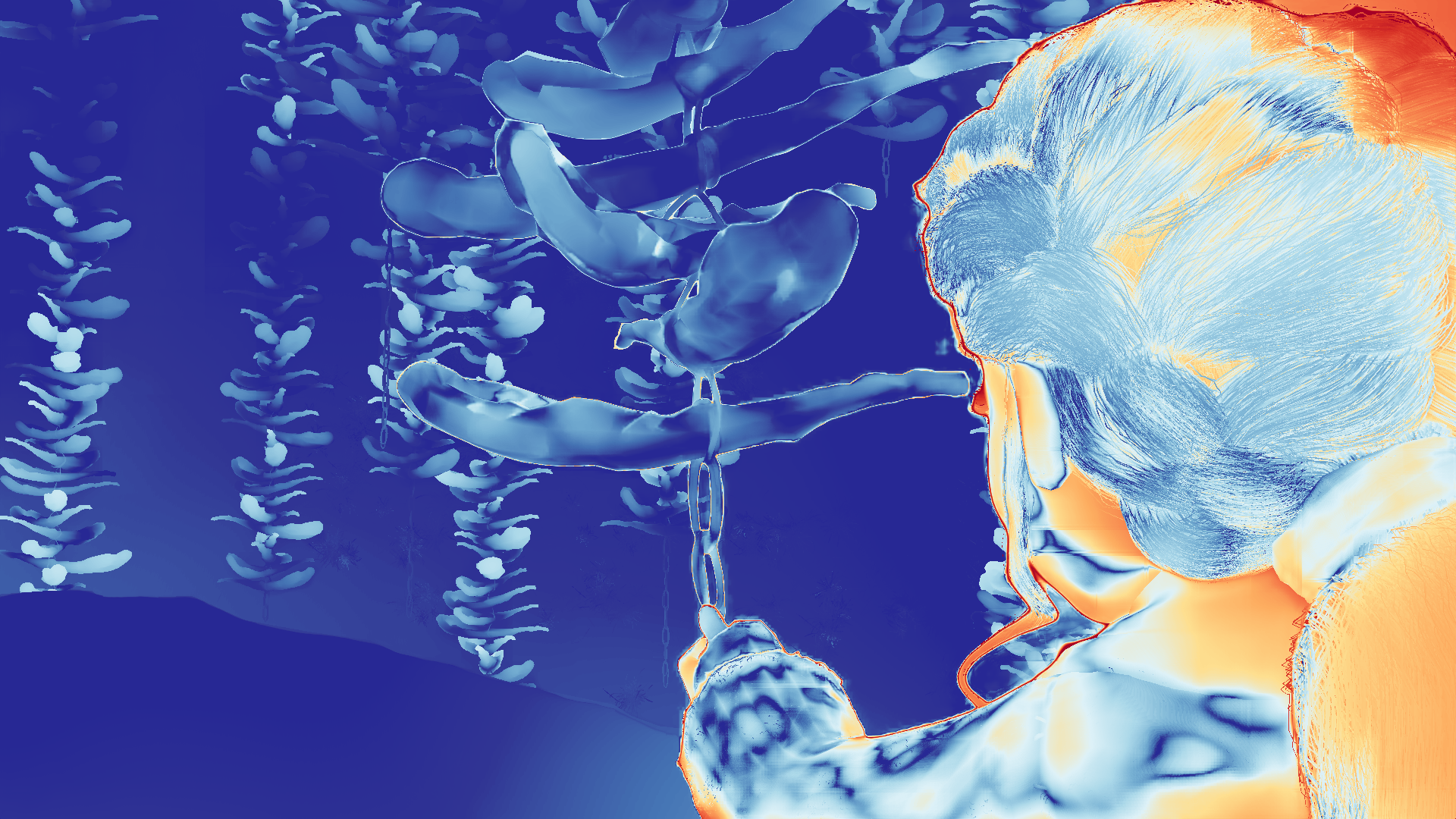}};
\end{tikzpicture} &
\begin{tikzpicture}
\draw (0, 0) node[imgstyle] (img) {
\includegraphics[width=0.195\textwidth]{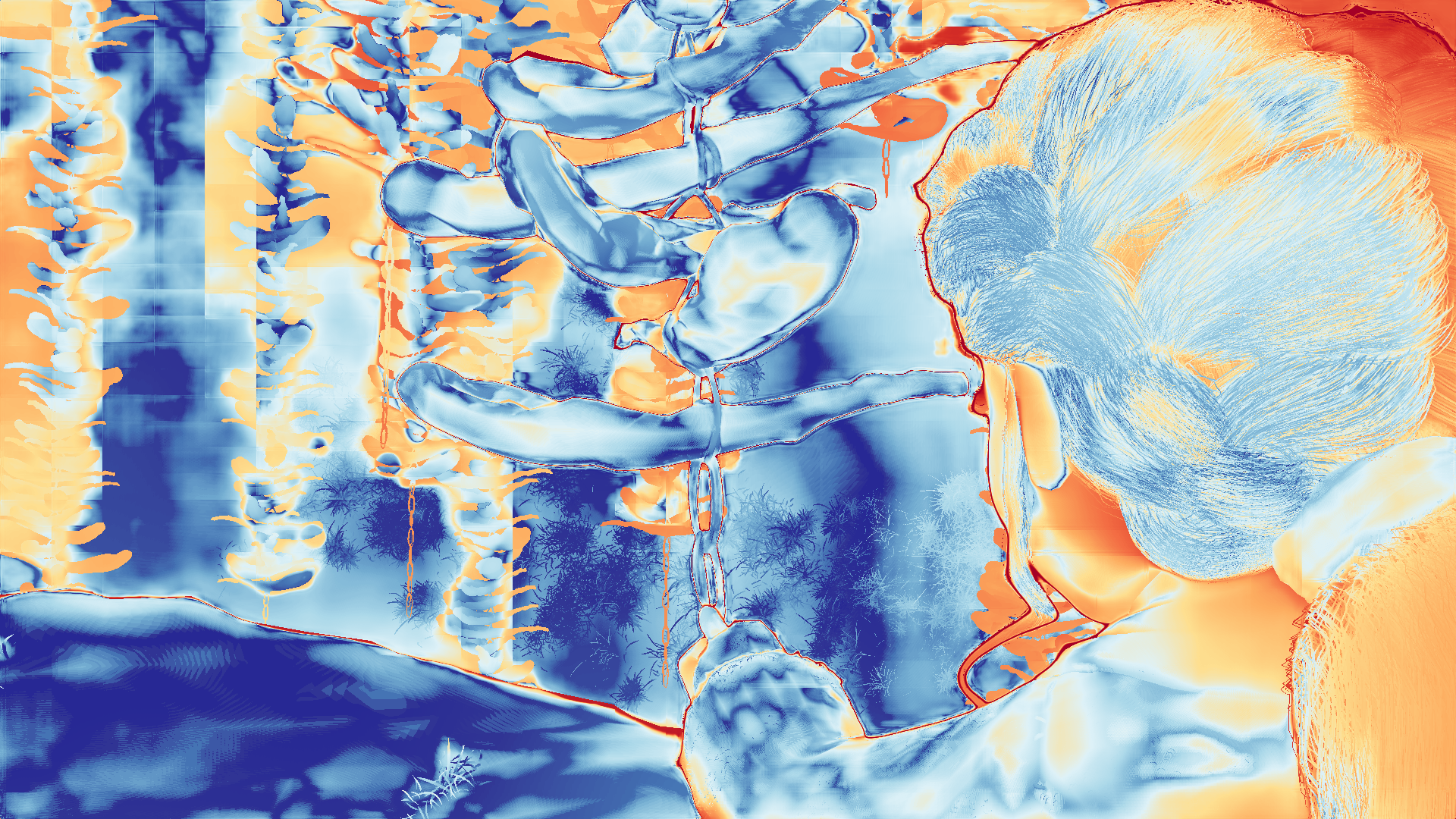}};
\end{tikzpicture}

\\[-0.6mm]
\begin{tikzpicture}
\draw (0, 0) node[imgstyle] (img) {
\includegraphics[width=0.195\textwidth]{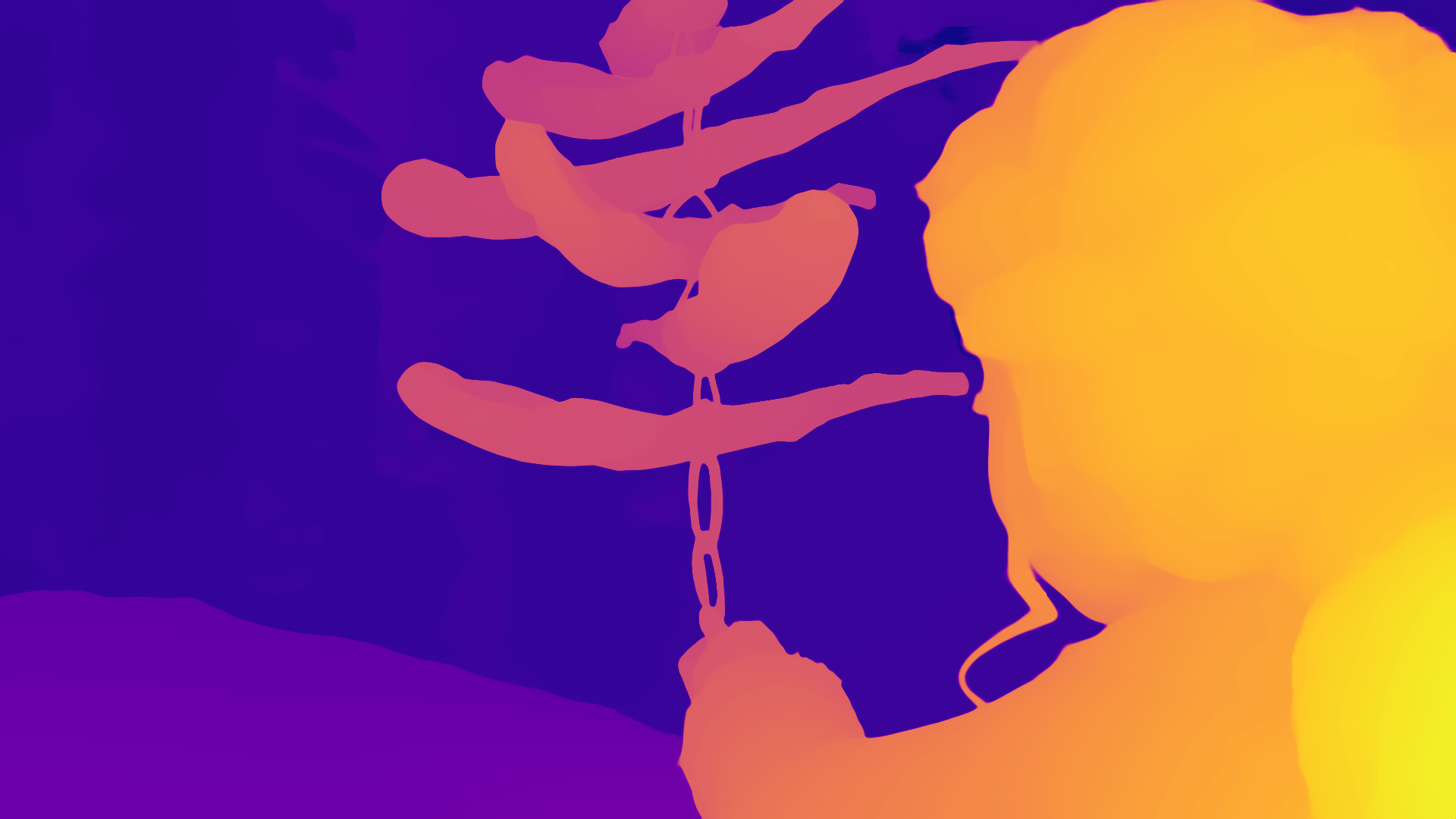}};
\drawlabel{MS-RAFT-3D+}
\end{tikzpicture} &
\begin{tikzpicture}
\draw (0, 0) node[imgstyle] (img) {
\includegraphics[width=0.195\textwidth]{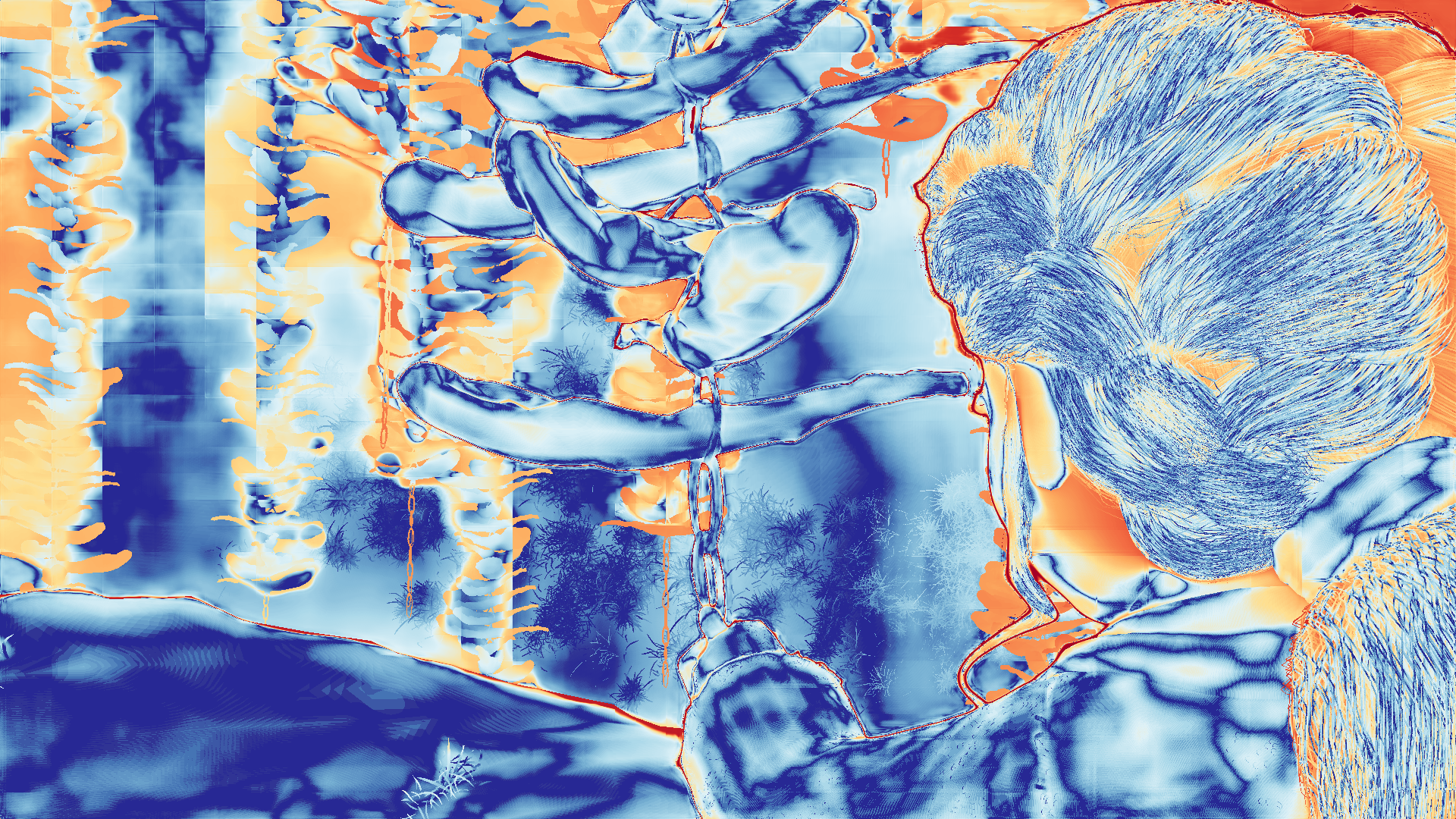}};
\end{tikzpicture} &
\begin{tikzpicture}
\draw (0, 0) node[imgstyle] (img) {
\includegraphics[width=0.195\textwidth]{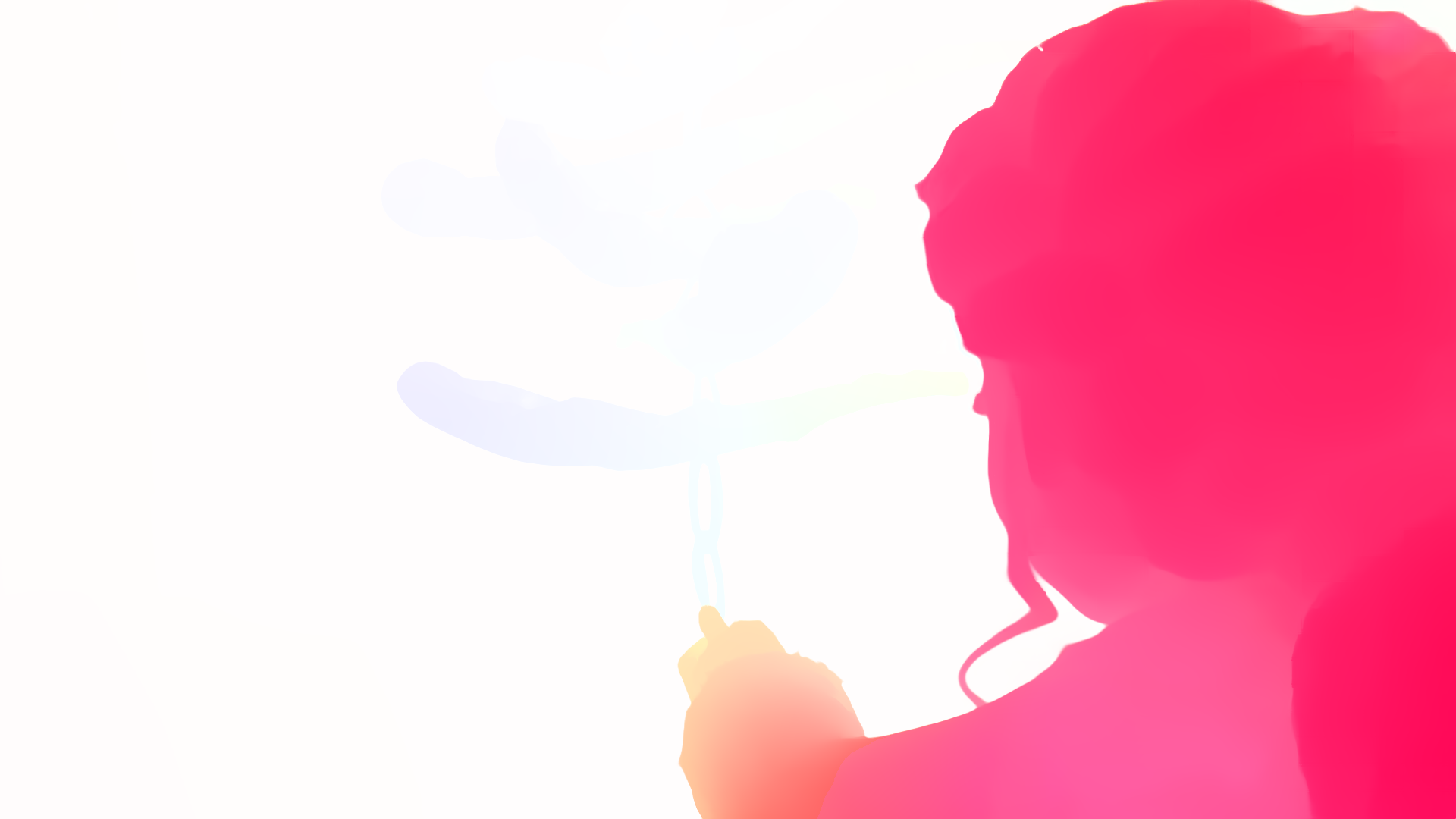}};
\end{tikzpicture} &
\begin{tikzpicture}
\draw (0, 0) node[imgstyle] (img) {
\includegraphics[width=0.195\textwidth]{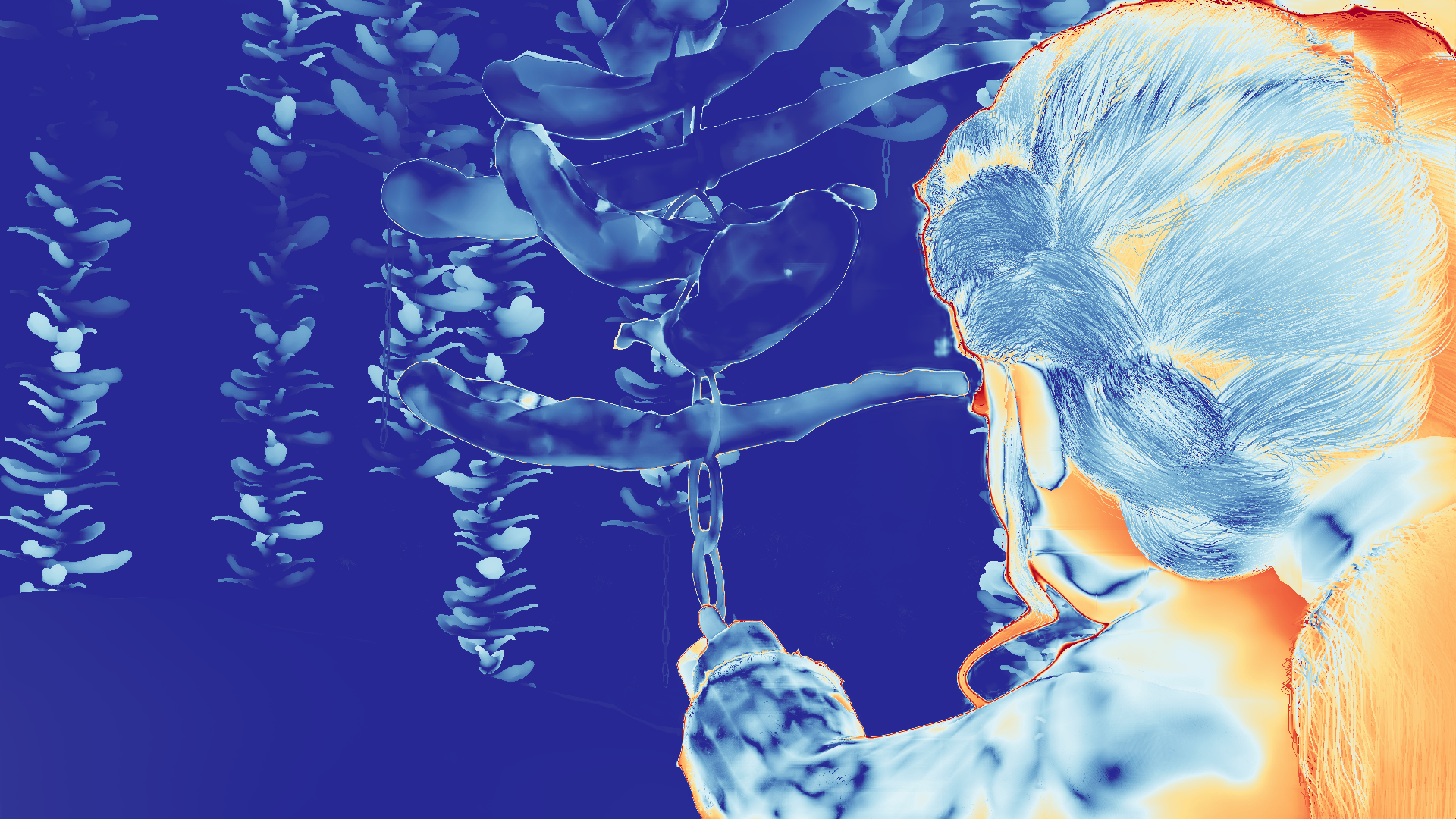}};
\end{tikzpicture} &
\begin{tikzpicture}
\draw (0, 0) node[imgstyle] (img) {
\includegraphics[width=0.195\textwidth]{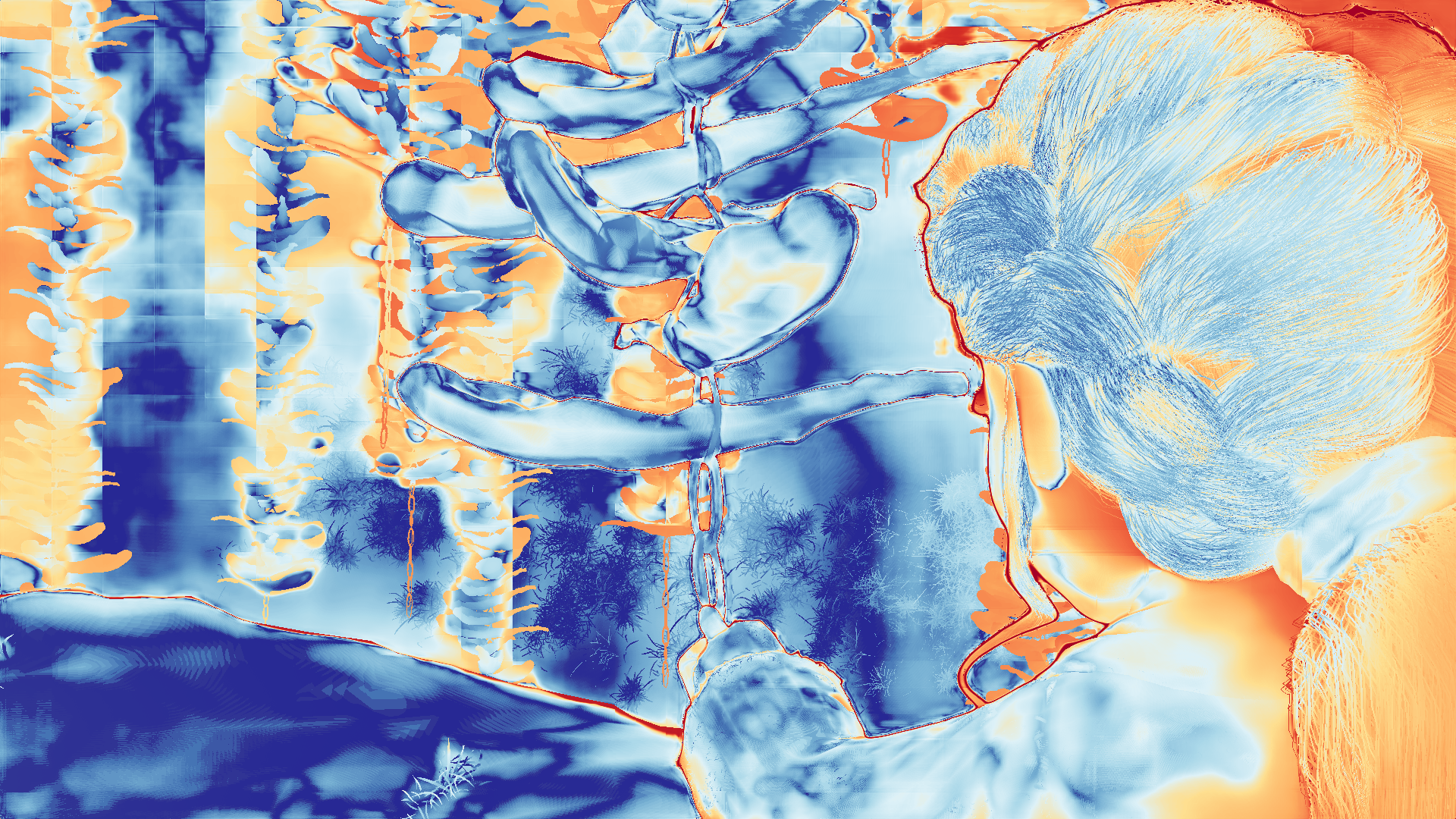}};
\end{tikzpicture}
\end{tabular}
}
\vspace{-2mm}
\caption{Qualitative results of our method and the current SOTA on Spring.
\label{fig:qulaitative_spring}
}
\end{figure*}

\begin{figure*}
\centering{
\setlength\tabcolsep{1pt}
\newcommand*\imgtrimtop{3cm}
\tikzset{labelstyle/.style={anchor=north west, fill=black, inner sep=1, text opacity=1, fill opacity=0.7, scale=0.5, xshift=3, yshift=-3, text=white}}
\tikzset{imgstyle/.style={inner sep=0,anchor=south west}}
\tikzset{rectstyle/.style={draw=black,densely dotted}}
\tikzset{rectstyle2/.style={draw=white, densely dotted}}
\newcommand{\drawlabel}[1]{%
    \draw (img.north west) node[labelstyle] {#1};
}
\newcommand{\drawrecttop}{
    \coordinate (A1) at (1.6,0.27);
    \coordinate (A2) at (3.4,0.7);
    \draw[rectstyle2] (A1) rectangle (A2);
}

\newcommand{\drawrectbottom}{
    \coordinate (B1) at (0.2,0.2);
    \coordinate (B2) at (1.45,0.76);
    \draw[rectstyle2] (B1) rectangle (B2); 
}

\newcommand{\croppedincludegraphics}[2][]{%
    \includegraphics[trim=0 9.0cm 0 3.8cm, clip, #1]{#2}%
}

\begin{tabular}{ccccc}
 target disparity & \textit{D2} error & optical flow & \textit{Fl} error& \textit{SF} error\\
\begin{tikzpicture}
\draw (0, 0) node[imgstyle] (img) {
\includegraphics[width=0.195\textwidth]{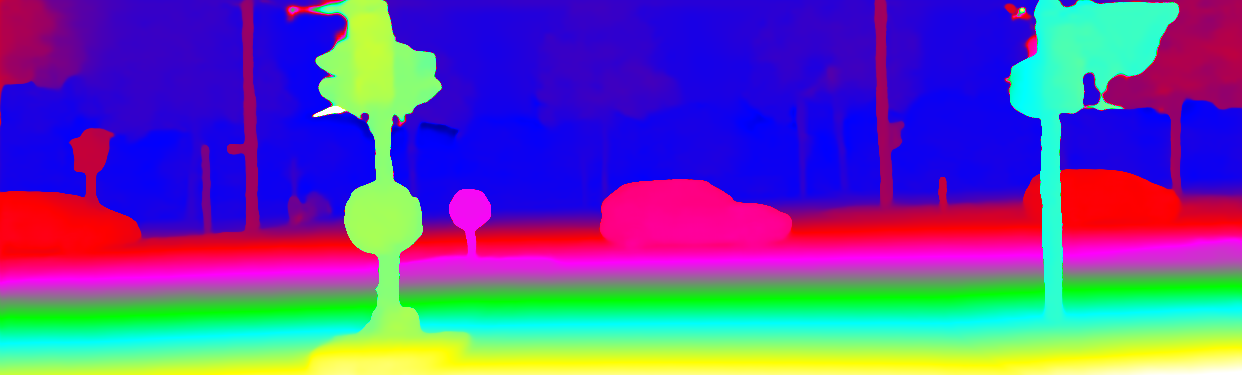}};
\drawrecttop
\drawlabel{CamLiRAFT \cite{CamLiRAFT}}
\end{tikzpicture} &
\begin{tikzpicture}
\draw (0, 0) node[imgstyle] (img) {
\includegraphics[width=0.195\textwidth]{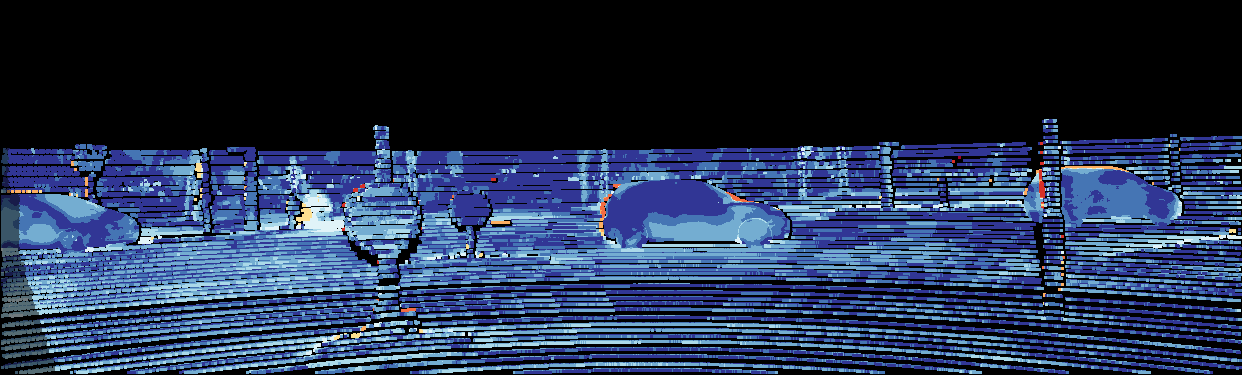}};
\drawrecttop
\drawlabel{D2-fg: 1.37, D2-bg: 0.42, D2-all: 0.66}
\end{tikzpicture} &
\begin{tikzpicture}
\draw (0, 0) node[imgstyle] (img) {
\includegraphics[width=0.195\textwidth]{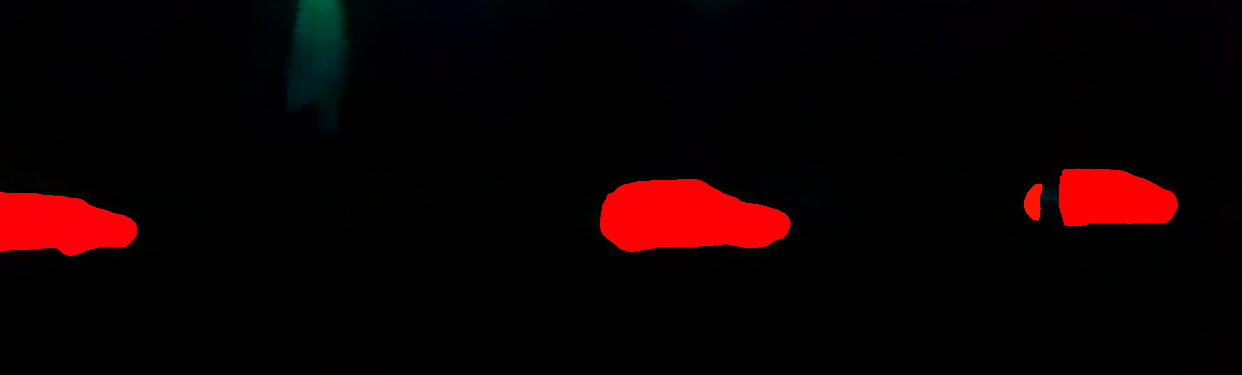}};
\drawrecttop
\end{tikzpicture} &
\begin{tikzpicture}
\draw (0, 0) node[imgstyle] (img) {
\includegraphics[width=0.195\textwidth]{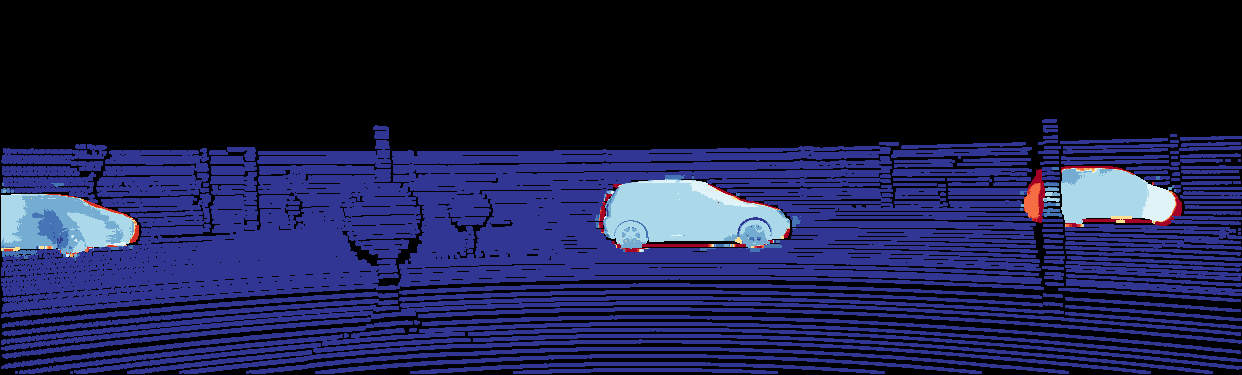}};
\drawrecttop
\drawlabel{Fl-fg: 6.85, Fl-bg: 0.42, Fl-all: 2.06}
\end{tikzpicture} &
\begin{tikzpicture}
\draw (0, 0) node[imgstyle] (img) {
\includegraphics[width=0.195\textwidth]{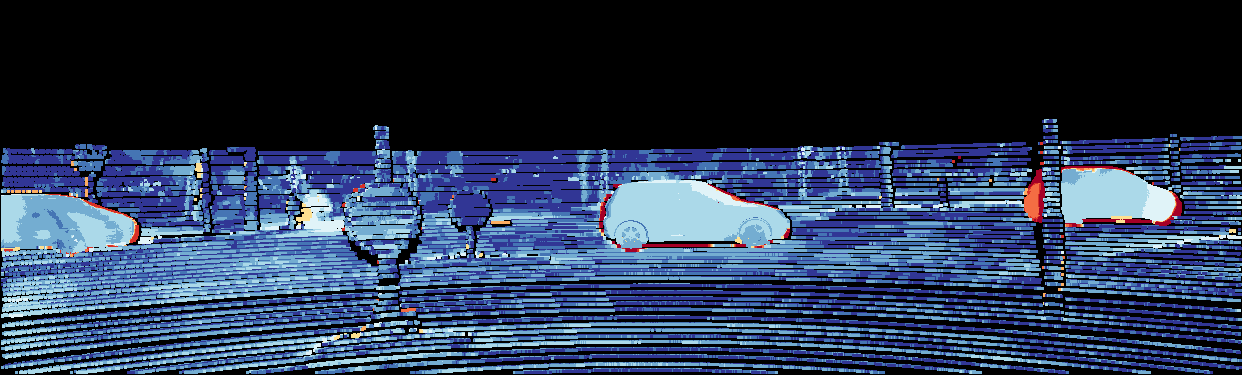}};
\drawrecttop
\drawlabel{SF-fg: 6.92, SF-bg: 0.80, SF-all: 2.36}
\end{tikzpicture}
\\[-0.6mm]
\begin{tikzpicture}
\draw (0, 0) node[imgstyle] (img) {
\includegraphics[width=0.195\textwidth]{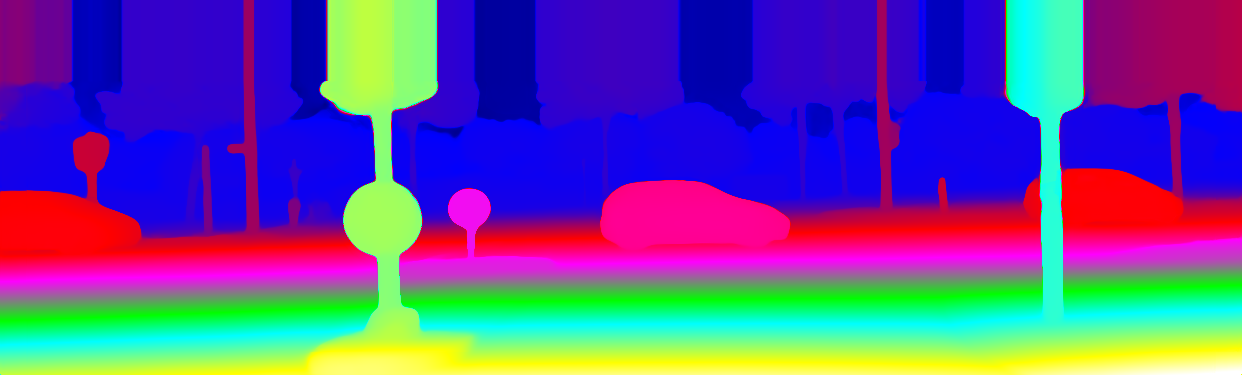}};
\drawrecttop
\drawlabel{MS-RAFT-3D+}
\end{tikzpicture} &
\begin{tikzpicture}
\draw (0, 0) node[imgstyle] (img) {
\includegraphics[width=0.195\textwidth]{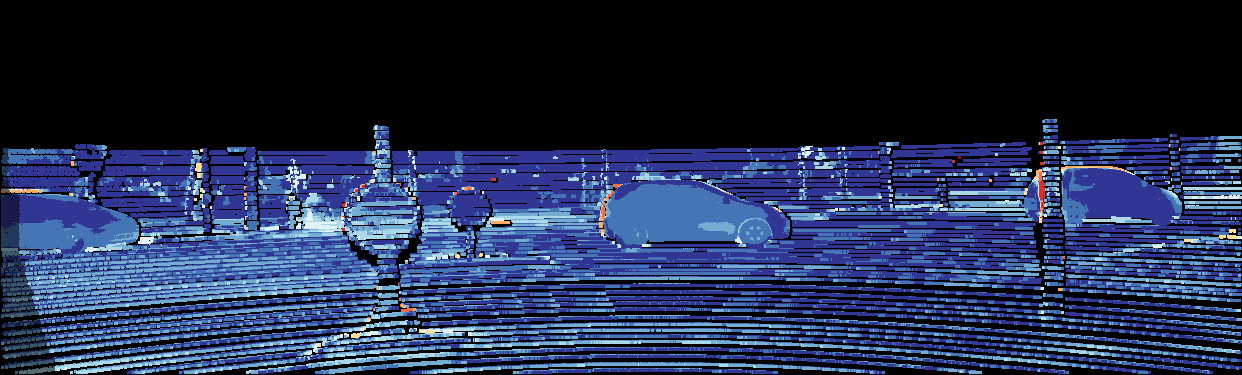}};
\drawrecttop
\drawlabel{D2-fg: 1.28, D2-bg: 0.38, D2-all: 0.61}
\end{tikzpicture} &
\begin{tikzpicture}
\draw (0, 0) node[imgstyle] (img) {
\includegraphics[width=0.195\textwidth]{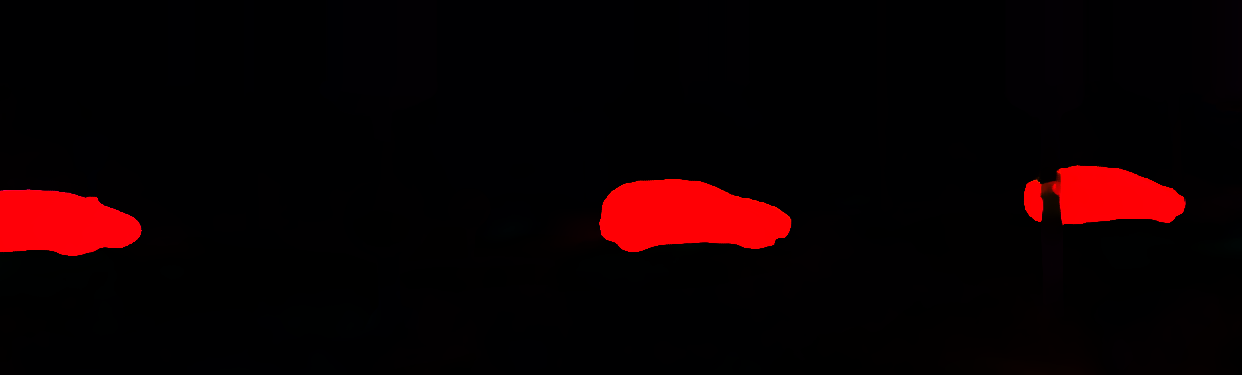}};
\drawrecttop
\end{tikzpicture} &
\begin{tikzpicture}
\draw (0, 0) node[imgstyle] (img) {
\includegraphics[width=0.195\textwidth]{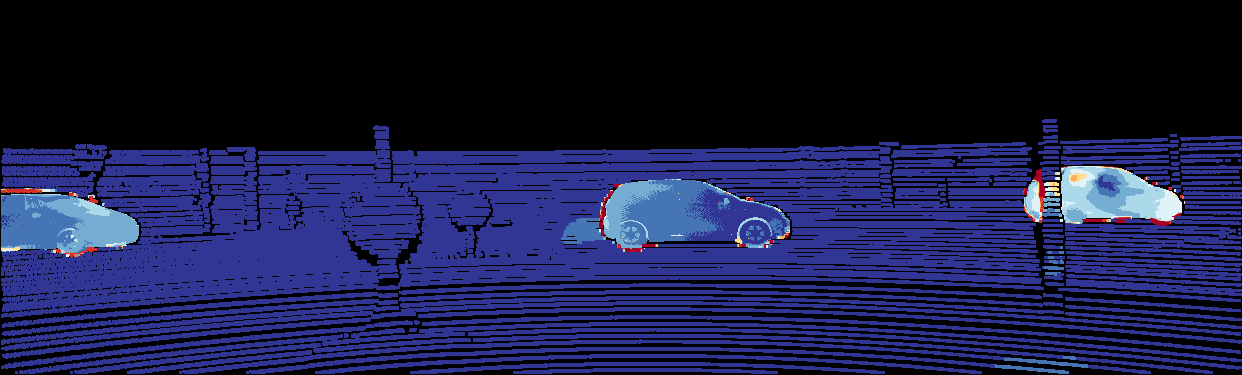}};
\drawrecttop
\drawlabel{Fl-fg: 2.52, Fl-bg: 0.43, Fl-all: 0.96}
\end{tikzpicture} &
\begin{tikzpicture}
\draw (0, 0) node[imgstyle] (img) {
\includegraphics[width=0.195\textwidth]{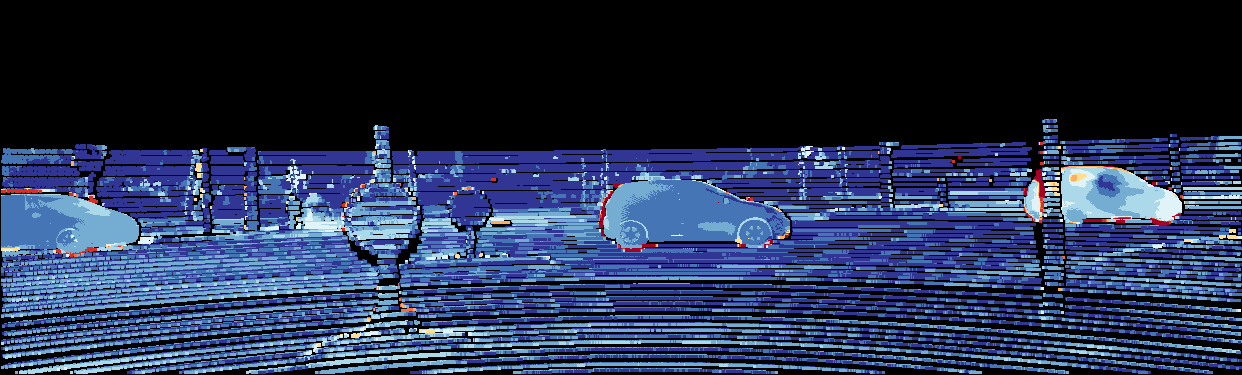}};
\drawrecttop
\drawlabel{SF-fg: 2.92, SF-bg: 0.73, SF-all: 1.29}
\end{tikzpicture}
\\[-0.6mm]
\begin{tikzpicture}
\draw (0, 0) node[imgstyle] (img) {
\includegraphics[width=0.195\textwidth]{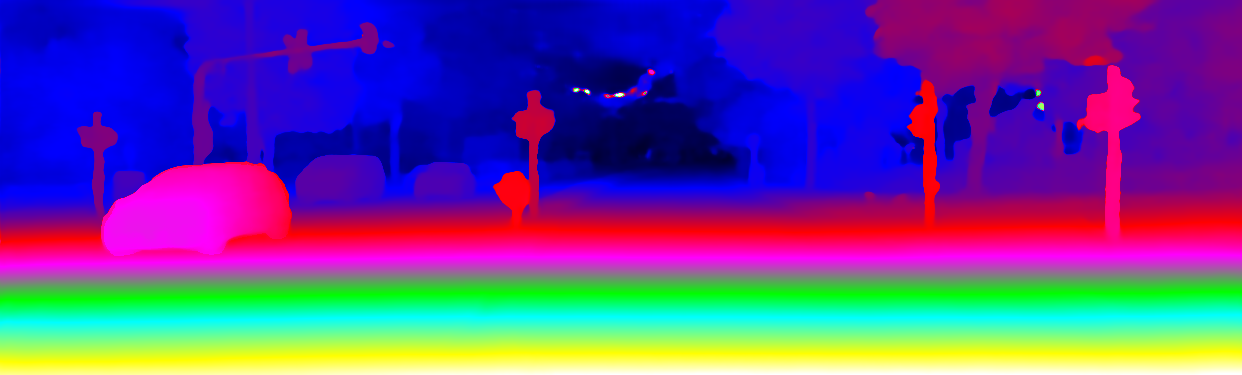}};
\drawrectbottom
\drawlabel{CamLiRAFT \cite{CamLiRAFT}}
\end{tikzpicture} &
\begin{tikzpicture}
\draw (0, 0) node[imgstyle] (img) {
\includegraphics[width=0.195\textwidth]{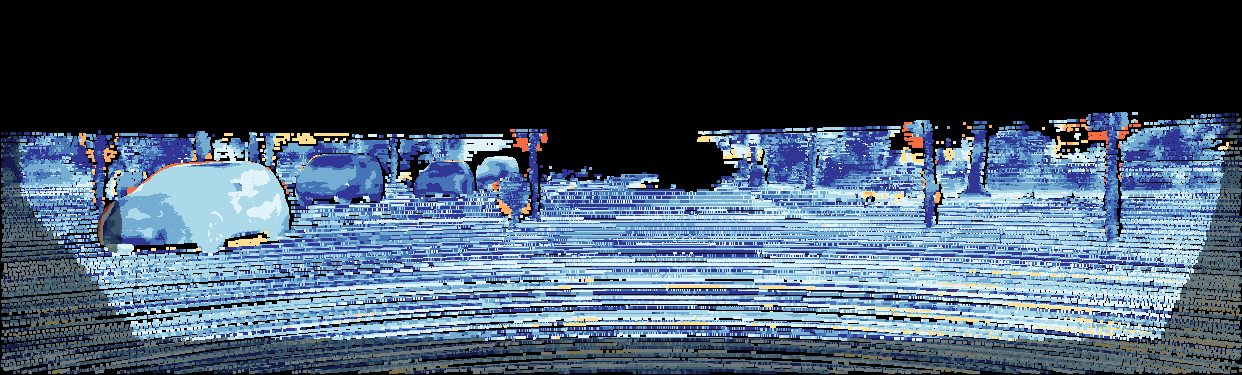}};
\drawrectbottom
\drawlabel{D2-fg: 2.13, D2-bg: 2.37, D2-all: 2.31}
\end{tikzpicture} &
\begin{tikzpicture}
\draw (0, 0) node[imgstyle] (img) {
\includegraphics[width=0.195\textwidth]{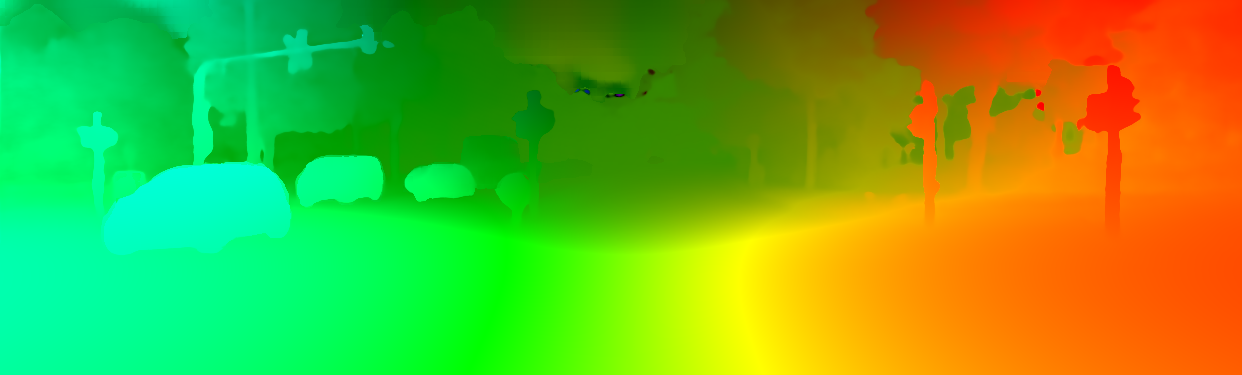}};
\drawrectbottom
\end{tikzpicture} &
\begin{tikzpicture}
\draw (0, 0) node[imgstyle] (img) {
\includegraphics[width=0.195\textwidth]{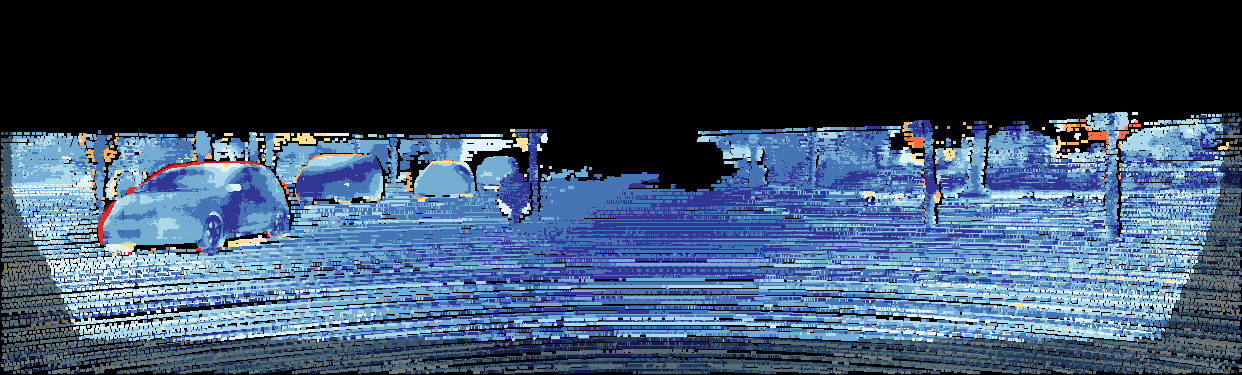}};
\drawrectbottom
\drawlabel{Fl-fg: 5.07, Fl-bg: 0.97, Fl-all: 1.91}
\end{tikzpicture} &
\begin{tikzpicture}
\draw (0, 0) node[imgstyle] (img) {
\includegraphics[width=0.195\textwidth]{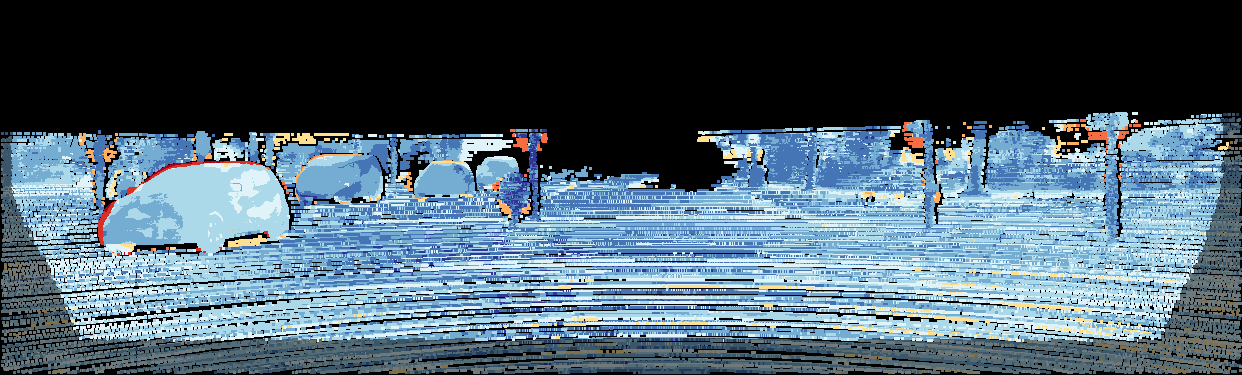}};
\drawrectbottom
\drawlabel{SF-fg: 5.15, SF-bg: 2.48, SF-all: 3.09}
\end{tikzpicture}
\\[-0.6mm]
\begin{tikzpicture}
\draw (0, 0) node[imgstyle] (img) {
\includegraphics[width=0.195\textwidth]{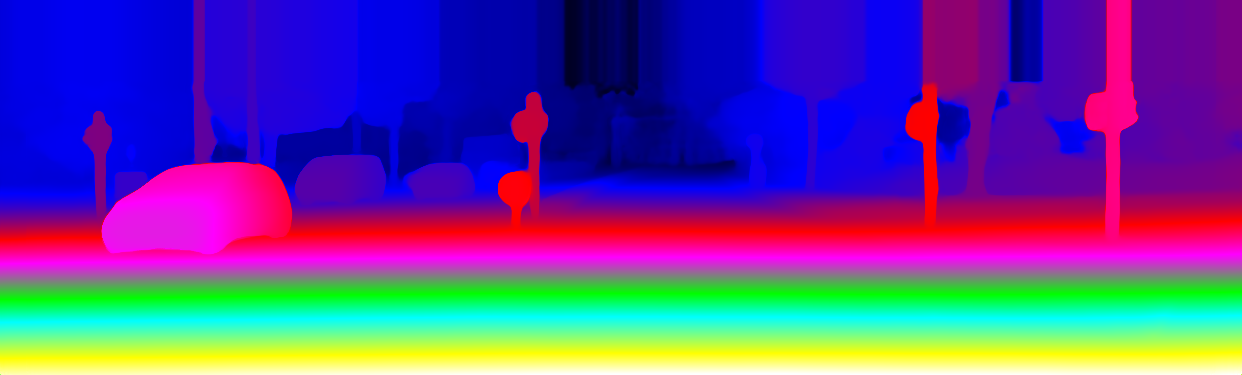}};
\drawrectbottom
\drawlabel{MS-RAFT-3D+}
\end{tikzpicture} &
\begin{tikzpicture}
\draw (0, 0) node[imgstyle] (img) {
\includegraphics[width=0.195\textwidth]{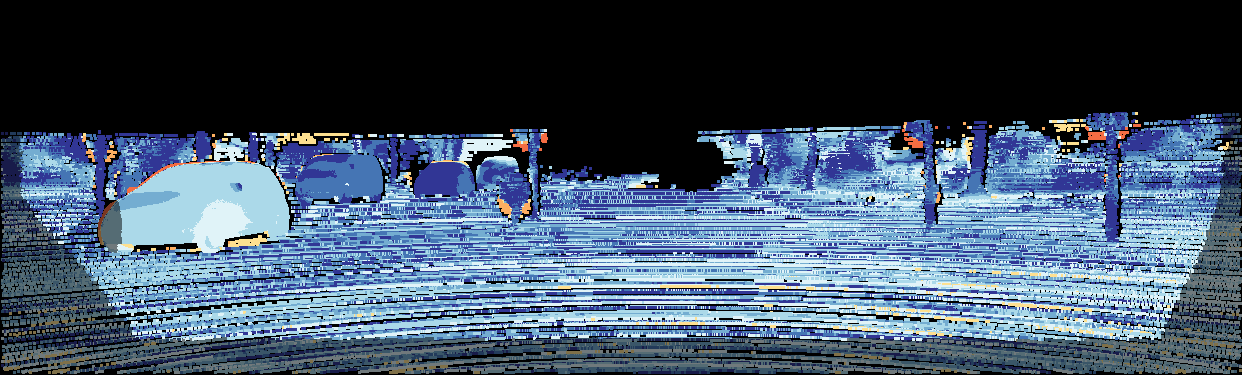}};
\drawrectbottom
\drawlabel{D2-fg: 2.60, D2-bg: 1.86, D2-all: 2.03}
\end{tikzpicture} &
\begin{tikzpicture}
\draw (0, 0) node[imgstyle] (img) {
\includegraphics[width=0.195\textwidth]{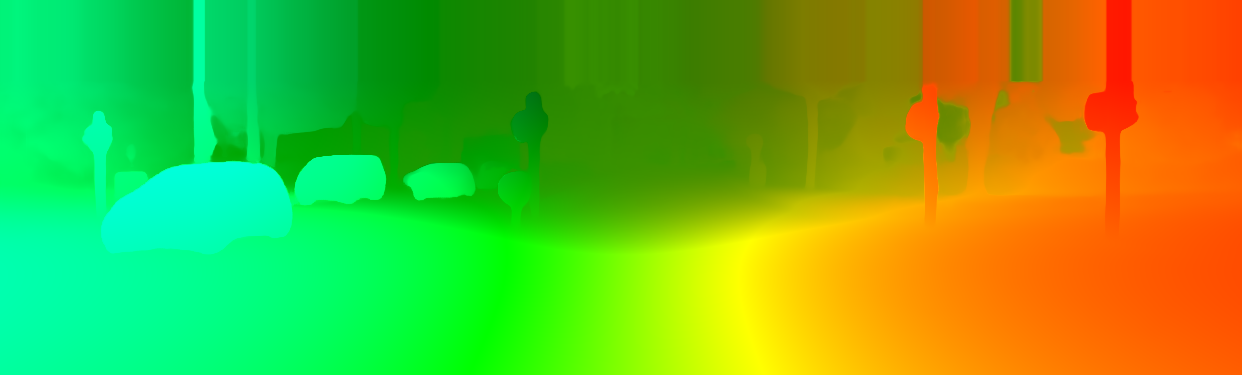}};
\drawrectbottom
\end{tikzpicture} &
\begin{tikzpicture}
\draw (0, 0) node[imgstyle] (img) {
\includegraphics[width=0.195\textwidth]{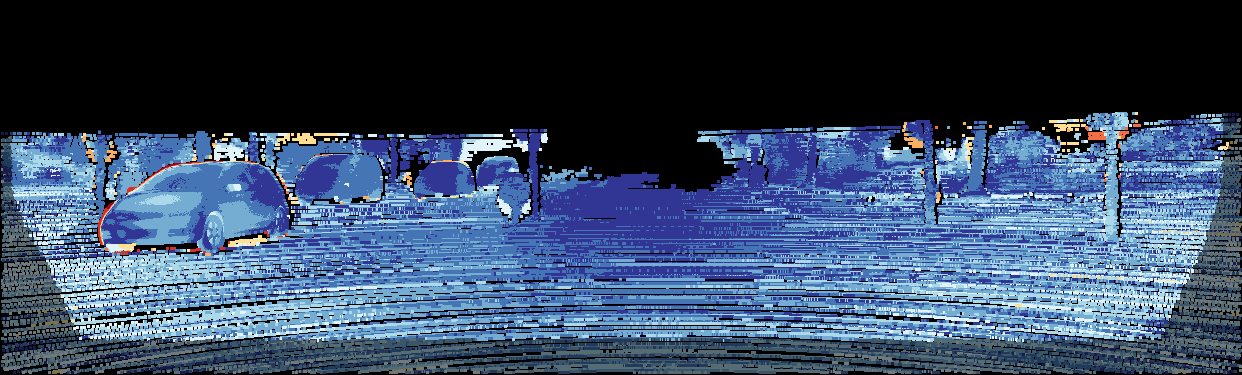}};
\drawrectbottom
\drawlabel{Fl-fg: 3.09, Fl-bg: 0.77, Fl-all: 1.30}
\end{tikzpicture} &
\begin{tikzpicture}
\draw (0, 0) node[imgstyle] (img) {
\includegraphics[width=0.195\textwidth]{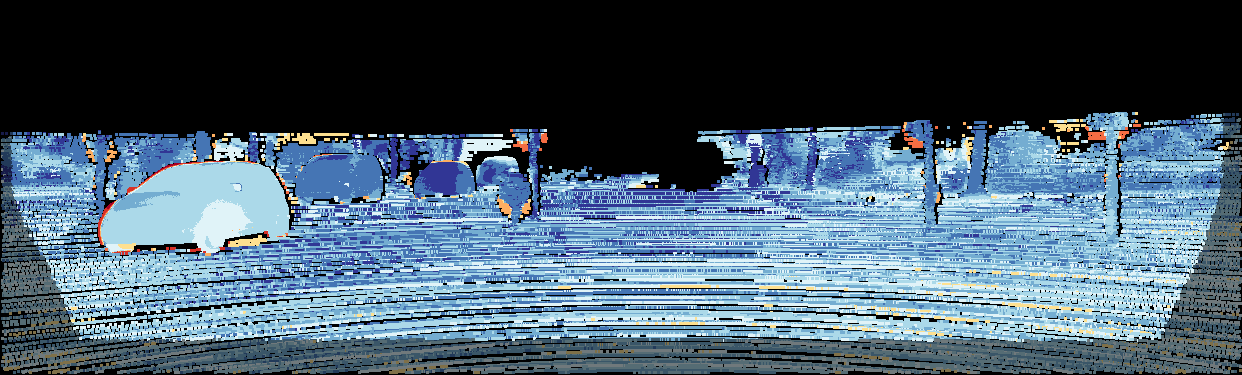}};
\drawrectbottom
\drawlabel{SF-fg: 3.13, SF-bg: 1.97, SF-all: 2.23}
\end{tikzpicture}

\end{tabular}
}
\vspace{-2mm}
\caption{
Qualitative results of our method and the current SOTA on KITTI.
\label{fig:qulaitative_KITTI}
}
\end{figure*}
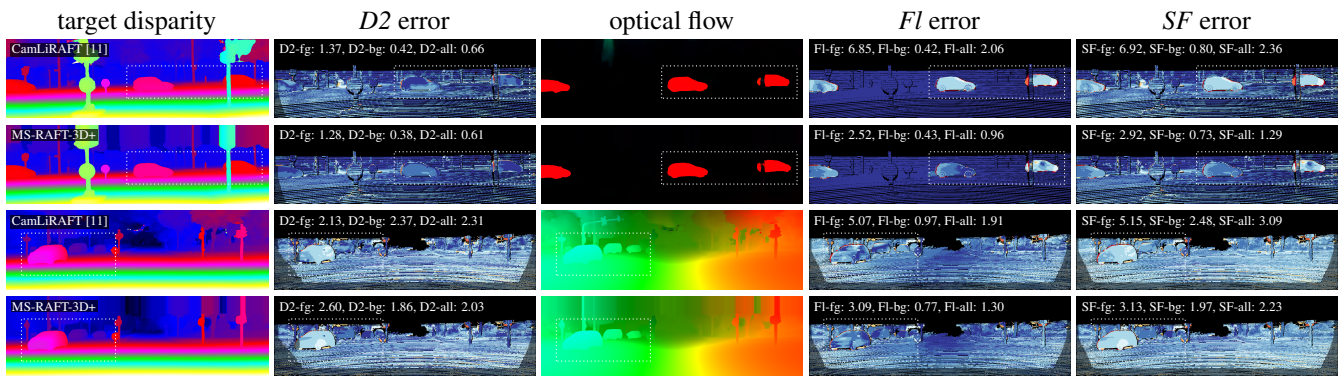

We present more qualitative results of our method from the Spring benchmark in \Cref{fig:qulaitative_spring} and from the KITTI benchmark in \Cref{fig:qulaitative_KITTI}.
In both cases, our approach achieves detailed results and lower errors. 
Importantly, in the case of KITTI, as the top 80 pixels of samples are not considered in the evaluation, they are also not computed but extended from the last row's estimate, as in  RAFT-3D \cite{RAFT-3D}.

\end{document}